%% file: output.tex
\documentclass[10pt,journal,compsoc]{IEEEtran}

\usepackage[OT1]{fontenc}

\usepackage{booktabs}
\usepackage{times}
\usepackage{epsfig}
\usepackage{graphicx}
\usepackage{mathtools,amsmath,amsfonts,amssymb,bm}
\usepackage{adjustbox}
\usepackage{multicol,multirow}
\usepackage{siunitx}
\usepackage{makecell}
\usepackage{float}
\usepackage[hyphens]{url}
\usepackage{placeins}
\usepackage{subcaption} 
\usepackage{cite}

\usepackage[ruled,noend,linesnumbered]{algorithm2e}
\SetKwComment{Comment}{$\triangleright$\ }{}

\usepackage{pgfplots}
\usepackage{pgfplotstable}
\usepgfplotslibrary{groupplots,dateplot,colorbrewer,fillbetween}
\usetikzlibrary{patterns,shapes.arrows,external}
\pgfplotsset{compat=newest}

\newif\ifarxiv
\arxivtrue 

\def\MYTITLE{Real-time Noise Source Estimation of a Camera System from an Image and Metadata}

\ifarxiv
\definecolor{eccvblue}{rgb}{0.12,0.49,0.85}
\usepackage[pagebackref=false,breaklinks=true,colorlinks,bookmarks=true,bookmarksnumbered=true,citecolor=eccvblue]{hyperref}
\else 
\usepackage[bookmarks=false,breaklinks,hidelinks]{hyperref}
\fi
\hypersetup{
  pdftitle={\MYTITLE},
  pdfsubject={Computer Vision, State Estimation},
  pdfauthor={Maik Wischow, Patrick Irmisch, Anko Boerner, Guillermo Gallego},
  pdfkeywords={Sensor AI, Smart Sensing Systems, Noise Estimation, Perception, Deep Learning, Denoising}
}

\usepackage{orcidlink}

\usepackage[capitalize]{cleveref}
\crefname{section}{Section}{Sections}
\Crefname{section}{Section}{Sections}
\Crefname{table}{Table}{Tables}
\crefname{table}{Table}{Tables}
\Crefname{figure}{Figure}{Figures}
\crefname{figure}{Figure}{Figures}

\input{mathMacros}

\makeatletter
\long\def\@IEEEtitleabstractindextextbox#1{\parbox{0.922\textwidth}{#1}}
\makeatother

\ifarxiv
\usepackage[absolute]{textpos}
\fi

\begin{document}

\title{\MYTITLE}

\ifarxiv
\definecolor{somegray}{gray}{0.5}
\newcommand{\darkgrayed}[1]{\textcolor{somegray}{#1}}
\begin{textblock}{11}(2.5, 0.4)  
\begin{center}
\darkgrayed{This journal paper has been accepted for publication at Advanced Intelligent Systems, 2024.}
\end{center}
\end{textblock}
\fi 

\author{%
Maik Wischow$^{1,2}$ \orcidlink{0000-0001-5777-3475}, %
Patrick Irmisch$^1$ \orcidlink{0009-0004-3621-530X}, %
Anko Boerner$^1$ \orcidlink{0000-0002-7176-3588},
Guillermo Gallego$^{2,3}$ \orcidlink{0000-0002-2672-9241}\\
$^{1}$German Aerospace Center, Germany.\\
$^{2}$Technische Universit\"at Berlin, Germany.\\
$^{3}$Einstein Center Digital Future and SCIoI Excellence Cluster, Germany.
\ifarxiv
\thanks{Preprint of accepted paper, doi: 10.1002/aisy.202300479}
\fi
}

\input{abstract}
\maketitle
\input{introduction}
\input{relatedWork}
\input{noiseSourceEstimation}
\input{experiments}
\input{conclusion}

\ifarxiv
\newpage 
\input{appendix}

\clearpage 
\bibliographystyle{IEEEtran}
\input{output.bbl}

\else
\bibliographystyle{IEEEtran}

\input{output.bbl}
\cleardoublepage \input{appendix}
\fi 

\end{document}

%% file: mathMacros.tex
\def\estimatedNoiseLevel{\hat\sigma}
\def\totalNoise{\sigma_{\text{Total}}}
\def\imageNoise{\sigma_{\text{Image}}}
\def\modelNoise{\sigma_{\text{Model}}}
\def\photonNoise{\sigma_{\text{PN}}}
\def\dcNoise{\sigma_{\text{DCSN}}}
\def\readoutNoise{\sigma_{\text{RN}}}
\def\restNoise{\xi_{\text{M/I}}}

\def\totalNoiseEstimate{\hat\sigma_{\text{Total}}}

\def\photonNoiseEstimate{\hat\sigma_{\text{PN}}}
\def\dcNoiseEstimate{\hat\sigma_{\text{DCSN}}}
\def\readoutNoiseEstimate{\hat\sigma_{\text{RN}}}
\def\restNoiseEstimation{\hat\xi_{\text{M/I}}}

\DeclareSIUnit\px{px}
\DeclareSIUnit\lines{lines}
\DeclareSIUnit\patch{patch}
\DeclareSIUnit\DN{DN}
\DeclareSIUnit\electron{e^-}

\def\synth{\emph{Sim}}
\def\cellar{\emph{Cellar}}
\def\parkingLot{\emph{Parking Lot}}
\def\EV76C661{\emph{EV76C661}}
\def\ICX285{\emph{ICX285}}
\def\OV2740{\emph{OV2740}}

\def\baseline{$\text{\emph{DRNE}}_\text{cust.}$}
\def\withoutMetadata{\emph{w/o-Meta}}
\def\minimalMetadata{\emph{Min-Meta}}
\def\fullMetadata{\emph{Full-Meta}}
\def\B+F{\emph{B+F}}
\def\PCA{\emph{PCA}}
\def\PGE{\emph{PGE-Net}}

\def\baselineTitle{$\text{DRNE}_\text{cust.}$}
\def\withoutMetadataTitle{W/o-Meta}
\def\minimalMetadataTitle{Min-Meta}
\def\fullMetadataTitle{Full-Meta}
\def\PGETitle{PGE-Net}

\newcommand{\gblue}[1]{#1} 

%% file: abstract.tex
\IEEEtitleabstractindextext{%
\begin{abstract}
Autonomous machines must self-maintain proper functionality to ensure the safety of humans and themselves.
This pertains particularly to its cameras as predominant sensors to perceive the environment and support actions.
A fundamental camera problem addressed in this study is noise.
Solutions often focus on denoising images a posteriori, that is, fighting symptoms rather than root causes.
However, tackling root causes requires identifying the noise sources, considering the limitations of mobile platforms.
This work investigates a real-time, memory-efficient and reliable noise source estimator that combines data- and physically-based models.
To this end, a DNN that examines an image with camera metadata for major camera noise sources is built and trained.
In addition, it quantifies unexpected factors that impact image noise or metadata.
This study investigates seven different estimators on six datasets that include synthetic noise, real-world noise from two camera systems, and real field campaigns.
For these, only the model with most metadata is capable to accurately and robustly quantify all individual noise contributions.
This method outperforms total image noise estimators and can be plug-and-play deployed.
It also serves as a basis to include more advanced noise sources, or as part of an automatic countermeasure feedback-loop to approach fully reliable machines.
\end{abstract}
}

%% file: introduction.tex
\ifarxiv
\section*{Project page}
\url{https://github.com/MaikWischow/Noise-Source-Estimation}
\vspace{-1ex}
\fi

\section{Introduction}
\label{sec:intro}
Machines in various fields (e.g., vehicles, robots) are increasingly moving away from manual control towards autonomy, which implies that they should ensure proper operation (e.g., \cite{Atab2020Soft,Jin2020Automated,Zhang2022Human}).
This applies to each component of a machine, and in particular to its perception system, as all subsequent actions depend on it.
Cameras are the predominant sensors for perceiving the environment and are therefore the subject of our study.
As any physical sensor, a camera is afflicted with noise, whose influence on subsequent computer vision tasks has justified extensive research.
To guarantee a machine's dependability and durability, which in turn guarantees the safety of both humans and machines, counteracting noise is mandatory.
However, to counteract noise in an active system, 
one needs first to identify and quantify its root causes.

Previous studies approach this task by using estimated noise levels to denoise images \cite{Chen15iccv,jain2008natural,tan2019pixelwise,wang2019enhancing,zhang2017beyond}.
This process has matured for various noise models and use cases, but they often yield undesired visual artifacts.
That is, they only fight symptoms and do not target noise source identification, although noise sources and countermeasures are well researched \cite[ch.~7]{janesick2001Scientific}.
This can be attributed to three reasons:
($i$) The camera system control is often inaccessible, which makes denoising more applicable if only image datasets are available. 
($ii$) The need for more autonomy of machines with consumer-grade cameras emerged only recently and noise could only be approached manually so far. 
($iii$) Reliable and real-time noise source estimation is challenging;
it relies on accurate image noise estimation and extensive noise models, which gained interest and matured only in recent years (see~\cref{sec:relatedWork}). 
Moreover, noise source identification from an image alone is ambiguous, since most noise sources follow similar statistics --auxiliary data is needed for disambiguation.
Last but not least, at present, only deep neural networks (DNNs) are able to perform the implied complex operations (extensive noise modeling and heterogeneous data-fusion) in real-time.

\input{figEyeCatcher}
This paper proposes a real-time, memory-efficient, and reliable noise source estimator (\cref{fig:eyeCatcher}).
During operation it analyzes single images together with metadata from the camera system and quantifies the respective contributions to major noise sources of the system. 
Moreover, we include a verification mechanism that quantifies noise mismatches between the metadata and the image noise, which serves for self-control and detection of unexpected events (e.g., camera damages).
Without loss of generality, our study analyzes: time-varying noise (since any time-invariant noise is usually mitigated by camera calibration), and spatially-varying noise (since image patches are used).

 We make the following technical contributions:
 \begin{itemize}
     \item We propose a real-time, memory-efficient, and reliable DNN-based noise source estimator (\cref{sec:noiseSourceEstimation}) that is able to quantify contributions of different camera noise sources and to detect unexpected factors that impact image noise or metadata.
     \item We demonstrate seven different estimators in comprehensive experiments on six datasets and two real camera systems (\cref{sec:experiments}).
     Our experiments investigate synthetic noise, real-world noise extracted from camera systems, qualitative field campaigns, and also create unexpected noise events in images or metadata.
     \item We provide the source code of our experiments, the data used for training and benchmarking, and the ready-to-use estimators. 
     \ifarxiv
     \else
        \url{https://github.com/MaikWischow/Noise-Source-Estimation}
     \fi
 \end{itemize}

%% file: figEyeCatcher.tex
\begin{figure}[t]
\centering
\includegraphics[width=1.0\columnwidth, trim=0 0 0 0, clip]{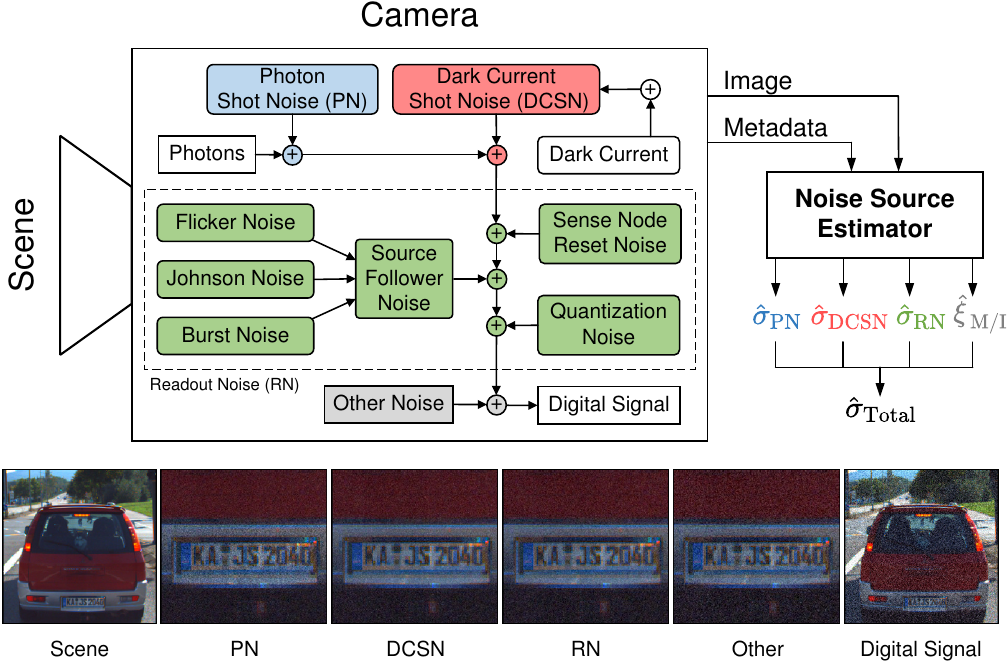}
\caption{\emph{Proposed Camera Noise Source Estimation.} 
Different noise sources affect the image formation process of a scene. 
Our noise source estimator quantifies major noise source contributions $\hat\sigma_{i \in \{ \text{PN, DCSN, RN}\}}$, unexpected noise $\restNoiseEstimation$, and the total image noise $\hat\sigma_{\text{Total}}$ using an image and camera metadata.
}
\label{fig:eyeCatcher}
\end{figure}

%% file: relatedWork.tex
\section{Related Work}
\label{sec:relatedWork}
We first survey general image noise level estimators and then discuss noise models from related studies that account for multiple noise sources and utilize camera metadata.

\textbf{Noise Level Estimation. }%
Motivated by applications in the field, we focus our study on estimators that assume unknown noise levels (i.e., \emph{blind} estimation) using \emph{single} images. 
These may be further divided into \emph{traditional} and \emph{learning-based} approaches.
Traditional approaches comprise one or more of the following paradigms: 
($i$) \emph{block-based} \cite{shin2005block,uss2011image} (estimate noise using low-textured regions), 
($ii$) \emph{filtering-based} \cite{corner2003noise,shin2005block} (subtract a low-pass filtered image and estimate noise from high frequency components) and 
($iii$) \emph{transform-based} \cite{de2004training,pyatykh2012image,Chen15iccv} (represent the image in a different space, e.g., using wavelets, and estimate noise therein).
All have their own pros and cons, with over-/underestimation in low/high noise and textured areas.

Learning-based methods either determine the noise level \emph{explicitly} \cite{zhang2017beyond,tan2019pixelwise,byun2021fbi} (e.g., using residual learning and scale pyramids) or \emph{implicitly} \cite{jain2008natural,lyu2020degan} (e.g., with generative adversarial networks) as part of an end-to-end denoising pipeline. 
In terms of real-image denoising performance, traditional methods are still considered the state of the art, closely followed by learning-based methods \cite{plotz2017benchmarking}.

\textbf{Noise Models. }%
Driven by Space camera systems, extensive noise models on a subatomic level have been developed in recent decades \cite{healey1994radiometric,janesick2001Scientific,konnik2014high}.
However, applications on Earth tend to employ simpler models, as follows.

The majority of research presumes an additive white Gaussian noise (AWGN) source \cite{shin2005block,uss2011image, corner2003noise,de2004training,pyatykh2012image,Chen15iccv,zhang2017beyond,jain2008natural}.
Given the influence of light on camera noise \cite{blanksby1997noise}, 
signal-dependent noise models have been developed considering ($i$) photon shot noise and ($ii$) noise due to camera electronics (e.g., the Poissonian-Gaussian noise model) \cite{foi2008practical, tan2019pixelwise, byun2021fbi}.
A special case is the noise level function (NLF) that characterizes the dependence of noise levels on image intensity \cite{liu2007automatic,sutour2015estimation,yang2015estimation}.
To account for non-linear camera processes that affect noise statistics \cite{matsushita2007radiometric}, some works employ the camera response function for NLF estimation; 
they describe a camera's physical processing as a black-box in a single function \cite{yao2016novel,yao2021signal}.

A few noise studies break down the noise caused by camera electronics and therefore consider more than two noise sources \cite{wang2019enhancing, wei2020physics, zhang2021rethinking, onzon2021neural}, but generally ``[...] \emph{noise sources caused by digital camera electronics are still largely overlooked, despite their significant effect on raw measurement}'' \cite{wei2020physics}.
The works \cite{wei2020physics, onzon2021neural} propose ``simpler'' extensive noise models that account, e.g., for the camera system gain, read noise, or quantization noise, which are partially analyzed in more detail.
More sophisticated noise models from \cite{wang2019enhancing} and \cite{zhang2021rethinking} also address camera specifics like the shutter mechanism, individual color channel biases or differentiate between analog/digital gain.
There have also been attempts to approximate noise models by DNNs \cite{chen2018image, abdelhamed2019noise, chang2020learning, chen2022Temperature} for synthesis, but \cite{zhang2021rethinking} shows that ``\emph{The DNN-based [noise generators] still cannot outperform physics-based statistical methods}''.

All the above models calibrate their parameters (temperature, exposure time, ISO gain, ...) offline 
and only implicitly account for changing camera parameters during training data generation, 
but they do not consider camera parameters at inference time. 
We investigate this gap and show in our experiments on DNN noise level estimation that the system is only able to identify the contribution of different noise sources when these parameters are available at runtime.
Furthermore, to the best of our knowledge, our approach is the first estimator (traditional or learning-based) to explicitly quantify not only two (PN, other) but four (PN, DCSN, RN, other) individual noise source contributions.

%% file: noiseSourceEstimation.tex
\section{Noise Source Estimation}
\label{sec:noiseSourceEstimation}
\input{figNNModel}
Given a possibly corrupted image patch $I'$ and metadata from the camera system, 
the goal of our image noise source estimator is to determine the image's total noise level
\begin{equation}
\totalNoise = \sqrt{\photonNoise^2 + \dcNoise^2 + \readoutNoise^2} + \restNoise
\label{eq:totalNoise}%
\end{equation}
\emph{and its individual components}: the photon shot noise (PN) level $\photonNoise$, the dark current shot noise (DCSN) level $\dcNoise$, the readout noise (RN) level $\readoutNoise$ and a component $\restNoise$ that quantifies unexpected (i.e., residual) noise (details about noise types are in the supplementary material).
We assume grayscale patches, of size $128 \times 128~\si{\px}$.
Next, we describe the base architecture (\cref{subsec:baseline}), subsequently detail our extensions (\cref{subsec:extendBaseline}), and lastly focus on training the noise source estimator (\cref{subsec:training}).

\subsection{Base Architecture}
\label{subsec:baseline}%
Our method is inspired by the deep residual noise-level estimator DRNE from \cite{tan2019pixelwise}, 
which has been shown to be superior compared to traditional state-of-the-art approaches in terms of runtime and accuracy \cite{wischow2021camera}.
It takes an RGB image patch as input and predicts a pixel-wise noise level. 
It consists of 16 convolution layers (with \num{15} of them separated into three residual blocks). 
Pooling layers, interpolation operations and convolution strides larger than $3 \times 3$ are omitted to keep the focus on low-level noise features.

We customize the above architecture so that the neural network takes grayscale images as input and estimates only one noise level per image patch (left part of \cref{fig:NNModels}).
Specifically, we replace the first $3 \times 3 \times 3$ convolution kernel by a $3\times 3$ one, replace the last residual block by a fully connected block (FCB) with three layers having \num{32}, \num{16}, and \num{8} neurons, respectively, and apply global max pooling before the FCB to fit the dimensions.
As a consequence, we are able to reduce the total number of network parameters by 35\%, from \num{519}k to \num{336}k 
while achieving similar estimation accuracy as \cite{tan2019pixelwise}.
Lastly, we retrain the network as described in \cref{subsec:training}. 
In the upcoming sections we refer to this customized model as \baseline{}.

\subsection{Noise Source Estimation}
\label{subsec:extendBaseline}%
The previous method estimates the noise level of the patch, but does not identify its origin (i.e., type and amount of noise), which is critical information for a camera's maintenance operation.
\gblue{%
In order to identify the noise origin, additional information is needed alongside the noised image.
Our approach is to train the baseline network on a physical noise model \cite{konnik2014high} that relates image intensity and camera metadata to different noise distributions.
To this end, the baseline network needs to be extended to separate the different noise contributions and to process image and metadata together.
Moreover, we expect an improved noise estimation accuracy as a result of learned awareness of separate noise sources and thus increased physical consistency.
}

In the following, we describe the three major extensions to the above method for noise \emph{source} estimation (right part of \cref{fig:NNModels}): noise type identification (with or without the inclusion of camera metadata), and quantification of unexpected noise.

\textbf{1.~Noise Type Identification / Separation. }%
In a first step we duplicate the FCB and its preceding global max pooling layer to get three independent network branches.
Each branch will predict the noise level of one noise type.

\textbf{2.~Inclusion of Camera Metadata. }%
In a second step we separate the camera's metadata pertaining to the noise model into fixed and variable metadata (see \cref{tab:cameraMetadata}).
We assume the fixed metadata to be constant at training and inference times due to multiple reasons: 
($i$) Only parameters that in a sensitivity analysis lead to significant noise changes in the noise model are picked as variable parameters (see supp. material). From these parameters, we also fix ($ii$) the offset, for simplicity, and the ones that ($iii$) we consider as too difficult to obtain from a consumer-grade camera.

For the variable metadata, we survey existing camera systems in the literature to determine parameter ranges that are typical for our application scenarios (excluding unique systems for specialized use cases).
The variable parameters are arranged into ``minimal'' and ``full'' metadata.
We consider minimal metadata as easy to obtain\footnote{Camera gain (digital gain for simplicity) and exposure time are typically configurable, while most camera systems comprise a temperature sensor to approach dark current compensation.} and full metadata as more comprehensively include parameters often provided by the camera manufacturer.
For comparison, we derive three models, where each one is fed with different metadata: 
one without any (\withoutMetadata{}), one with minimal (\minimalMetadata{}) and one with full metadata (\fullMetadata{}).

\input{tabVariableCameraParameterRanges}
In preparation to use the metadata as input for the neural network, each parameter is first normalized to a \gblue{floating point number} in the range $[0,1]$ (using \gblue{the physical units, and the respective minimum and maximum} values in \cref{tab:cameraMetadata}).
\gblue{Subsequently, all parameters are combined and passed as a single array to the neural network (see attached source code for details)}.
\gblue{Inside the network,} the metadata subset associated to its respective noise type is then concatenated with the output of the corresponding global max pooling layer and passed into its FCB. 
Note that using FCBs over the noise model itself to estimate the noise levels is: 
($i$) fast (using a GPU), ($ii$) allows us to train on real noise data that is not covered by the noise model, and ($iii$) allows us to perform non-trivial feature-wise fusion with the feature maps from the processed input image.

\textbf{3.~Unexpected Noise Quantification. }%
In the proposed system (\cref{fig:NNModels}-right), we add a fourth FCB that quantifies unexpected noise, i.e., when the metadata does not agree with the considered image noise model.
If we ensure that image noise is only generated inside the camera system (by preventing image pre- and post-processing) and assume a radiometrically calibrated camera (including a correct determination of the relevant metadata), there are two reasons for noise-metadata mismatch: 
($i$) corrupted metadata (e.g., by camera malfunctioning) or 
($ii$) unmodeled noise sources (e.g., also by hardware damages, or a general mismatch between the noise model and the real image noise).

Specifically, we train this fourth FCB to quantify
\begin{equation}
    \begin{split}
        \restNoise  &\doteq \modelNoise - \imageNoise\\
                    &\stackrel{\eqref{eq:totalNoise}}{=} \sqrt{\photonNoise^2{\scriptstyle(M_1)} + \dcNoise^2{\scriptstyle(M_2)} + \readoutNoise^2{\scriptstyle(M_3)}} \hspace{5pt}\\
                    &{\hspace{12pt}} - \sqrt{\photonNoise^2{\scriptstyle(M'_1)} + \dcNoise^2{\scriptstyle(M'_2)} + \readoutNoise^2{\scriptstyle(M'_3)}}
    \end{split}
    \label{eq:restNoise}%
\end{equation}
with $\restNoise$ normalized to $[-1, 1]$ for training, the total image noise $\imageNoise$, the total modeled noise $\modelNoise$, and metadata sets $M_1,\ldots,M_3$, and altered sets $M'_1,\ldots,M'_3$ having a different randomly generated camera gain.
The metadata sets $M_{(\cdot)}$ are only fed to the FCBs (corresponding to noise level $\modelNoise$) while the altered sets $M'_{(\cdot)}$ are used to corrupt the image (with corresponding noise level $\imageNoise$).
In this way, the network learns to capture the mismatch between the metadata and the image noise in $\restNoise$.

With all the aforementioned extensions, the number of network parameters slightly increases, from \num{336}k to \num{345}k.

\subsection{Training Details}%
\label{subsec:training}%
We utilize an almost noise-free dataset with natural images (\emph{TAMPERE21} \cite{bahnemiri2022learning}), whose noise variance is ensured to be $\sigma^2 < 1$.
These images are first augmented by a small random image intensity change of $[-20, 20]$ \si{\DN} and afterwards corrupted with noise generated by the noise model of \cite{konnik2014high}.
Each image patch is corrupted independently with its own set of randomly generated variable metadata.
In this way we generate $\approx$\num{103}k data tuples to train the estimators in a supervised manner.
Our motivation to train on simulated noise only is to cover a large extent of different metadata and to keep the limited real noise data available for model evaluation. 
The network's branches are collectively trained utilizing the mean squared error loss function along with the Adam optimizer \cite{kingma2014adam} and an initial learning rate of $10^{-4}$.
Further implementation details and the training configuration can be found in the code base.

%% file: figNNModel.tex
\begin{figure*}[ht]
    \noindent
    \begin{minipage}{.48\textwidth}
    \centering
            \includegraphics[width=0.85\textwidth,clip]{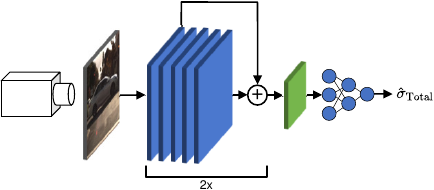}
    \end{minipage}%
    \hfill
    \begin{minipage}{.48\textwidth}
    \centering
        \includegraphics[width=0.9\textwidth,clip]{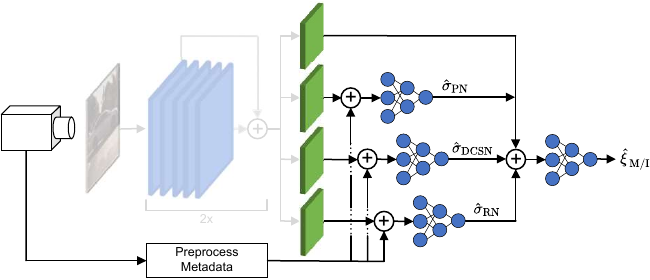}
    \end{minipage}
    %
    \caption{\emph{Noise level estimator vs.~proposed noise source and level estimator}. 
    Left: Customized baseline estimator \baseline{}, which predicts the noise level of the input image's total noise. 
    Right: Proposed noise source estimator that additionally employs camera metadata and predicts the noise levels of four different noise types. \gblue{Architectural changes from the baseline are highlighted.}}
    \label{fig:NNModels}
\end{figure*}

%% file: tabVariableCameraParameterRanges.tex
\begin{table}[t]
    \centering
    \caption{\emph{Camera metadata} used for noise source estimation.  
    We split these into fixed and variable parameters, and consider only variable ones.
    Fixed parameters and all parameter definitions can be found in the supplementary material.
    \label{tab:cameraMetadata}
    }    
    \begin{adjustbox}{valign=t, max width=0.49\linewidth}
        \begin{tabular}{ll}
            \toprule
            Variable Parameter & Value Range \\
            \midrule
            \textbf{Minimal Metadata} & \\
            Camera Gain & [0, 24] \si{\deci\bel}\\
            Exposure Time & [0.001, 0.2] \si{\second}\\
            Sensor Temperature & [0, 80] \si{\celsius}\\
            [2ex]
            \textbf{Full Metadata} & \\
            Dark Signal FoM & [0, 1]\\
            Full Well Capacity & [2, 100] $\times 10^{3}$ $\text{e}^-$\\
            \bottomrule%
        \end{tabular}%
    \end{adjustbox}%
    \hspace{.5ex}
    \begin{adjustbox}{valign=t, max width=0.49\linewidth}%
       \begin{tabular}{ll}
            \toprule
            Variable Parameter & Value Range \\
            \midrule
            \textbf{Full Metadata} & \textbf{(cont.)}\\
            Pixel Clock Rate & [8, 150] $\times 10^{6}$ \si{\hertz}\\
            Sense Node Gain & [1, 5] $\times 10^{-6}$ \si{\milli\meter}\\
            Sense Node -- & \\
            Reset Factor & [0, 1]\\
            Sensor Pixel Size & [0.0009, 0.01] \si{\milli\meter}\\
            Sensor Type & \{CCD, CMOS\}\\
            Thermal Wh. Noise & [1, 60] $\times 10^{-9}$ \si{\hertz}\\
            \bottomrule%
        \end{tabular}%
    \end{adjustbox}    
\end{table}

%% file: experiments.tex
\section{Experiments}
\label{sec:experiments}
We first describe the datasets used and the image noise applied (\cref{subsec:datasets}).
Depending on whether a dataset includes ground truth (GT) labels or not, we conduct either quantitative or qualitative experiments.
Our quantitative experiments comprise performance evaluations on simulated and real-world data (\cref{subsec:quantitativeExperiments}).
In qualitative experiments, we evaluate our methods in real field campaigns and on three use cases of unexpected noise (\cref{subsec:qualitativeExperiments}).
Besides the ability to quantify individual noise sources, we additionally demonstrate the improved total noise estimation performance on the downstream task of real-world image denoising (\cref{sec:denoising}).
Subsequently, we analyze the effects of each camera metadata on noise estimation in comparison to the applied theoretical noise model (\cref{sec:noiseSourceEstimation:sensitivityAnalysis}).
Finally, we provide runtime measurements (\cref{subsec:runtime}).

We compare our proposed estimators against:
($i$) \emph{B+F} \cite{shin2005block}, \baseline{}, \emph{PCA} \cite{Chen15iccv}, and \PGE{} \cite{byun2021fbi} in the case of $\totalNoise$, 
($ii$) \PGE{} for $\photonNoise$, and 
($iii$) noise model predictions from the respective metadata for all individual noise levels $\sigma_{i \in \{\text{PN, DCSN, RN}\}}$.
Note that \PGE{} is only applicable in the quantitative experiments, since it requires (unnoised) GT images in order to calculate $\hat\sigma_{i \in \{\text{PN, Total}\}}$.

All experiments are executed on an Intel Xeon W-2145 CPU and an NVIDIA Quadro RTX 6000 GPU, with the neural networks running on the GPU.
Noise levels are reported as digital numbers in the range $[0,255] \, \si{\DN}$.

\subsection{Datasets}
\label{subsec:datasets}
\input{figDatasets}
\input{figDatasetsGTNoise}
We augment four datasets with ground truth labels and two datasets with pseudo ground truth labels (\cref{fig:datasets}).

\subsubsection{Datasets with Ground Truth} 
\label{subsec:datasetsWithGT}
We employ one simulated and three real-world datasets: $\synth{}$, KITTI\cite{KITTI}, TAMPERE17 \cite{ponomarenko2018blind}, and Udacity \cite{udacity}.
$\synth$ is created with the simulator \cite{Sim} to provide accurate ground truth for noise estimation. 
It comprises \num{1000} images of a village environment acquired from different viewpoints and includes vehicles, such as cars and bikes. 
Similar to our training dataset TAMPERE21 (\cref{subsec:training}), TAMPERE17 provides \num{300} natural images with a controlled noise level of $\sigma^2 < 1$.
From TAMPERE17 we use the grayscale version.
KITTI and Udacity contain images from transportation scenarios.
From KITTI we use the annotated object detection sub-dataset and from Udacity sub-dataset \#2.
We consider only the first 1000 images from both datasets to match the number with $\synth$ and reduce computation time.
\gblue{Depending on the respective image size, one image yields several image patches.}

Note that KITTI and Udacity do not include noise control.
To assess how much noise both original datasets already contain, we apply three state-of-the-art noise level estimators (cf. \cref{sec:relatedWork}).
The results in \cref{fig:datasetsGTNoise} indicate similarly small noise levels for Udacity as for both TAMPERE datasets, but significantly higher noise for KITTI. 
For this reason, we consider Udacity in the main paper and provide KITTI results in the supplementary material.

We corrupt all datasets with \emph{simulated} or \emph{real-world} noise.
In the simulated case, we added noise to the images like our training dataset (cf. \cref{subsec:training}).
In the real-world case, we generated in total \num{12}k RN and DCSN image tuples $(I_{\text{RN}}, I_{\text{DCSN}})$ with about \num{600} different metadata sets from two different camera systems%
\footnote{We investigated several more camera types (e.g., Realsense D435i RGB and Huawei P30), but we reached the point where camera manufacturers would only provide metadata for private usage (i.e., not for publication) behind an NDA. Thus, we cannot include them in this paper.} %
that we abbreviate according to their implemented camera sensors: \ICX285{} \cite{prosilicaCamera} and \EV76C661{} \cite{ximeCamera} (\cref{fig:cameras}).
The first one is considered a scientific-grade CCD and the latter an industrial-grade CMOS camera system.
PN is calculated synthetically as the quantum nature of light determines PN to strictly follow the Poisson distribution.
Details about the real-world noise acquisition, noise post-processings, and used metadata can be found in the supplementary material.
\input{figCameras}

\subsubsection{Datasets without Ground Truth} 
We collect two datasets from field campaigns without ground truth labels: \cellar{} and \parkingLot{}.
Both datasets contain about \num{1000} grayscale images from respective eponymous environments and were recorded with both camera systems. 
We ensured high noise levels by applying the minimum exposure time of $\SI{1}{\milli\second}$ (to capture low but detectable signals), maximum gain of $\SI{24}{\deci\bel}$ (to strongly amplify signal and noise without saturation), and by disabling all image post-processing (that could reduce noise).

We further evaluate the fourth noise type $\restNoise$ as part of these field campaign experiments to demonstrate the detection of unexpected noise during operation time.
Therefore, we split these experiments into two cases: $\restNoise = 0$ and $\restNoise \neq 0$.
The case $\restNoise \neq 0$ is further subdivided into $\restNoise < 0$ and $\restNoise > 0$.
For $\restNoise < 0$, we simulate an additional image noise source by adding randomly generated Gaussian noise $\mathcal{N}(\mu=0, \sigma=\SI{5}{\DN})$ to the images.
For $\restNoise > 0$, we increase the model noise by synthetically doubling the value of the camera metadata \emph{thermal white noise}.
This parameter adjustment can be interpreted as a mis-calibration of the camera sensor's readout profile or a malfunctioned sensor component (e.g., the source follower).
Moreover, we demonstrate the case of doubling the metadata \emph{sensor temperature} in the supplementary material. 

\subsection{Quantitative Experiments}
\label{subsec:quantitativeExperiments}
\textbf{Metrics}. We follow \cite{chen2015efficient} and evaluate our noise source estimators in terms of 
accuracy $\text{Bias} \doteq \left|\mathbb{E}[\sigma - \mathbb{E}(\estimatedNoiseLevel)]\right|$, 
robustness $\text{Std} \doteq (\mathbb{E}[(\estimatedNoiseLevel - \mathbb{E}(\estimatedNoiseLevel))^2])^{1/2}$, 
and overall performance $\text{RMS} \doteq (\text{Bias}^2(\estimatedNoiseLevel) + \text{Std}^2(\estimatedNoiseLevel))^{1/2}$,
where $\estimatedNoiseLevel$ is the estimated noise level and $\sigma$ is the true noise level. 
Smaller RMS, Bias, and Std values indicate better performance.

\subsubsection{Simulated Noise}
\label{subsec:simulatedQuantitativeExperiments}
\input{figRandomSensorNoiseEstimationPlots}
\input{tabRandomSensorSimulatedNoiseResults}
The performance on the synthetically-added noise datasets are summarized in \cref{tab:randomSensorSimulatedEstimation},
while mean noise estimation results on \synth{} are depicted in \cref{fig:randomSensorSimNoisePlots}.

Let us focus on results from \cref{tab:randomSensorSimulatedEstimation} first.
Among the reference methods, we observe that \PGE{} performs worst due to underestimation (cf.~\cref{fig:randomSensorSimNoisePlots}), which agrees with \cite{byun2021fbi}.
We can further see that \baseline{} generally produces better results than \emph{PCA} for all metrics, and both better yield results than \emph{B+F}.
This observation matches \cite{wischow2021camera}.
Considering our proposed methods, we observe that all three estimators accurately and robustly determine $\totalNoise$, where \fullMetadata{} is generally best, and \fullMetadata{} and \withoutMetadata{} perform slightly more robust than \minimalMetadata{} (smaller Std).
In comparison to the reference methods, \fullMetadata{} is on par with \baseline{}.
When it comes to noise source estimation, \fullMetadata{} performs best.
Both accuracy and robustness span sub-intensity levels for all three noise sources in all three datasets.
\withoutMetadata{} and \minimalMetadata{} also accurately quantify the single noise types within sub-intensity levels on average (small bias). 
However, they have worse robustness in all datasets, particularly for DCSN and RN (large Std).
We considered that this might be a problem of insufficient model capacity, but increasing the number of layers and neurons of the FCBs did not produce any change.
We further make two detailed observations: all three methods estimate PN best, and \minimalMetadata{} determines the DCSN amount more robustly than \withoutMetadata{}. 
We attribute the former observation to the strong link between image intensity and PN in the noise model, and the weaker influence of any metadata.
However, only \fullMetadata{} obtains the camera's \emph{full well capacity} parameter, which seems to slightly improve PN estimation.
The more robust DCSN estimation performance of \minimalMetadata{} can be ascribed to its access to \emph{temperature} and \emph{exposure time metadata}, since both have a major impact on thermal noise \cite{konnik2014high}.
The significance of metadata on separating the noise sources is further underpinned by the minor performance on RN estimation (as the minimal metadata only have a minor impact on the noise model) and by the prevailing performance of \fullMetadata{}, which has access to the largest amount of metadata.

\Cref{fig:randomSensorSimNoisePlots} confirms the results of \cref{tab:randomSensorSimulatedEstimation}.
It further indicates the increasing bias for \withoutMetadata{}, the increasing Std (spread of the distributions) for \withoutMetadata{} and \minimalMetadata{} with increasing noise levels $\sigma_{i \in \{\text{Total, PN, DCSN, RN}\}}$.

In summary, only \fullMetadata{} with access to the full set of camera metadata can accurately and robustly quantify the contribution of each noise source.
Although all variants of the proposed method can estimate the total noise level well, the lack of camera metadata for \withoutMetadata{} and \minimalMetadata{} makes it difficult to disambiguate the origin of the noise (i.e., identify the noise sources).

\subsubsection{Real-World Noise}
\label{subsec:mixedQuantitativeExperiments}
\input{figRealWorldNoiseEstimation.tex}
\input{tabSN03SimulatedRealNoiseResults}
Next we discuss the estimation performances of the real-world DCSN/RN produced by ICX285 and EV76C661 using \cref{tab:SN03SimulatedRealNoiseEstimation} and \cref{fig:realNoiseEstimation}.
Note that these two noise-optimized sensors produce lower noise levels compared to our simulated sensors ($\sigma_{i \in \{ \text{DCSN}, \text{RN}\}} \leq \SI{5}{\DN}$).
Both sensors lead to similar results, hence we focus on ICX285 here and consider EV76C661 in the supplementary material.

In contrast to the fully simulated noise experiments (\cref{tab:randomSensorSimulatedEstimation}), the absolute DCSN and RN estimation performances of \withoutMetadata{} and \minimalMetadata{} seem to have improved in \cref{tab:SN03SimulatedRealNoiseEstimation}.
These results should not be overrated due to the generally smaller noise levels and because the errors in the fully simulated cases started to majorly increase for noise levels $\sigma_{i \in \{ \text{DCSN}, \text{RN}\}} \geq \SI{5}{\DN}$.
However, we observe two significant relative performance changes: \fullMetadata{} worsened for RN, and \withoutMetadata{} improved for DCSN/RN.
We attribute the change of both methods in the case of RN to the simulation-reality-gap of the noise model that \withoutMetadata{} coincidentally profits from (cf. \cref{fig:realNoiseEstimation}), because both methods are trained on simulated data only where it has been shown that \fullMetadata{} matches it better (\cref{tab:randomSensorSimulatedEstimation}). 
In the case of DCSN, the better performance of \withoutMetadata{} is misleading, since only \fullMetadata{} seems to approximately fit the ground truth, while the others fail (see \cref{fig:realNoiseEstimation}).
These errors also propagate to the overall noise estimation.
The estimations of the simulated PN have not changed significantly.

In summary, despite the simulation-to-reality gap observed in these experiments 
the access to the full metadata still leads to the best results in terms of noise source quantification,
thus providing evidence for the generalization capabilities of the method.

\subsection{Experiments on Real-world Platforms}
\label{subsec:qualitativeExperiments}
We recorded datasets \cellar{} and \parkingLot{} with camera systems \ICX285{} and \EV76C661{} in field campaigns (\cref{fig:cameras}).
For comparison, we use \emph{B+F}, \baseline{}, and \emph{PCA} in the case of total noise and the noise model predictions with live recorded metadata for the individual noise sources.
Since we observed similar results for both cameras and both datasets, we focus on \ICX285{} and \cellar{} here, and consider the rest in the supplementary material.
We first evaluate the raw dataset (\cref{subsec:soundQualitativeExperiments}) and subsequently test three altered versions with unexpected noise (\cref{subsec:corruptedQualitativeExperiments}).

\subsubsection{Expected Noise ($\modelNoise \approx \imageNoise$)}
\label{subsec:soundQualitativeExperiments}
\input{figICX285CellarParkingLotNoiseSources}
Let us first focus on the noise source identification (top row in \cref{fig:ICX285CellarParkingLotNoiseSources}).
We see that \fullMetadata{} matches the noise model best with $|\hat\sigma_{\text{\fullMetadataTitle{}}} - \hat\sigma_{\text{Reference}}| < \SI{1}{\DN}$ in each noise case, followed by \minimalMetadata{} and \withoutMetadata{}.
These results are generally in accordance to the simulated noise evaluations in \cref{subsec:simulatedQuantitativeExperiments}.
The only significant difference we observe is that \minimalMetadata{} matches the relative value range of the PN noise model curve better than \withoutMetadata{} (i.e., smaller Std).
This can be explained with the \emph{camera gain} parameter that \minimalMetadata{} obtains as one key parameter in the noise model to determine PN (already be indicated on the simulated \ICX285{} in \cref{tab:SN03SimulatedRealNoiseEstimation}).
The residual noise plot depicts only a small mismatch between the noise model and the detected image noise for \fullMetadata{} and \minimalMetadata{}.
Only the nearly constant value of \withoutMetadata{} indicates that it has not learned to detect any residual noises.
From this residual noise estimation of \fullMetadata{} (and later results from \cref{fig:ICX285CellarParkingLotNoiseSources}) we assume for \cellar{} that
\begin{equation}
    \restNoise \approx 0 \stackrel{\eqref{eq:restNoise}}{\implies} \modelNoise \approx \imageNoise.
    \label{eq:uncorruptedNoiseAssumption}
\end{equation}

\input{figICX285CellarParkingLotTotalNoise.tex}
In the total noise inspection (\cref{fig:ICX285CellarParkingLotTotalNoise}), we consider only \fullMetadata{} from our proposed methods to avoid clutter.
We see from both plots that \fullMetadata{} produces similar estimations as the reference methods.
Hence, we consider its results as plausible.

In summary, the results agree with those of the synthetic noise experiments, meaning that our model is applicable to an actual real-world robotic platform. 
The more metadata are available, the better the noise source estimation of all noise types (with \fullMetadata{} as the best method).

\subsubsection{Unexpected Noise ($\modelNoise \neq \imageNoise$)}
\label{subsec:corruptedQualitativeExperiments}
Here we evaluate three scenarios where we synthetically increase image noise or model noise to reach $\restNoise < 0$ (i.e., $\modelNoise < \imageNoise$) or $\restNoise > 0$ (i.e., $\modelNoise > \imageNoise$), respectively.
We investigate these scenarios on the basis of the raw \cellar{} dataset for that we assume that the applied noise model follows the actual image noise (i.e., \eqref{eq:uncorruptedNoiseAssumption}: $\restNoise \approx 0$).

\textbf{One scenario of the form $\modelNoise < \imageNoise$: } In our first scenario we increase the image noise by adding randomly sampled Gaussian noise from $\mathcal{N}(\mu = 0, \sigma_{\mathcal{N}} = \SI{5}{\DN})$ to the raw \cellar{} images. 
Note that this Gaussian noise is statistically independent from the other image noise sources and so its noise level adds in quadrature to the new total image noise level $\sigma_{\text{Image+}\mathcal{N}}$ (cf.~\eqref{eq:totalNoise}).
We calculated the resulting ground truth $\restNoise$ as
\begin{equation}
        \restNoise  \stackrel{\eqref{eq:restNoise}}{=} \modelNoise - \sigma_{\text{Image+}\mathcal{N}}
                    \stackrel{\eqref{eq:uncorruptedNoiseAssumption}}{\approx} \modelNoise - \sqrt{\modelNoise^2 + \sigma_{\mathcal{N}}^2}.
    \label{eq:restNoiseGreaterZero}
\end{equation}

The middle row of \cref{fig:ICX285CellarParkingLotNoiseSources} illustrates the results.
We expect only a reduction of $\restNoise$ and unchanged values otherwise, with respect to the first row.
It can be seen that only \fullMetadata{} captures the unexpected noise (note the initial error of $\approx \SI{0.5}{\DN}$ is propagated), whereas \withoutMetadata{} remains unchanged (cf. \cref{subsec:soundQualitativeExperiments}) and \minimalMetadata{} incorrectly estimate increased values. 
Furthermore, \withoutMetadata{} and \minimalMetadata{} split $\sigma_{\mathcal{N}}$ among the other noise sources (especially \minimalMetadata{} increases $\readoutNoiseEstimate$ significantly).
Only \fullMetadata{} maintains its noise source estimated values.

\textbf{Two scenarios of the form $\modelNoise > \imageNoise$: } 
In this second test we increase the model noise by doubling the metadata \emph{thermal white noise}.
This parameter only affects \fullMetadata{}.
The new ground truth noise levels are calculated using the noise model.
(In a third test, we prepared an example with a doubled metadata \emph{sensor temperature}, however, without new findings.
Thus, it is treated in the suppl.~mat.)

The results are shown in the bottom row of \cref{fig:ICX285CellarParkingLotNoiseSources}.
In this case, we expect an increasing $\readoutNoiseEstimate$ in accordance to the increased \emph{thermal white noise}, 
an increasing $\restNoiseEstimation$ (which indicates the unexpected higher model noise) and unchanged values otherwise.
We can see that \fullMetadata{} meets these expectations (note the initial propagated error here as well).

We conclude that unexpected noise in either images or from metadata could only be reliably quantified with the full set of variable camera metadata.

\subsection{Experiments on Real-World Image Denoising}
\label{sec:denoising}
Our proposed noise source estimator is able to improve total image noise estimation (see Tables \ref{tab:randomSensorSimulatedEstimation} and \ref{tab:SN03SimulatedRealNoiseEstimation}), thereby offering potential advantages for downstream vision tasks.
Although we use the symptom-fighting denoising as a rationale for noise source estimation, denoising is the most studied downstream application for noise level estimation and therefore best suited to assess the effects of estimating total noise more accurately.

We investigate the effect of more accurate total noise level estimation on denoising on the example of two traditional denoisers that input expected noise levels (BM3D \cite{dabov2007image} and NLM \cite{buades2005non}) and compare results to two state-of-the-art learning-based denoisers (FBI-Denoiser \cite{byun2021fbi} and Blind2Unblind \cite{wang2022blind2unblind}).
BM3D and NLM both assume Gaussian noise, the FBI denoiser internally uses PGE-Net for Poisson-Gaussian noise estimation, and Blind2Unblind does not explicit assume a noise distribution.
We apply all denoisers with default parameter values and pre-trained weights provided by the respective authors (we select respective weights for real-noise images that lead to the best results for our datasets, i.e., ``DND''-weights for FBI-Denoiser and ``raw RGB''-weights for Blind2Unblind).
For a fair comparison, all denoisers are applied to whole images.
Denoising results are compared using peak signal-to-noise ratio (PSNR [dB]) and structural similarity index measure (SSIM) \cite{wang2004image}.
\input{tabDenoisingICX285}
\input{tabDenoisingEV76C661}

Table~\ref{tab:noiseSourceEstimation:denoising} presents quantitative results using the \ICX285{} camera.
For the \cellar{} scene, \fullMetadata{} + BM3D leads to the best scores in all cases, followed by \minimalMetadata{} + BM3D for cases of higher noise (less images used for averaging), and \baseline{} + BM3D for lower noise cases (more images for averaging).
We observe similar results in combination with the NLM denoiser.
This is in accordance with the results from \cref{tab:randomSensorSimulatedEstimation} that \baseline{} and \fullMetadata{} perform best and in accordance with \cref{tab:SN03SimulatedRealNoiseEstimation} that \minimalMetadata{} is not far off, but it counteracts the non-intuitive results from \cref{tab:SN03SimulatedRealNoiseEstimation} that \withoutMetadata{} occasionally leads to more accurate total noise estimations.
The denoising results rather underpin the intuition that the more metadata available for the noise source estimators, the better the total noise estimation.
The learning-based denoisers perform less accurate than BM3D, which differs from the results reported in their respective original studies, as these denoisers were neither trained on large and diverse real-world datasets (with the weights we employ) nor fine-tuned to our datasets.
The performance gap between the traditional and the learning-based denoisers increase with decreasing noise level.

We note similar results for the \parkingLot{} dataset with the difference that Blind2Unblind and \minimalMetadata{} both score best in the two highest noise cases.
However, the better performance of \minimalMetadata{} compared to \fullMetadata{} may be specific to BM3D, as \fullMetadata{} is relatively more accurate when combined with NLM.
Experiments on the \EV76C661{} camera yield comparable findings (\cref{tab:noiseSourceEstimation:denoising:EV76C661}).

Figure~\ref{fig:realWorldDenoisingResults} illustrates qualitative results on the example of both scenes recorded with the \ICX285{} camera and a medium noise level (four averaged images each).
It can be seen that the FullMeta+BM3D combination is the best at visually removing noise while preserving image detail, closely followed by DRNE+BM3D (e.g., FullMeta+BM3D restores the edges of the shadows less pixelated in the first row of \cref{fig:realWorldDenoisingResults}).
In contrast, FBI-Denoiser and Blind2Unblind visually remove noise the best, but smooth the entire image (both methods, see especially bottom rows in \cref{fig:realWorldDenoisingResults}) and introduce square artifacts (Blind2Unblind).
NLM tends to generally retain noise at edges (e.g., around the door handle in the first row and around the silver frame in the third row of \cref{fig:realWorldDenoisingResults}).
\input{figDenoisingResultImages}

In summary, \fullMetadata{} in combination with the traditional BM3D denoiser leads to the best denoising results in most cases.
This supports previous findings that \fullMetadata{} generally estimates total noise levels the best and thus that noise estimation can benefit from camera metadata.

\subsection{Metadata Sensitivity Analysis}
\label{sec:noiseSourceEstimation:sensitivityAnalysis}
\input{tabMetadataSensitivityAnalysis}
In this section, we investigate the individual influence of camera metadata on total noise level estimations.
Building upon previous results, we focus only on \fullMetadata{} and compare it to the theoretical noise model.

We conduct a black box analysis by uniformly sampling different input parameter values from the respective parameter ranges for both approaches and observing respective outputs (i.e., the total noise estimations).
One input parameter is sampled at a time and other parameters are fixed to their respective maximum value (to aim for sufficiently high noise levels).
Note that the only image feature the noise model depends on is the image intensity, while \fullMetadata{} may have learned to employ more image features (e.g., image noise).
As we focus on the influence of camera metadata only, we only input uncorrupted homogeneous images with uniform intensities.

In the case of \fullMetadata{}, we further omit the residual noise estimation $\restNoise$ to calculate the total noise level, as a mismatch between image noise and camera metadata is expected.
Finally, we compare the estimated noise levels of both models to quantify the impact of each metadata and whether \fullMetadata{} has learned the theoretical model (we consider deviations in $[1,2] \, \si{\DN}$ as minor but worth noting and those larger than $\SI{2}{\DN}$ as significant).

We first examine the effect of each parameter on the estimated noise level according to the theoretical noise model (top part of \cref{tab:cameraMetadataSensitivityAnalysisEstimator}).
For each row, the bigger the difference between estimated noise levels in the “Min” and “Max” columns, the more important the parameter.
The table shows that the full well capacity is most important because it determines the photon shot noise in the noise model, followed by the camera gain that amplifies noise. 
The pixel clock rate, thermal white noise, and sense node reset factor, which all contribute to readout noise, have a negligible effect on the estimated total noise level.
Note that these three parameters are only insignificant for the considered set of fixed parameters in our experiments (see supplementary material). 
To illustrate, for instance, the impact of the pixel clock rate on the total noise level is closely tied to the correlated double sampling dominant time constant \cite{konnik2014high}.

Comparing the noise level of the noise model with the estimations from the \fullMetadata{} model (bottom part in \cref{tab:cameraMetadataSensitivityAnalysisEstimator}), \fullMetadata{} has generally learned the relations between input parameters and the noise levels.
Yet, there are specific cases where deviations can be observed (indicated by colored values).
Let us consider the two severe model deviations first (red values).
The first differences are the estimated noise levels for minimum and maximum mean image intensities.
This corresponds to the reduced noise estimation accuracy that we observed for under- and overexposed images (see supplementary material).
The second deviation can be observed for full well capacities $\leq 24$k electrons.
The corresponding noise levels are most different from the others learned by \fullMetadata{} (that range between about $[0,13] \, \si{\DN}$). 
The farther the noise values depart from this range, the larger the observed model deviation.
This indicates that these noise levels are underrepresented in the training data.
Minor deviations from the noise model (orange values) are limited to small respective parameter values, with the exception of the sensor temperature. 
However, we do not see a specific pattern in these deviations; they are mostly slightly above $\SI{1}{\DN}$. 

In conclusion, the \fullMetadata{} model learned to capture the theoretical camera metadata relations, with notable exceptions for low and high exposed images, and large noise levels resulting from camera full well capacities $\leq 24$k electrons.
The full well capacity and the camera gain could be identified as the most significant camera metadata, while pixel clock rate, sense nose reset factor, and thermal white noise could be neglected.

\subsection{Computational Cost}
\label{subsec:runtime}
The computation time is determined by averaging the noise estimation inference times for \num{13.5}k Udacity image patches (i.e., 100 Udacity images).
We repeated the measurements 5 times and took the average to eliminate the influence of background processes and caching.
We measured the following average runtimes per image patch: $\SI{1.4}{\milli\second}$ (\withoutMetadata{}), $\SI{1.3}{\milli\second}$ (\minimalMetadata{}), $\SI{1.3}{\milli\second}$ (\fullMetadata{}), $\SI{1.2}{\milli\second}$ (\baseline{}), $\SI{0.1}{\milli\second}$ (\PGE{}), $\SI{9.8}{\milli\second}$ (\PCA{}), and $\SI{13.2}{\milli\second}$ (\B+F{}). 
Note that \PGE{} is faster because it processes a whole image at once, but it does not estimate as many noise sources nor is as accurate as the proposed method(s).

%% file: figDatasets.tex
\begin{figure}[t]
\centering
\includegraphics[width=.99\columnwidth, trim=16 16 16 16, clip]{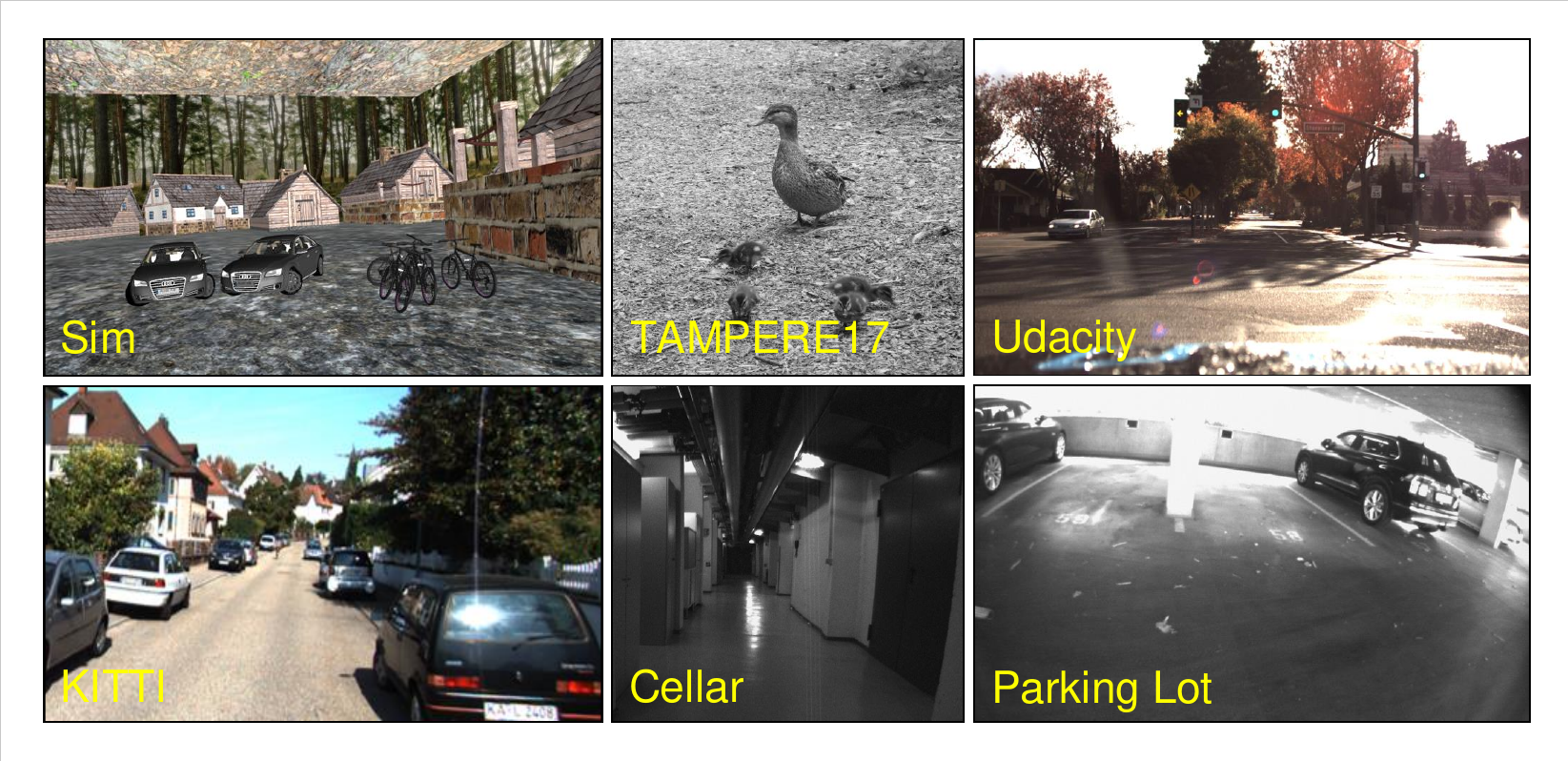}
\caption{\emph{Datasets}. 
Exemplary image snippets from
\synth{} ($896 \times 768 \,\si{\px}$), 
TAMPERE17 ($512 \times 512 \,\si{\px}$), 
Udacity ($1920 \times 1200 \,\si{\px}$), 
KITTI ($1242 \times 375 \,\si{\px}$),
\cellar{} and \parkingLot{} (\ICX285: $1360 \times 1024 \,\si{\px}$, \EV76C661: $1280 \times 1024 \,\si{\px}$).
}
\label{fig:datasets}
\end{figure}

%% file: figDatasetsGTNoise.tex
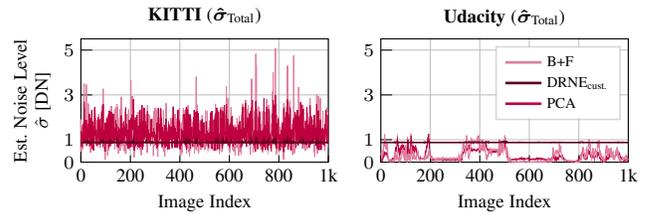
\begin{figure}
    \centering
    \tikzsetnextfilename{FigdatasetsGTNoise}
    \begingroup
        \pgfplotsset{every axis/.style={scale=0.48}}
        \input{plotDatasetsGTNoise.tex}
    \endgroup
    \caption{\emph{Noise estimation of uncorrupted KITTI and Udacity datasets.} 
    The reference methods estimate significant noise in KITTI images ($\hat\sigma \leq 5$) 
    and low noise in Udacity data ($\hat\sigma \leq 1.25$).
    }
    \label{fig:datasetsGTNoise}
\end{figure}

%% file: plotDatasetsGTNoise.tex
\pgfplotstableread[col sep = comma]{dataKITTIGT1Branch.csv}\loadedtableA
\pgfplotstableread[col sep = comma]{dataKITTIGTPCA.csv}\loadedtableB
\pgfplotstableread[col sep = comma]{dataKITTIGTB+F.csv}\loadedtableC

\pgfplotstableread[col sep = comma]{dataUdacityGT1Branch.csv}\loadedtableD
\pgfplotstableread[col sep = comma]{dataUdacityGTPCA.csv}\loadedtableE
\pgfplotstableread[col sep = comma]{dataUdacityGTB+F.csv}\loadedtableF

\begin{tikzpicture}
    
    {\scriptsize
    \begin{axis}[%
            name=axis1,
            clip mode=individual,
            yscale=0.6,
            tick pos=left,
            xmajorgrids,
            xmin=0, xmax=10,
            xtick style={color=black},
            xtick={0, 2, 4, 6, 8, 10},
            xticklabels={0, 200, 400, 600, 800, 1k},
            ymajorgrids,
            ymin=0, ymax=5.5,
            ytick style={color=black},
            ytick={0, 1, 3, 5},
            xlabel near ticks,
            ylabel near ticks,
            xlabel={Image Index},
            xlabel style={yshift=-4pt},
            ylabel style={align=center}, 
            ylabel={Est. Noise Level \\ $\hat{\sigma}$ [DN]},
            title={\textbf{KITTI} ($\bm{\hat\sigma_\text{Total}}$)},
            title style={yshift=16pt},
    ]
        
        \addplot[color=purple!50!white] table [x=3, y=7] {\loadedtableC};
        \addplot[color=purple] table [x=3, y=7] {\loadedtableB};
        \addplot[color=purple!50!black, semithick] table [x=3, y=7] {\loadedtableA};
    \end{axis}
    \begin{axis}[%
            name=axis2,
            at=(axis1.right of south east), 
            clip mode=individual,
            tick pos=left,
            xshift=14pt,
            xmajorgrids,
            xmin=0, xmax=10,
            xtick style={color=black},
            xtick={0, 2, 4, 6, 8, 10},
            xticklabels={0, 200, 400, 600, 800, 1k},
            ymajorgrids,
            ymin=0, ymax=5.5,
            ytick style={color=black},
            ytick={0, 1, 3, 5},
            xlabel near ticks,
            ylabel near ticks,
            xlabel={Image Index},
            xlabel style={yshift=-4pt},
            yscale=0.6,
            title={\textbf{Udacity} ($\bm{\hat\sigma_\text{Total}}$)},
            title style={yshift=15pt},
            legend cell align={left},
            legend image post style={scale=0.8},
            legend columns=1,
            legend style={
              nodes={scale=0.8, transform shape},
              text opacity=1,
              at={(0.97, 1.57)},
              anchor=north east,
              font={\scriptsize\color{black}},
              line width=0.6pt,
              draw=lightgray,
              draw opacity=0.6%
            },
            legend entries={%
                B+F,
                \baselineTitle,
                PCA,
                }
    ]
        \addlegendimage{color=purple!50!white, line width=1.0pt, draw opacity=1.0}
        \addlegendimage{color=purple!50!black, no markers, line width=1.0pt, draw opacity=1.0}
        \addlegendimage{color=purple, line width=1.0pt, draw opacity=1.0}
        
        \addplot[color=purple] table [x=3, y=7] {\loadedtableE};
        \addplot[color=purple!50!white] table [x=3, y=7] {\loadedtableF};
        \addplot[color=purple!50!black, semithick] table [x=3, y=7] {\loadedtableD};
    \end{axis}
    }
\end{tikzpicture}

%% file: figCameras.tex
\begin{figure}[t]
\centering
\includegraphics[width=1.0\columnwidth, trim=16 16 20 16, clip]{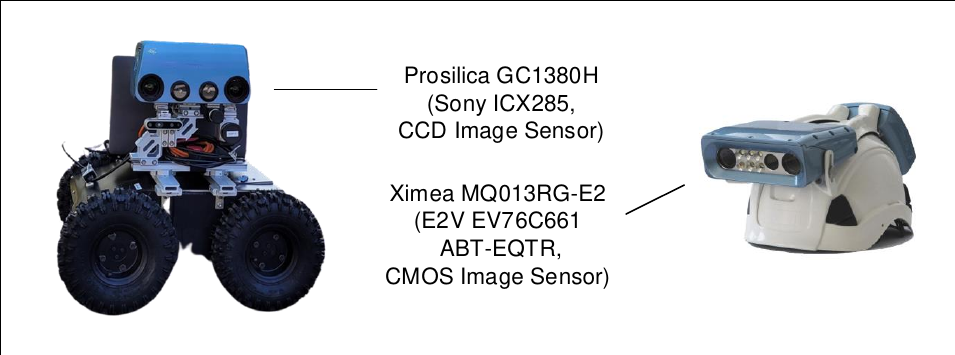}
\caption{\emph{Camera systems.} \ICX285 is attached on an autonomous robotic platform and \EV76C661 on an inspection helmet.
}
\label{fig:cameras}
\end{figure}

%% file: figRandomSensorNoiseEstimationPlots.tex
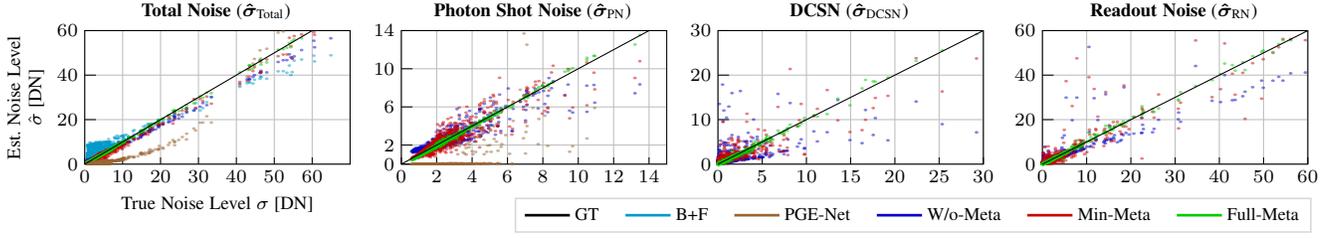
\begin{figure*}[ht]
    \tikzsetnextfilename{FigrandomSensorSimNoisePlots}
    \begingroup
        \pgfplotsset{every axis/.style={scale=0.52}}
        \input{plotRandomSensorNoise.tex}
    \endgroup
    \caption{\emph{Noise source estimation on synthetic noise  (dataset: \synth{}, camera: random)}. 
    Each dot represents the mean noise estimation of one image. 
    The plots of \baseline{} and \emph{PCA} are omitted in the case of $\totalNoiseEstimate$ due to a strong similarity with the other plots (to avoid clutter).}
    \label{fig:randomSensorSimNoisePlots}
\end{figure*}

%% file: plotRandomSensorNoise.tex
\pgfplotstableread[col sep = comma]{datarandomSensorSim1Branch.csv}\loadedtableA
\pgfplotstableread[col sep = semicolon]{datarandomSensorSim4Branch.csv}\loadedtableB
\pgfplotstableread[col sep = semicolon]{datarandomSensorSim4BranchConf.csv}\loadedtableC
\pgfplotstableread[col sep = semicolon]{datarandomSensorSim4BranchAll.csv}\loadedtableD
\pgfplotstableread[col sep = semicolon]{datarandomSensorSimPGE-Net.csv}\loadedtableE
\pgfplotstableread[col sep = comma]{datarandomSensorSimB+F.csv}\loadedtableF
\pgfplotstableread[col sep = comma]{datarandomSensorSimPCA.csv}\loadedtableG
\begin{tikzpicture}
{\scriptsize
\tikzset{mark options={mark size=0.5, opacity=0.5}}

\begin{axis}[%
        name=axis1,
        clip mode=individual,
        tick pos=left,
        xmajorgrids,
        xmin=0, xmax=70,
        xtick style={color=black},
        xtick={0, 10, 20, 30, 40, 50, 60},
        ymajorgrids,
        ymin=0, ymax=60,
        ytick style={color=black},
        ytick={0, 20, 40, 60},
        xlabel near ticks,
        ylabel near ticks,
        xlabel={True Noise Level $\sigma$ [DN]},
        xlabel style={yshift=-4pt},
        ylabel style={align=center}, 
        ylabel={Est. Noise Level \\ $\hat{\sigma}$ [DN]},
        title={\textbf{Total Noise} ($\bm{\hat\sigma_\text{Total}}$)},
        title style={yshift=16pt},
        yscale=0.6,
        xscale=0.99
    ]
    
    \addplot[color=cyan!80!black, only marks] table [x=3, y=7] {\loadedtableF};
    \addplot[color=blue!80!black, only marks] table [x=4, y=9] {\loadedtableB};
    \addplot[color=red!80!black, only marks] table [x=4, y=9] {\loadedtableC};
    \addplot[color=green!80!black, only marks] table [x=4, y=9] {\loadedtableD};
    \addplot[color=brown!80!black, only marks] table [x=4, y=9] {\loadedtableE};
    \addplot[color=black] coordinates {(0,0) (70,70)};
\end{axis}
  
\begin{axis}[%
    name=axis2,
    at=(axis1.right of south east), 
    anchor=left of south west,
    xshift=2ex, 
    clip mode=individual,
    tick pos=left,
    xmajorgrids,
    xmin=0, xmax=15,
    xtick style={color=black},
    xtick={0, 2, 4, 6, 8, 10, 12, 14},
    ymajorgrids,
    ymin=0, ymax=14,
    ytick style={color=black},
    ytick={0, 2, 6, 10, 14},
    xlabel near ticks,
    ylabel near ticks,
    title={\textbf{Photon Shot Noise} ($\bm{\hat\sigma_\text{PN}}$)},
    title style={yshift=16pt},
    yscale=0.6,
    xscale=0.99
]
    
    \addplot[color=blue!80!black, only marks] table [x=0, y=5] {\loadedtableB};
    \addplot[color=red!80!black, only marks] table [x=0, y=5] {\loadedtableC};
    \addplot[color=green!80!black, only marks] table [x=0, y=5] {\loadedtableD};
    \addplot[color=brown!80!black, only marks] table [x=0, y=5] {\loadedtableE};
    \addplot[color=black] coordinates {(0,0) (15,15)};
\end{axis}

\begin{axis}[%
    name=axis3,
    at=(axis2.right of south east), 
    anchor=left of south west,
    xshift=2ex,
    clip mode=individual,
    tick pos=left,
    xmajorgrids,
    xmin=0, xmax=30,
    xtick style={color=black},
    xtick={0, 5, 10, 15, 20, 25, 30},
    ymajorgrids,
    ymin=0, ymax=30,
    ytick style={color=black},
    ytick={0, 10, 20, 30},
    xlabel near ticks,
    ylabel near ticks,
    title={\textbf{DCSN} ($\bm{\hat\sigma_\text{DCSN}}$)},
    title style={yshift=16pt},
    yscale=0.6,
    xscale=0.99
]

    \addplot[color=blue!80!black, only marks] table [x=1, y=6] {\loadedtableB};
    \addplot[color=red!80!black, only marks] table [x=1, y=6] {\loadedtableC};
    \addplot[color=green!80!black, only marks] table [x=1, y=6] {\loadedtableD};
    \addplot[color=black] coordinates {(0,0) (30, 30)};
\end{axis}
    
\begin{axis}[%
        name=axis4,
        at=(axis3.right of south east), 
        anchor=left of south west,
        xshift=1.0ex,
        clip mode=individual,
        tick pos=left,
        xmajorgrids,
        xmin=0, xmax=60,
        xtick style={color=black},
        xtick={0, 10, 20, 30, 40, 50, 60},
        ymajorgrids,
        ymin=0, ymax=60,
        ytick style={color=black},
        ytick={0, 20, 40, 60},
        xlabel near ticks,
        ylabel near ticks,
        title={\textbf{Readout Noise} ($\bm{\hat\sigma_\text{RN}}$)},
        title style={yshift=16pt},
        yscale=0.6,
        xscale=0.99,
        legend cell align={left},
        legend image post style={scale=1.0},
        legend columns=6,
        legend style={
          nodes={scale=0.75, transform shape},
          text opacity=1,
          at={(0, 0)},
          anchor=north east,
          /tikz/every even column/.append style={column sep=8pt},
          font={\small\color{black}},
          line width=0.6pt,
          draw=lightgray,
          draw opacity=0.6,
          fill=none
          },
        legend entries={GT,
                B+F,
                \PGETitle,
                \withoutMetadataTitle,
                \minimalMetadataTitle,
                \fullMetadataTitle},
        legend to name=UniversalLegendRandomSensorSimNoise
    ]

    \addlegendimage{color=black, no markers, line width=1.0pt, draw opacity=1.0}
    \addlegendimage{color=cyan!80!black, line width=1.0pt, draw opacity=1.0}
    \addlegendimage{color=brown!80!black, line width=1.0pt, draw opacity=1.0}
    \addlegendimage{color=blue!80!black, line width=1.0pt, draw opacity=1.0}
    \addlegendimage{color=red!80!black, line width=1.0pt, draw opacity=1.0}
    \addlegendimage{color=green!80!black, line width=1.0pt, draw opacity=1.0}

    \addplot[color=blue!80!black, only marks] table [x=2, y=7] {\loadedtableB};
    \addplot[color=red!80!black, only marks] table [x=2, y=7] {\loadedtableC};
    \addplot[color=green!80!black, only marks] table [x=2, y=7] {\loadedtableD};
    \addplot[color=black] coordinates {(0,0) (60, 60)};
\end{axis}
}
\end{tikzpicture}

\tikzexternaldisable
    \hspace*{191.5pt}
    \raisebox{6pt}[0cm][0cm]{\ref*{UniversalLegendRandomSensorSimNoise}}
\vspace{-6pt}
\tikzexternalenable

%% file: tabRandomSensorSimulatedNoiseResults.tex
\begin{table*}[t]
\caption{\emph{Noise source estimation on synthetically corrupted datasets.} The simulated noise is generated on the basis of randomly simulated camera sensors. The best results per method and dataset are highlighted in bold. \label{tab:randomSensorSimulatedEstimation}}
    \centering
    \begin{adjustbox}{max width=1.00\linewidth}
        \begin{tabular}{llcccccccccccc}
            \toprule
            &&\multicolumn{3}{c}{Photon Shot Noise}&\multicolumn{3}{c}{DCSN}&\multicolumn{3}{c}{Readout Noise} &\multicolumn{3}{c}{Total Noise}%
            \\
            \cmidrule(r){3-5}\cmidrule(r){6-8}\cmidrule(r){9-11}\cmidrule(r){12-14}
            & & Bias & Std & RMS & Bias & Std & RMS & Bias & Std & RMS & Bias & Std & RMS \\
            \midrule
            \parbox[t]{3.0mm}{\multirow{7}{*}{\rotatebox[origin=c]{90}{Sim}}}
                 &B+F \cite{shin2005block}%
                        & - & - & - %
                        & - & - & - %
                        & - & - & - %
                        & 2.51 & 3.00 & 3.91 \\
                 &\baselineTitle%
                        & - & - & - %
                        & - & - & - %
                        & - & - & - %
                        & \textbf{0.07} & \textbf{0.23} & \textbf{0.23}\\
                &PCA \cite{Chen15iccv}%
                        & - & - & - %
                        & - & - & - %
                        & - & - & - %
                        & 0.75 & 1.07 & 1.30 \\
                &\PGETitle \cite{byun2021fbi}%
                        & 1.74 & 3.02 & 3.49 %
                        & - & - & - %
                        & - & - & - %
                        & 3.23 & 4.36 & 5.43\\
                &\withoutMetadataTitle%
                        & \textbf{0.01} & 0.75 & 0.75 %
                        & 0.35 & 4.23 & 4.24 %
                        & 0.35 & 3.40 & 3.42 %
                        & 0.50 & 1.22 & 1.32\\
                &\minimalMetadataTitle%
                        & 0.05 & 0.75 & 0.76 %
                        & 0.13 & 2.82 & 2.83 %
                        & 0.13 & 3.38 & 3.39 %
                        & 0.47 & 0.97 & 1.08\\
                &\fullMetadataTitle%
                        & 0.09 & \textbf{0.07} & \textbf{0.09} %
                        & \textbf{0.07} & \textbf{0.34} & \textbf{0.35} %
                        & \textbf{0.09} & \textbf{0.46} & \textbf{0.47} %
                        & 0.16 & 0.29 & 0.33\\
            [0.9ex]%
            \parbox[t]{3.0mm}{\multirow{7}{*}{\rotatebox[origin=c]{90}{Tamp.17}}}
                &B+F \cite{shin2005block}%
                        & - & - & - %
                        & - & - & - %
                        & - & - & - %
                        & 2.22 & 4.19 & 4.74 \\
                &\baselineTitle%
                        & - & - & - %
                        & - & - & - %
                        & - & - & - %
                        & 0.21 & 0.44 & 0.49\\
                &PCA \cite{Chen15iccv}%
                        & - & - & - %
                        & - & - & - %
                        & - & - & - %
                        & 2.81 & 3.04 & 4.14 \\
                &\PGETitle%
                        & 2.06 & 1.72 & 2.68 %
                        & - & - & - %
                        & - & - & - %
                        & 3.15 & 3.34 & 4.59\\
                &\withoutMetadataTitle%
                        & 0.16 & 0.84 & 0.85 %
                        & \textbf{0.07} & 3.11 & 3.11 %
                        & \textbf{0.02} & 3.04 & 3.04 %
                        & \textbf{0.02} & 1.18 & 1.18\\
                &\minimalMetadataTitle%
                        & \textbf{0.09} & 0.82 & 0.83 %
                        & 0.21 & 2.01 & 2.02 %
                        & 0.39 & 3.73 & 3.75 %
                        & \textbf{0.02} & 1.05 & 1.05\\
                &\fullMetadataTitle%
                        & 0.10 & \textbf{0.13} & \textbf{0.16} %
                        & 0.09 & \textbf{0.29} & \textbf{0.30} %
                        & 0.17 & \textbf{0.37} & \textbf{0.41} %
                        & 0.05 & \textbf{0.43} & \textbf{0.43}\\
            [0.9ex]%
            \parbox[t]{3.0mm}{\multirow{7}{*}{\rotatebox[origin=c]{90}{Udacity}}} 
                &B+F \cite{shin2005block}%
                        & - & - & - %
                        & - & - & - %
                        & - & - & - %
                        & 1.09 & 2.19 & 2.44 \\
               &\baselineTitle%
                        & - & - & - %
                        & - & - & - %
                        & - & - & - %
                        & 0.24 & 0.50 & 0.54\\
                &PCA \cite{Chen15iccv}%
                        & - & - & - %
                        & - & - & - %
                        & - & - & - %
                        & 0.70 & 0.93 & 1.17 \\
                &\PGETitle%
                        & 1.58 & 2.05 & 2.59 %
                        & - & - & - %
                        & - & - & - %
                        & 3.04 & 3.70 & 4.79\\
                &\withoutMetadataTitle%
                        & \textbf{0.05} & 0.54 & 0.54 %
                        & 0.28 & 3.31 & 3.33 %
                        & 0.45 & 2.54 & 2.58 %
                        & 0.44 & 1.39 & 1.46\\
                &\minimalMetadataTitle%
                        & 0.19 & 0.66 & 0.68 %
                        & \textbf{0.03} & 2.21 & 2.21 %
                        & 0.27 & 2.38 & 2.40 %
                        & \textbf{0.11} & 0.88 & 0.89\\
                &\fullMetadataTitle%
                        & 0.06 & \textbf{0.14} & \textbf{0.15} %
                        & 0.04 & \textbf{0.30} & \textbf{0.30} %
                        & \textbf{0.10} & \textbf{0.44} & \textbf{0.45} %
                        & 0.14 & \textbf{0.42} & \textbf{0.45}\\
            \bottomrule
        \end{tabular}%
    \end{adjustbox}
\end{table*}

%% file: figRealWorldNoiseEstimation.tex
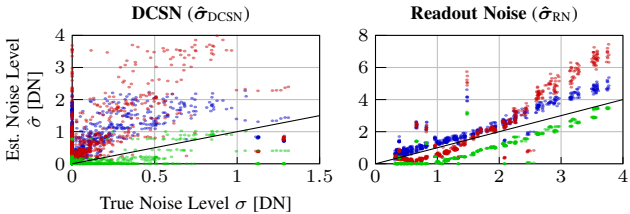
\begin{figure}[t]
    \centering
    \tikzsetnextfilename{FigrealNoiseEstimation}
    \begingroup
        \pgfplotsset{every axis/.style={scale=0.50}}
        \pgfplotsset{cycle list/Set1, cycle multiindex* list={
                mark list*\nextlist
                Set1\nextlist
            }
        }
        \input{plotRealNoise.tex}
    \endgroup
    \caption{\emph{Noise source estimation on real-world noise (dataset: \synth{}, camera: \ICX285{}).} 
    Compare to \cref{fig:randomSensorSimNoisePlots}.}
    \label{fig:realNoiseEstimation}
\end{figure}

%% file: plotRealNoise.tex
\pgfplotstableread[col sep = comma]{dataICX285sim4Branch.csv}\loadedtableA
\pgfplotstableread[col sep = comma]{dataICX285sim4BranchConf.csv}\loadedtableB
\pgfplotstableread[col sep = comma]{dataICX285sim4BranchAll.csv}\loadedtableC
\begin{tikzpicture}
{\scriptsize
\tikzset{mark options={mark size=0.5, opacity=0.5}}

\begin{axis}[%
    name=axis1,
    clip mode=individual,
    tick pos=left,
    xmajorgrids,
    xmin=0, xmax=1.5,
    xtick style={color=black},
    ymajorgrids,
    ymin=0, ymax=4,
    ytick style={color=black},
    ytick={0, 1, 2, 3, 4},
    xlabel near ticks,
    ylabel near ticks,
    xlabel={True Noise Level $\sigma$ [DN]},
    xlabel style={yshift=-4pt},
    ylabel style={align=center}, 
    ylabel={Est. Noise Level \\ $\hat{\sigma}$ [DN]},
    title={\textbf{DCSN} ($\bm{\hat\sigma_\text{DCSN}}$)},
    title style={yshift=16pt},
    yscale=0.6,
    xscale=0.96
]

    \addplot[color=blue!80!black, only marks] table [x=1, y=5] {\loadedtableA};
    \addplot[color=red!80!black, only marks] table [x=1, y=5] {\loadedtableB};
    \addplot[color=green!80!black, only marks] table [x=1, y=5] {\loadedtableC};
    \addplot[color=black] coordinates {(0,0) (30, 30)};
\end{axis}
    
\begin{axis}[%
        name=axis2,
        at=(axis1.right of south east), 
        anchor=left of south west,
        xshift=1.5ex,
        clip mode=individual,
        tick pos=left,
        xmajorgrids,
        xmin=0, xmax=4,
        xtick style={color=black},
        ymajorgrids,
        ymin=0, ymax=8,
        ytick style={color=black},
        ytick={0, 2, 4, 6, 8},
        xlabel near ticks,
        ylabel near ticks,
        title={\textbf{Readout Noise} ($\bm{\hat\sigma_\text{RN}}$)},
        title style={yshift=16pt},
        yscale=0.6,
        xscale=0.96
    ]

    \addplot[color=blue!80!black, only marks] table [x=2, y=6] {\loadedtableA};
    \addplot[color=red!80!black, only marks] table [x=2, y=6] {\loadedtableB};
    \addplot[color=green!80!black, only marks] table [x=2, y=6] {\loadedtableC};
    \addplot[color=black] coordinates {(0,0) (60, 60)};
\end{axis}
}
\end{tikzpicture}

%% file: tabSN03SimulatedRealNoiseResults.tex
\begin{table*}[t]
\caption{\emph{Noise source estimation on real-world noise extracted from a Sony ICX285 CCD Sensor.} 
    DCSN and RN with corresponding metadata were recorded from the camera. 
    PN was generated synthetically using the real metadata. 
    The best results per method and dataset are highlighted in bold.\label{tab:SN03SimulatedRealNoiseEstimation}}
    \centering
    \begin{adjustbox}{max width=1.0\linewidth}
        \begin{tabular}{llcccccccccccc}
            \toprule
            &&\multicolumn{3}{c}{Photon Shot Noise}&\multicolumn{3}{c}{DCSN}&\multicolumn{3}{c}{Readout Noise} &\multicolumn{3}{c}{Total Noise}%
            \\
            \cmidrule(r){3-5}\cmidrule(r){6-8}\cmidrule(r){9-11}\cmidrule(r){12-14}
            & & Bias & Std. & RMS & Bias & Std & RMS & Bias & Std & RMS & Bias & Std & RMS \\
            \midrule
            \parbox[t]{3.0mm}{\multirow{7}{*}{\rotatebox[origin=c]{90}{Sim}}}
                &B+F \cite{shin2005block}%
                        & - & - & - %
                        & - & - & - %
                        & - & - & - %
                        & 3.12 & 1.60 & 3.51 \\
                &\baselineTitle%
                        & - & - & - %
                        & - & - & - %
                        & - & - & - %
                        & 0.17 & 0.28 & 0.33\\
                &PCA \cite{Chen15iccv}%
                        & - & - & - %
                        & - & - & - %
                        & - & - & - %
                        & 1.11 & 0.82 & 1.38 \\
                &\PGETitle \cite{byun2021fbi}%
                        & 3.01 & 1.22 & 3.25 %
                        & - & - & - %
                        & - & - & - %
                        & 3.11 & 1.26 & 3.35\\
                &\withoutMetadataTitle%
                        & 0.63 & 0.63 & 0.89 %
                        & 0.68 & 0.59 & 0.90 %
                        & 0.43 & \textbf{0.61} & \textbf{0.75} %
                        & 0.08 & 0.27 & 0.29\\
                &\minimalMetadataTitle%
                        & 1.03 & 0.21 & 1.05 %
                        & 0.80 & 0.86 & 1.18 %
                        & \textbf{0.26} & 1.35 & 1.38 %
                        & 0.77 & 0.65 & 1.00\\
                &\fullMetadataTitle%
                        & \textbf{0.14} & \textbf{0.09} & \textbf{0.17} %
                        & \textbf{0.15} & \textbf{0.45} & \textbf{0.47} %
                        & 0.82 & 0.95 & 1.25 %
                        & \textbf{0.04} & \textbf{0.19} & \textbf{0.20}\\
            [0.9ex]%
            \parbox[t]{3.0mm}{\multirow{7}{*}{\rotatebox[origin=c]{90}{Tamp.17}}}
                &B+F \cite{shin2005block}%
                        & - & - & - %
                        & - & - & - %
                        & - & - & - %
                        & 2.71 & 3.54 & 4.43 \\
                &\baselineTitle%
                        & - & - & - %
                        & - & - & - %
                        & - & - & - %
                        & 0.37 & 0.40 & 0.55\\
                &PCA \cite{Chen15iccv}%
                        & - & - & - %
                        & - & - & - %
                        & - & - & - %
                        & 3.07 & 2.77 & 4.14 \\
                &\PGETitle \cite{byun2021fbi}%
                        & 3.03 & 1.35 & 3.32 %
                        & - & - & - %
                        & - & - & - %
                        & 2.74 & 1.71 & 3.23\\
                &\withoutMetadataTitle%
                        & 0.46 & 0.68 & 0.82 %
                        & 0.83 & 0.55 & 1.00 %
                        & 0.74 & \textbf{0.76} & \textbf{1.06} %
                        & 0.26 & 0.53 & 0.59\\
                &\minimalMetadataTitle%
                        & 0.95 & 0.28 & 0.99 %
                        & 0.85 & 0.82 & 1.18 %
                        & \textbf{0.37} & 1.36 & 1.41 %
                        & 0.59 & 0.78 & 0.98\\
                &\fullMetadataTitle%
                        & \textbf{0.22} & \textbf{0.14} & \textbf{0.26} %
                        & \textbf{0.14} & \textbf{0.41} & \textbf{0.44} %
                        & 0.85 & 0.87 & 1.21 %
                        & \textbf{0.13} & \textbf{0.36} & \textbf{0.38}\\
            [0.9ex]%
            \parbox[t]{3.0mm}{\multirow{7}{*}{\rotatebox[origin=c]{90}{Udacity}}} 
                &B+F \cite{shin2005block}%
                        & - & - & - %
                        & - & - & - %
                        & - & - & - %
                        & 0.33 & 0.58 & 0.66 \\
                &\baselineTitle%
                        & - & - & - %
                        & - & - & - %
                        & - & - & - %
                        & \textbf{0.01} & 0.53 & 0.53\\
                &PCA \cite{Chen15iccv}%
                        & - & - & - %
                        & - & - & - %
                        & - & - & - %
                        & 0.14 & 0.63 & 0.64 \\
                &\PGETitle \cite{byun2021fbi}%
                        & 2.44 & 1.02 & 2.64 %
                        & - & - & - %
                        & - & - & - %
                        & 3.00 & 1.48 & 3.35\\
                &\withoutMetadataTitle%
                        & 0.44 & 0.49 & 0.66 %
                        & 0.64 & 0.57 & 0.85 %
                        & \textbf{0.27} & \textbf{0.65} & \textbf{0.70} %
                        & 0.04 & \textbf{0.27} & \textbf{0.27}\\
                &\minimalMetadataTitle%
                        & 0.63 & 0.21 & 0.66 %
                        & 0.76 & 0.84 & 1.14 %
                        & 0.28 & 1.33 & 1.36 %
                        & 0.41 & 0.68 & 0.79\\
                &\fullMetadataTitle%
                        & \textbf{0.04} & \textbf{0.10} & \textbf{0.11} %
                        & \textbf{0.17} & \textbf{0.44} &\textbf{ 0.47} %
                        & 0.87 & 0.97 & 1.30 %
                        & 0.25 & 0.30 & 0.39\\
            \bottomrule
        \end{tabular}%
    \end{adjustbox}
\end{table*}

%% file: figICX285CellarParkingLotNoiseSources.tex
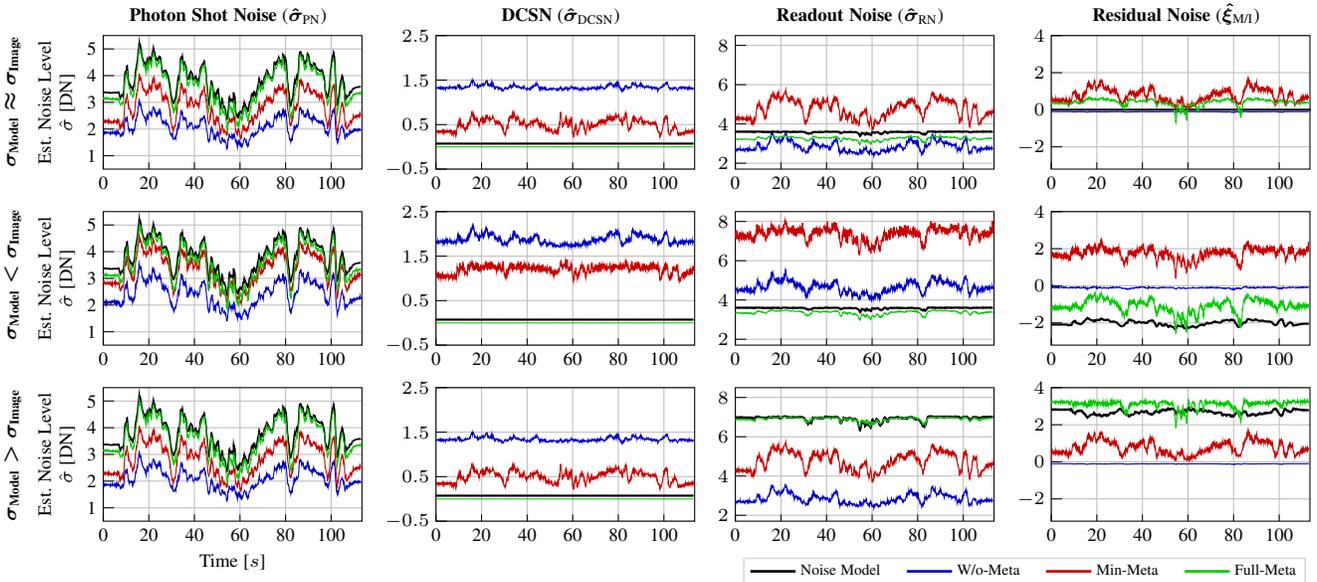
\begin{figure*}[t]
    \tikzsetnextfilename{FigICX285CellarParkingLotNoiseSources}
    \begingroup
        \pgfplotsset{every axis/.style={scale=0.52}}
        \input{plotICX285CellarParkingLotNoiseSources.tex}
    \endgroup
    \caption{\emph{Noise source estimation with and without unexpected noise (dataset: \cellar{}, camera: \ICX285{})}.
    Top row: Estimation on the uncorrupted dataset.
    Middle row: Image noise increased by random Gaussian noise $\mathcal{N}(\mu = 0, \sigma = \SI{5}{\DN})$. 
    Bottom row: Model noise increased by doubling camera parameter \emph{thermal white noise}. 
    }
    \label{fig:ICX285CellarParkingLotNoiseSources}
\end{figure*}

%% file: plotICX285CellarParkingLotNoiseSources.tex
\pgfplotstableread[col sep = comma]{dataICX285cellar1Branch.csv}\csvCellarA
\pgfplotstableread[col sep = comma]{dataICX285cellar4Branch.csv}\csvCellarB
\pgfplotstableread[col sep = comma]{dataICX285cellar4BranchConf.csv}\csvCellarC
\pgfplotstableread[col sep = comma]{dataICX285cellar4BranchAll.csv}\csvCellarD
\pgfplotstableread[col sep = comma]{dataICX285cellarnoiseModel.csv}\csvCellarE
\pgfplotstableread[col sep = semicolon]{dataICX285cellar+5GaussnoiseModel.csv}\loadedtableA
\pgfplotstableread[col sep = comma]{dataICX285cellar+5Gauss4Branch.csv}\loadedtableB
\pgfplotstableread[col sep = comma]{dataICX285cellar+5Gauss4BranchConf.csv}\loadedtableC
\pgfplotstableread[col sep = comma]{dataICX285cellar+5Gauss4BranchAll.csv}\loadedtableD
\pgfplotstableread[col sep = semicolon]{dataICX285cellar+doubleThermalNoisenoiseModel.csv}\loadedtableE
\pgfplotstableread[col sep = comma]{dataICX285cellar+doubleThermalNoise4Branch.csv}\loadedtableF
\pgfplotstableread[col sep = comma]{dataICX285cellar+doubleThermalNoise4BranchConf.csv}\loadedtableG
\pgfplotstableread[col sep = comma]{dataICX285cellar+doubleThermalNoise4BranchAll.csv}\loadedtableH
\begin{tikzpicture}
{\scriptsize
  
\begin{axis}[%
    name=axis1,
    clip mode=individual,
    tick pos=left,
    xmajorgrids,
    xmin=0, xmax=11.35,
    xtick style={color=black},
    xtick={0, 2, 4, 6, 8, 10, 12},
    xticklabels={0,20, 40, 60, 80, 100, 120},
    ymajorgrids,
    ymin=0.5, ymax=5.50,
    ytick style={color=black},
    ytick={1, 2, 3, 4, 5},
    yticklabels={1, 2, 3, 4, 5},
    xlabel near ticks,
    ylabel near ticks,
    ylabel style={align=center}, 
    ylabel={Est. Noise Level \\ $\hat{\sigma}$ [DN]},
    title={\textbf{Photon Shot Noise} ($\bm{\hat\sigma_\text{PN}}$)},
    title style={yshift=16pt},
    yscale=0.6,
    xscale=0.965
]
    
\addplot[color=black, semithick] table [x=0, y=4] {\csvCellarE};
\addplot[color=blue!80!black] table [x=0, y=4] {\csvCellarB};
\addplot[color=red!80!black] table [x=0, y=4] {\csvCellarC};
\addplot[color=green!80!black] table [x=0, y=4] {\csvCellarD};
\end{axis}

\begin{axis}[%
    name=axis2,
    at=(axis1.right of south east), 
    anchor=left of south west,
    xshift=2ex,
    clip mode=individual,
    tick pos=left,
    xmajorgrids,
    xmin=0, xmax=11.35,
    xtick style={color=black},
    xtick={0, 2, 4, 6, 8, 10, 12},
    xticklabels={0,20, 40, 60, 80, 100, 120},
    ymajorgrids,
    ymin=-0.5, ymax=2.5,
    ytick style={color=black},
    ytick={-0.5, 0.5, 1.5, 2.5},
    xlabel near ticks,
    ylabel near ticks,
    title={\textbf{DCSN} ($\bm{\hat\sigma_\text{DCSN}}$)},
    title style={yshift=16pt},
    yscale=0.6,
    xscale=0.965
]

\addplot[color=blue!80!black] table [x=0, y=5] {\csvCellarB};
\addplot[color=red!80!black] table [x=0, y=5] {\csvCellarC};
\addplot[color=green!80!black] table [x=0, y=5] {\csvCellarD};
\addplot[color=black, thick] table [x=0, y=5] {\csvCellarE};
    \end{axis}
    
\begin{axis}[%
        name=axis3,
        at=(axis2.right of south east), 
        anchor=left of south west,
        xshift=2ex,
        clip mode=individual,
        tick pos=left,
        xmajorgrids,
        xmin=0, xmax=11.35,
        xtick style={color=black},
        xtick={0, 2, 4, 6, 8, 10, 12},
        xticklabels={0,20, 40, 60, 80, 100, 120},
        ymajorgrids,
        ymin=1.7, ymax=8.5,
        ytick style={color=black},
        ytick={2, 4, 6, 8},
        xlabel near ticks,
        ylabel near ticks,
        title={\textbf{Readout Noise} ($\bm{\hat\sigma_\text{RN}}$)},
        title style={yshift=16pt},
        yscale=0.6,
        xscale=0.965
    ]

\addplot[color=blue!80!black] table [x=0, y=6] {\csvCellarB};
\addplot[color=red!80!black] table [x=0, y=6] {\csvCellarC};
\addplot[color=green!80!black] table [x=0, y=6] {\csvCellarD};
\addplot[color=black, thick] table [x=0, y=6] {\csvCellarE};
\end{axis}
    
\begin{axis}[%
        name=axis4,
        at=(axis3.right of south east), 
        anchor=left of south west,
        xshift=2ex,
        clip mode=individual,
        tick pos=left,
        xmajorgrids,
        xmin=0, xmax=11.35,
        xtick style={color=black},
        xtick={0, 2, 4, 6, 8, 10, 12},
        xticklabels={0, 20, 40, 60, 80, 100, 120},
        ymajorgrids,
        ymin=-3.2, ymax=4.0,
        ytick style={color=black},
        ytick={-2, 0, 2, 4},
        xlabel near ticks,
        ylabel near ticks,
        title={\textbf{Residual Noise} ($\bm{\restNoiseEstimation}$)},
        title style={yshift=16pt},
        yscale=0.6,
        xscale=0.965,
        legend cell align={left},
            legend image post style={scale=1.0},
            legend columns=4,
            legend style={
              nodes={scale=0.72, transform shape},
              inner xsep=2pt, 
              inner ysep=1pt,
              text opacity=1,
              at={(0,0)},
              anchor=north east,
              /tikz/every even column/.append style={column sep=8pt},
              font={\footnotesize\color{black}},
              line width=0.6pt,
              draw=lightgray,
              draw opacity=0.6,
              fill=none
            },
            legend entries={Noise Model,
                    \withoutMetadataTitle,
                    \minimalMetadataTitle,
                    \fullMetadataTitle},
            legend to name=cellarParkingLot
        ]
        
\addlegendimage{color=black, no markers, line width=1.0pt, draw opacity=1.0}
\addlegendimage{color=blue!80!black, line width=1.0pt, draw opacity=1.0}
\addlegendimage{color=red!80!black, line width=1.0pt, draw opacity=1.0}
\addlegendimage{color=green!80!black, line width=1.0pt, draw opacity=1.0}

\addplot[color=black, thick] coordinates {(0,0) (12, 0)};
\addplot[color=blue!80!black] table [x=3, y=7] {\csvCellarB};
\addplot[color=red!80!black] table [x=3, y=7] {\csvCellarC};
\addplot[color=green!80!black] table [x=3, y=7] {\csvCellarD};
\end{axis}

\begin{axis}[%
        name=axis5,
        at={(axis1.below south west)},
        anchor=north west,
        yshift=-2ex, 
        clip mode=individual,
        tick pos=left,
        xmajorgrids,
        xmin=0, xmax=11.35,
        xtick style={color=black},
        xtick={0, 2, 4, 6, 8, 10, 12},
        xticklabels={0, 20, 40, 60, 80, 100, 120},
        ymajorgrids,
        ymin=0.5, ymax=5.50,
        ytick style={color=black},
        ytick={1, 2, 3, 4, 5},
        yticklabels={1, 2, 3, 4, 5},
        xlabel near ticks,
        ylabel near ticks,
        ylabel style={align=center}, 
        ylabel={Est. Noise Level \\ $\hat{\sigma}$ [DN]},
        yscale=0.6,
        xscale=0.965
    ]
    
    \addplot[color=black, semithick] table [x=0, y=4] {\loadedtableA};
    \addplot[color=blue!80!black] table [x=0, y=4] {\loadedtableB};
    \addplot[color=red!80!black] table [x=0, y=4] {\loadedtableC};
    \addplot[color=green!80!black] table [x=0, y=4] {\loadedtableD};
\end{axis}
  
\begin{axis}[%
    name=axis6,
    at=(axis5.right of south east), 
    anchor=left of south west,
    xshift=2ex, 
    clip mode=individual,
    tick pos=left,
    xmajorgrids,
    xmin=0, xmax=11.35,
    xtick style={color=black},
    xtick={0, 2, 4, 6, 8, 10, 12},
    xticklabels={0,20, 40, 60, 80, 100, 120},
    ymajorgrids,
    ymin=-0.5, ymax=2.5,
    ytick style={color=black},
    ytick={-0.5, 0.5, 1.5, 2.5},
    xlabel near ticks,
    ylabel near ticks,
    yscale=0.6,
    xscale=0.965
]
    
    \addplot[color=black, thick] table [x=1, y=5] {\loadedtableA};
    \addplot[color=blue!80!black] table [x=1, y=5] {\loadedtableB};
    \addplot[color=red!80!black] table [x=1, y=5] {\loadedtableC};
    \addplot[color=green!80!black] table [x=1, y=5] {\loadedtableD};
\end{axis}

\begin{axis}[%
    name=axis7,
    at=(axis6.right of south east), 
    anchor=left of south west,
    xshift=2ex,
    clip mode=individual,
    tick pos=left,
    xmajorgrids,
    xmin=0, xmax=11.35,
    xtick style={color=black},
    xtick={0, 2, 4, 6, 8, 10, 12},
    xticklabels={0,20, 40, 60, 80, 100, 120},
    ymajorgrids,
    ymin=1.7, ymax=8.5,
    ytick style={color=black},
    ytick={2, 4, 6, 8},
    xlabel near ticks,
    ylabel near ticks,
    yscale=0.6,
    xscale=0.965
]

    \addplot[color=black, thick] table [x=2, y=6] {\loadedtableA};
    \addplot[color=blue!80!black] table [x=2, y=6] {\loadedtableB};
    \addplot[color=red!80!black] table [x=2, y=6] {\loadedtableC};
    \addplot[color=green!80!black] table [x=2, y=6] {\loadedtableD};
\end{axis}
    
\begin{axis}[%
        name=axis8,
        at=(axis7.right of south east), 
        anchor=left of south west,
        xshift=2ex,
        clip mode=individual,
        tick pos=left,
        xmajorgrids,
        xmin=0, xmax=11.35,
        xtick style={color=black},
        xtick={0, 2, 4, 6, 8, 10, 12},
        xticklabels={0, 20, 40, 60, 80, 100, 120},
        ymajorgrids,
        ymin=-3.2, ymax=4.0,
        ytick style={color=black},
        ytick={-2, 0, 2, 4},
        xlabel near ticks,
        ylabel near ticks,
        yscale=0.6,
        xscale=0.965
    ]

    \addplot[color=black, thick] table [x=3, y=7] {\loadedtableA};
    \addplot[color=blue!80!black] table [x=3, y=7] {\loadedtableB};
    \addplot[color=red!80!black] table [x=3, y=7] {\loadedtableC};
    \addplot[color=green!80!black] table [x=3, y=7] {\loadedtableD};
\end{axis}

\begin{axis}[%
    name=axis9,
    at={(axis5.below south west)},
    anchor=north west,
    yshift=-2ex, 
    clip mode=individual,
    tick pos=left,
    xmajorgrids,
    xmin=0, xmax=11.35,
    xtick style={color=black},
    xtick={0, 2, 4, 6, 8, 10, 12},
    xticklabels={0,20, 40, 60, 80, 100, 120},
    ymajorgrids,
    ymin=0.5, ymax=5.50,
    ytick style={color=black},
    ytick={1, 2, 3, 4, 5},
    yticklabels={1, 2, 3, 4, 5},
    xlabel near ticks,
    ylabel near ticks,
    xlabel={Time [$s$]},
    xlabel style={yshift=-4pt},
    ylabel style={align=center}, 
    ylabel={Est. Noise Level \\ $\hat{\sigma}$ [DN]},
    yscale=0.6,
    xscale=0.965
]
    
\addplot[color=black, semithick] table [x=0, y=4] {\loadedtableE};
\addplot[color=blue!80!black] table [x=0, y=4] {\loadedtableF};
\addplot[color=red!80!black] table [x=0, y=4] {\loadedtableG};
\addplot[color=green!80!black] table [x=0, y=4] {\loadedtableH};
\end{axis}

\begin{axis}[%
    name=axis10,
    at=(axis9.right of south east), 
    anchor=left of south west,
    xshift=2ex, 
    clip mode=individual,
    tick pos=left,
    xmajorgrids,
    xmin=0, xmax=11.35,
    xtick style={color=black},
    xtick={0, 2, 4, 6, 8, 10, 12},
    xticklabels={0,20, 40, 60, 80, 100, 120},
    ymajorgrids,
    ymin=-0.5, ymax=2.5,
    ytick style={color=black},
    ytick={-0.5, 0.5, 1.5, 2.5},
    xlabel near ticks,
    ylabel near ticks,
    yscale=0.6,
    xscale=0.965
]
    
    \addplot[color=black, thick] table [x=1, y=5] {\loadedtableE};
    \addplot[color=blue!80!black] table [x=1, y=5] {\loadedtableF};
    \addplot[color=red!80!black] table [x=1, y=5] {\loadedtableG};
    \addplot[color=green!80!black] table [x=1, y=5] {\loadedtableH};
\end{axis}

\begin{axis}[%
    name=axis11,
    at=(axis10.right of south east), 
    anchor=left of south west,
    xshift=2ex,
    clip mode=individual,
    tick pos=left,
    xmajorgrids,
    xmin=0, xmax=11.35,
    xtick style={color=black},
    xtick={0, 2, 4, 6, 8, 10, 12},
    xticklabels={0,20, 40, 60, 80, 100, 120},
    ymajorgrids,
    ymin=1.7, ymax=8.5,
    ytick style={color=black},
    ytick={2, 4, 6, 8},
    xlabel near ticks,
    ylabel near ticks,
    yscale=0.6,
    xscale=0.965
]

    \addplot[color=black, thick] table [x=2, y=6] {\loadedtableE};
    \addplot[color=blue!80!black] table [x=2, y=6] {\loadedtableF};
    \addplot[color=red!80!black] table [x=2, y=6] {\loadedtableG};
    \addplot[color=green!80!black] table [x=2, y=6] {\loadedtableH};
\end{axis}

\begin{axis}[%
        name=axis12,
        at=(axis11.right of south east), 
        anchor=left of south west,
        xshift=2ex,
        clip mode=individual,
        tick pos=left,
        xmajorgrids,
        xmin=0, xmax=11.35,
        xtick style={color=black},
        xtick={0, 2, 4, 6, 8, 10, 12},
        xticklabels={0, 20, 40, 60, 80, 100, 120},
        ymajorgrids,
        ymin=-3.2, ymax=4.0,
        ytick style={color=black},
        ytick={-2, 0, 2, 4},
        xlabel near ticks,
        ylabel near ticks,
        yscale=0.6,
        xscale=0.965
    ]
    
    \addplot[color=black, thick] table [x=3, y=7] {\loadedtableE};
    \addplot[color=blue!80!black] table [x=3, y=7] {\loadedtableF};
    \addplot[color=red!80!black] table [x=3, y=7] {\loadedtableG};
    \addplot[color=green!80!black] table [x=3, y=7] {\loadedtableH};
\end{axis}

\node [rotate=90] at (-1.2, 0.9) {\textbf{$\bm{\sigma_\text{Model} \approx \sigma_\text{Image}}$}}; 
\node [rotate=90] at (-1.2, -1.4) {\textbf{$\bm{\sigma_\text{Model} < \sigma_\text{Image}}$}}; 
\node [rotate=90] at (-1.2, -3.8) {\textbf{$\bm{\sigma_\text{Model} > \sigma_\text{Image}}$}}; 
}
\end{tikzpicture}
\tikzexternaldisable
    \hspace*{279pt}
    \raisebox{8pt}[0cm][0cm]{\ref*{cellarParkingLot}}
\vspace{-9pt}
\tikzexternalenable

%% file: figICX285CellarParkingLotTotalNoise.tex
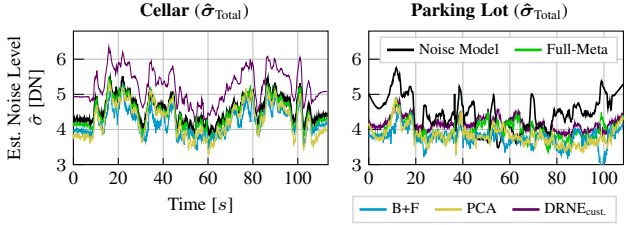
\begin{figure}[t]
    \centering
    \tikzsetnextfilename{FigICX285CellarParkingLotTotalNoise}
    \begingroup
        \pgfplotsset{every axis/.style={scale=0.52}}
        \input{plotICX285CellarParkingLotTotalNoise.tex}
    \endgroup
    \caption{\emph{Total noise estimation (datasets: \cellar{} and \parkingLot{}, camera: \ICX285{}).} 
    Compare the left plot to \cref{fig:ICX285CellarParkingLotNoiseSources}.}
    \label{fig:ICX285CellarParkingLotTotalNoise}
\end{figure}

%% file: plotICX285CellarParkingLotTotalNoise.tex
\pgfplotstableread[col sep = comma]{dataICX285cellar1Branch.csv}\csvCellarA
\pgfplotstableread[col sep = comma]{dataICX285cellar4Branch.csv}\csvCellarB
\pgfplotstableread[col sep = comma]{dataICX285cellar4BranchConf.csv}\csvCellarC
\pgfplotstableread[col sep = comma]{dataICX285cellar4BranchAll.csv}\csvCellarD
\pgfplotstableread[col sep = comma]{dataICX285cellarnoiseModel.csv}\csvCellarE
\pgfplotstableread[col sep = comma]{dataICX285cellarB+F.csv}\csvCellarF
\pgfplotstableread[col sep = comma]{dataICX285cellarPCA.csv}\csvCellarG

\pgfplotstableread[col sep = comma]{dataICX285parkingLot1Branch.csv}\csvParkingLotA
\pgfplotstableread[col sep = comma]{dataICX285parkingLot4Branch.csv}\csvParkingLotB
\pgfplotstableread[col sep = comma]{dataICX285parkingLot4BranchConf.csv}\csvParkingLotC
\pgfplotstableread[col sep = comma]{dataICX285parkingLot4BranchAll.csv}\csvParkingLotD
\pgfplotstableread[col sep = comma]{dataICX285parkingLotnoiseModel.csv}\csvParkingLotE
\pgfplotstableread[col sep = comma]{dataICX285parkingLotB+F.csv}\csvParkingLotF
\pgfplotstableread[col sep = comma]{dataICX285parkingLotPCA.csv}\csvParkingLotG
\begin{tikzpicture}
{\scriptsize

\begin{axis}[%
        name=axis1,
        clip mode=individual,
        tick pos=left,
        xmajorgrids,
        xmin=0, xmax=11.35,
        xtick style={color=black},
        xtick={0, 2, 4, 6, 8, 10, 12},
        xticklabels={0,20, 40, 60, 80, 100, 120},
        ymajorgrids,
        ymin=3, ymax=6.8,
        ytick style={color=black},
        ytick={3, 4, 5, 6, 7},
        xlabel near ticks,
        xlabel={Time [$s$]},
        xlabel style={yshift=-4pt},
        ylabel near ticks,
        ylabel style={align=center}, 
        ylabel={Est. Noise Level \\ $\hat{\sigma}$ [DN]},
        title={\textbf{Cellar} ($\bm{\hat\sigma_\text{Total}}$)},
        title style={yshift=16pt},
        yscale=0.6,
        xscale=0.95,
        legend cell align={left},
            legend image post style={scale=0.6},
            legend columns=3,
            legend style={
              nodes={scale=0.72, transform shape},
              inner xsep=2pt, 
              inner ysep=1pt,
              text opacity=1,
              at={(1.03,1.6)},
              anchor=north east,
              /tikz/every even column/.append style={column sep=4pt},
              font={\footnotesize\color{black}},
              line width=0.6pt,
              draw=lightgray,
              draw opacity=0.6
            },
            legend entries={B+F,
                PCA,
                \baselineTitle
            },
            legend to name=ICX285TotalNoiseComparisonMethodsLegend
    ]

    \addlegendimage{color=cyan!80!black, line width=1.0pt, draw opacity=1.0}
    \addlegendimage{color=yellow!80!black, line width=1.0pt, draw opacity=1.0}
    \addlegendimage{color=violet!80!black, line width=1.0pt, draw opacity=1.0}
    
    \addplot[color=black, semithick] table [x=3, y=7, col sep=comma] {\csvCellarA};
    \addplot[color=violet!80!black] table [x=3, y expr={sqrt(\thisrowno{4}^2 + \thisrowno{5}^2 + \thisrowno{6}^2)}] {\csvCellarE};
    \addplot[color=green!80!black] table [x=3, y expr={sqrt(\thisrowno{4}^2 + \thisrowno{5}^2 + \thisrowno{6}^2) - \thisrowno{7}}] {\csvCellarD};
    \addplot[color=cyan!80!black] table [x=0, y=7] {\csvCellarF};
    \addplot[color=yellow!80!black] table [x=0, y=7] {\csvCellarG};

\end{axis}

\begin{axis}[%
        name=axis2,
        at=(axis1.right of south east), 
        anchor=left of south west,
        xshift=2ex, 
        clip mode=individual,
        tick pos=left,
        xmajorgrids,
        xmin=0, xmax=10.83,
        xtick style={color=black},
        xtick={0, 2, 4, 6, 8, 10, 12},
        xticklabels={0,20, 40, 60, 80, 100, 120},
        ymajorgrids,
        ymin=3, ymax=6.8,
        ytick style={color=black},
        ytick={3, 4, 5, 6, 7},
        yscale=0.6,
        xscale=0.95,
        xlabel near ticks,
        ylabel near ticks,
        title={\textbf{Parking Lot} ($\bm{\hat\sigma_\text{Total}}$)},
        title style={yshift=16pt},
        legend cell align={left},
            legend image post style={scale=0.6},
            legend columns=3,
            legend style={
              nodes={scale=0.72, transform shape},
              inner xsep=2pt, 
              inner ysep=1pt,
              text opacity=1,
              at={(1.03,1.6)},
              anchor=north east,
              /tikz/every even column/.append style={column sep=4pt},
              font={\footnotesize\color{black}},
              line width=0.6pt,
              draw=lightgray,
              draw opacity=0.6
            },
            legend entries={Noise Model,
            \fullMetadataTitle
            }
    ]

    \addlegendimage{color=black, no markers, line width=1.0pt, draw opacity=1.0}
    \addlegendimage{color=green!80!black, line width=1.0pt, draw opacity=1.0}
    
    \addplot[color=violet!80!black] table [x=3, y=7] {\csvParkingLotA};
    \addplot[color=black, semithick] table [x=3,y expr={sqrt(\thisrowno{4}^2 + \thisrowno{5}^2 + \thisrowno{6}^2)}] {\csvParkingLotE};
    \addplot[color=green!80!black] table [x=3, y expr={sqrt(\thisrowno{4}^2 + \thisrowno{5}^2 + \thisrowno{6}^2) - \thisrowno{7}}] {\csvParkingLotD};
    \addplot[color=cyan!80!black] table [x=0, y=7] {\csvParkingLotF};
    \addplot[color=yellow!80!black] table [x=0, y=7] {\csvParkingLotG};
\end{axis}
}
\end{tikzpicture}

\tikzexternaldisable
    \hspace*{130pt}
    \raisebox{10pt}[0cm][0cm]{\ref*{ICX285TotalNoiseComparisonMethodsLegend}}
    \vspace{-6pt}
\tikzexternalenable

%% file: tabDenoisingICX285.tex
\DeclareFontSeriesDefault[rm]{bf}{b}%
\begin{table*}[t]
\caption[Denoising performance for real-world images.]{\emph{Denoising performance for real-world images (camera: ICX 285).}
    Best PSNR (dB $\uparrow$) and SSIM ($\uparrow$) scores per dataset, noise level, and metric are highlighted in bold, the second best are underlined (in the case of equal numbers, the decision is made on the basis of further decimal places).
    \label{tab:noiseSourceEstimation:denoising}
    }
    \centering
        \begin{adjustbox}{max width=1.00\linewidth}
        \begin{tabular}{l|lccccc}
            \toprule
            \multicolumn{2}{c}{} & \multicolumn{5}{c}{Number of raw images for averaging}\\
            \cmidrule(r){3-7}
            \multicolumn{2}{l}{\hspace{6mm} Method} & 1 & 2 & 4 & 8 & 16\\
            \midrule
            \multirow{11}{*}{\rotatebox[origin=c]{90}{Cellar}} & Raw & 34.75  /  0.7730 & 37.61 / 0.8703 & 40.42 / 0.9322 & 43.21 / 0.9680 & 46.13 / 0.9872\\
            &$\text{DRNE}_\text{cust.}$ + BM3D \cite{dabov2007image} & 43.01 / 0.9803 & 44.47 / 0.9853 & 45.53 / 0.9886 & 46.49 / \underline{0.9913} & \underline{47.59} / \underline{0.9932}\\
            &w/o-Meta + BM3D \cite{dabov2007image} & 41.42 / 0.9671 & 43.83 / 0.9817 & 45.01 / 0.9861 & 46.26 / 0.9905 & 47.56 / 0.9930\\
            &Min-Meta + BM3D \cite{dabov2007image} & \underline{43.30} / 0.9818 & \underline{44.72} / \underline{0.9864} & \underline{45.67} / \underline{0.9899} & \underline{46.49} / 0.9912 & 47.32 / 0.9922\\
		    &Full-Meta + BM3D  \cite{dabov2007image}& \textbf{43.74} / \textbf{0.9839} & \textbf{45.00} / \textbf{0.9875} & \textbf{45.99} / \textbf{0.9901} & \textbf{46.68} / \textbf{0.9916} & \textbf{47.63} / \textbf{0.9934}\\
			&$\text{DRNE}_\text{cust.}$ + NLM \cite{buades2005non}& 42.46 / 0.9793 & 43.91 / 0.9841 & 45.05 / 0.9874 & 46.09 / 0.9902 & 47.32 / 0.9925\\
			&w/o-Meta + NLM \cite{buades2005non}& 41.08 / 0.9701 & 43.33 / 0.9812 & 44.66 / 0.9322 & 45.94 / 0.9897 & 47.28 / 0.9924\\
			&Min-Meta + NLM \cite{buades2005non}& 42.77 / 0.9809 & 44.18 / 0.9852 & 45.18 / 0.9879 & 46.09 / 0.9902 & 46.92 / 0.9911\\
			&Full-Meta + NLM \cite{buades2005non}& 43.19 / 0.9828 & 44.47 / 0.9863 & 45.50 / 0.9889 & 46.26 / 0.9905 & 47.35 / 0.9927\\
			&FBI-Denoiser \cite{byun2021fbi}& 41.69 / \underline{0.9830} & 42.07 / 0.9851 & 42.29 / 0.9865 & 42.41 / 0.9871 & 42.62 / 0.9880\\
			&Blind2Unblind \cite{wang2022blind2unblind}& 43.02 / 0.9515 & 43.67 / 0.9574 & 44.10 / 0.9616 & 44.41 / 0.9643 & 44.77 / 0.9660\\
            \midrule
            \multirow{11}{*}{\rotatebox[origin=c]{90}{Parking Lot}} & Raw & 31.09 / 0.7890 & 32.34 / 0.8780 & 33.29 / 0.9330 & 34.07 / 0.9625 & 35.24 / 0.9786\\
            &$\text{DRNE}_\text{cust.}$ + BM3D \cite{dabov2007image}& 33.39 / 0.9546 & 33.78 / 0.9639 & 34.11 / \underline{0.9713} & \underline{34.51} / \underline{0.9770} & \underline{35.39} / \underline{0.9820}\\
            &w/o-Meta + BM3D \cite{dabov2007image}& 33.04 / 0.9394 & 33.70 / 0.9612 & 34.07 / 0.9705 & 34.50 / \underline{0.9770} & 35.37 / 0.9817\\
            &Min-Meta + BM3D \cite{dabov2007image}& \underline{33.47} / \textbf{0.9573} & \underline{33.83} / \textbf{0.9651} & \underline{34.11} / 0.9710 & 34.44 / 0.9749 & 35.20 / 0.9780\\
		    &Full-Meta + BM3D \cite{dabov2007image}& 33.40 / \underline{0.9553} & 33.81 / \underline{0.9648} & \textbf{34.11} / \textbf{0.9715} & \textbf{34.51} / \textbf{0.9771} & \textbf{35.40} / \textbf{0.9823}\\
			&$\text{DRNE}_\text{cust.}$ + NLM \cite{buades2005non}& 33.14 / 0.9464 & 33.56 / 0.9567 & 33.92 / 0.9655 & 34.36 / 0.9731 & 35.29 / 0.9799\\
			&w/o-Meta + NLM \cite{buades2005non}& 32.93 / 0.7890 & 33.52 / 0.9559 & 33.91 / 0.9657 & 34.37 / 0.9738 & 35.26 / 0.9794\\
			&Min-Meta + NLM \cite{buades2005non}& 33.11 / 0.9447 & 33.52 / 0.9547 & 33.84 / 0.9622 & 34.17 / 0.9670 & 34.88 / 0.9709\\
			&Full-Meta + NLM \cite{buades2005non}& 33.14 / 0.9468 & 33.56 / 0.9566 & 33.92 / 0.9657 & 34.37 / 0.9734 & 35.31 / 0.9804\\
			&FBI-Denoiser \cite{byun2021fbi}& 32.52 / 0.9450 & 32.67 / 0.9522 & 32.76 / 0.9573 & 32.87 / 0.9604 & 33.05 / 0.9630\\
			&Blind2Unblind \cite{wang2022blind2unblind}& \textbf{33.61} / 0.9156 & \textbf{33.86} / 0.9282 & 34.05 / 0.9379 & 34.27 / 0.9441 & 34.74 / 0.9488\\
            \bottomrule%
        \end{tabular}%
        \end{adjustbox}
            
\end{table*} 
\DeclareFontSeriesDefault[rm]{bf}{bx}%

%% file: tabDenoisingEV76C661.tex
\DeclareFontSeriesDefault[rm]{bf}{b}%
\begin{table*}[t!]
\caption[Denoising performance for real-world noised images (\EV76C661{}).]{\emph{Denoising performance for real-world noised images (camera: \EV76C661{}).}
    Best PSNR (dB $\uparrow$) and SSIM ($\uparrow$) scores per dataset, noise level, and metric are highlighted in bold, the second best are underlined.
    Compare to \cref{tab:noiseSourceEstimation:denoising}.
    \label{tab:noiseSourceEstimation:denoising:EV76C661}
    }
    \centering
        \begin{adjustbox}{max width=1.00\linewidth}
        \begin{tabular}{l|lccccc}
            \toprule
            \multicolumn{2}{c}{} & \multicolumn{5}{c}{Number of raw images for averaging}\\
            \cmidrule(r){3-7}
            \multicolumn{2}{l}{\hspace{6mm} Method} & 1 & 2 & 4 & 8 & 16\\
            \midrule
            \multirow{11}{*}{\rotatebox[origin=c]{90}{Cellar}} & Raw & 28.59 / 0.6052 & 30.73 / 0.7447 & 32.40 / 0.8517 & 33.80 / 0.9227 & \textbf{35.36} / \textbf{0.9651}\\
            &$\text{DRNE}_\text{cust.}$ + BM3D \cite{dabov2007image}& \underline{33.30} / \underline{0.9136} & \textbf{33.84} / \textbf{0.9221} & 34.06 / 0.9280 & \underline{34.30} / \underline{0.9342} & \underline{34.85} / \underline{0.9399}\\
            &w/o-Meta + BM3D \cite{dabov2007image}& 33.29 / 0.9132 & 33.83 / \textbf{0.9221} & \underline{34.08} / \underline{0.9298} & 34.27 / 0.9322 & 34.72 / 0.9337\\
            &Min-Meta + BM3D \cite{dabov2007image}& 33.28 / 0.9125 & 33.75 / 0.9192 & 34.05 / 0.9268& 34.29 / 0.9339 & 34.83 / 0.9387\\
		    &Full-Meta + BM3D \cite{dabov2007image}& \textbf{33.33} / \textbf{0.9151} & \underline{33.83} / \underline{0.9209} & \textbf{34.10} / \textbf{0.9317} & \textbf{34.37} / \textbf{0.9388} & 34.83 / 0.9386\\
			&$\text{DRNE}_\text{cust.}$ + NLM \cite{buades2005non}& 33.22 / 0.9132 & 33.73 / 0.9200 & 33.95 / 0.9248 & 34.19 / 0.9296 & 34.71 / 0.9341\\
			&w/o-Meta + NLM \cite{buades2005non}& 33.22 / 0.9133 & 33.74 / 0.9207 & 33.97 / 0.9263 & 34.17 / 0.9283 & 34.62 / 0.9299\\
			&Min-Meta + NLM \cite{buades2005non}& 33.22 / 0.9131 & 33.69 / 0.9177 & 33.93 / 0.9238 & 34.18 / 0.9294 & 34.70 / 0.9332\\
			&Full-Meta + NLM \cite{buades2005non}& 33.22 / 0.9133 & 33.73 / 0.9219 & 34.00 / 0.9285 & 34.24 / 0.9327 & 34.70 / 0.9333\\
			&FBI-Denoiser \cite{byun2021fbi}& 33.18 / 0.9127 & 33.65 / 0.9196 & 33.82 / 0.9231 & 33.99 / 0.9251 & 34.42 / 0.9264\\
			&Blind2Unblind \cite{wang2022blind2unblind}& 33.14 / 0.8921 & 33.60 / 0.9008 & 33.79 / 0.9075 & 33.99 / 0.9122 & 34.34 / 0.9149\\
            \midrule
            \multirow{11}{*}{\rotatebox[origin=c]{90}{Parking Lot}} & Raw & 31.06 / 0.6734 & 33.31 / 0.8050 & 35.40 / 0.8953 & 37.12 / 0.9499 & \textbf{37.78} / \textbf{0.9806}\\
            &$\text{DRNE}_\text{cust.}$ + BM3D \cite{dabov2007image}& \underline{35.94} / \textbf{0.9342} & \underline{36.50} / \textbf{0.9416} & 37.10 / 0.9476 & 37.47 / 0.9537 & 37.33 / 0.9605\\
            &w/o-Meta + BM3D \cite{dabov2007image}& 35.42 / 0.9198 & 36.38 / 0.9407 & 37.16 / 0.9502 & 37.56 / 0.9566 & 37.46 / 0.9661\\
            &Min-Meta + BM3D \cite{dabov2007image}& 35.62 / 0.9262 & 36.11 / 0.9334 & 37.18 / \underline{0.9519} & 37.57 / \underline{0.9568} & 37.43 / 0.9649\\
		    &Full-Meta + BM3D \cite{dabov2007image}& 35.85 / \underline{0.9324} & 36.43 / \underline{0.9415} & \underline{37.19} / \textbf{0.9524} & \textbf{37.82} / \textbf{0.9643} & \underline{37.70} / \underline{0.9757}\\
			&$\text{DRNE}_\text{cust.}$ + NLM \cite{buades2005non}& 35.67 / 0.9312 & 36.24 / 0.9377 & 36.87 / 0.9437 & 37.27 / 0.9498 & 37.78 / 0.9560\\
			&w/o-Meta + NLM \cite{buades2005non}& 35.52 / 0.9295 & 36.27 / 0.9404 & 36.96 / 0.9467 & 37.35 / 0.9522 & 37.30 / 0.9604\\
			&Min-Meta + NLM \cite{buades2005non}& 35.60 / 0.9314 & 36.13 / 0.9376 & 36.99 / 0.9483 & 37.36 / 0.9524 & 37.27 / 0.9594\\
			&Full-Meta + NLM \cite{buades2005non}& 35.67 / 0.9320 & 36.28 / 0.9403 & 37.00 / 0.9488 & 37.55 / 0.9581 & 37.52 / 0.9694\\
			&FBI-Denoiser \cite{byun2021fbi}& 35.58 / 0.9287 & 36.02 / 0.9345 & 36.51 / 0.9389 & 36.70 / 0.9420 & 36.59 / 0.9446\\
			&Blind2Unblind \cite{wang2022blind2unblind}& \textbf{36.37} / 0.8109 & \textbf{36.86} / 0.8243 & \textbf{37.37} / 0.8358 & \underline{37.65} / 0.8446 & 37.51 / 0.8511\\
            \bottomrule%
        \end{tabular}%
        \end{adjustbox}
            
\end{table*} 
\DeclareFontSeriesDefault[rm]{bf}{bx}%

%% file: figDenoisingResultImages.tex
\begin{figure*}[t!]
    \includegraphics[width=1.0\textwidth, trim=0 0 0 0, clip]{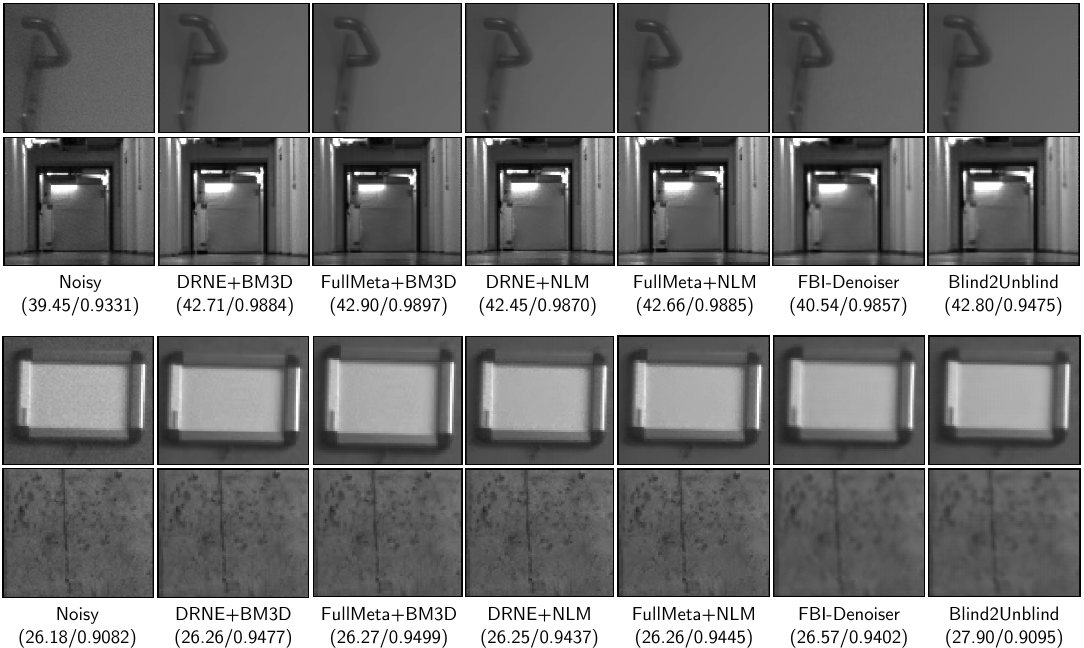}
    \caption[Exemplary denoised images.]{\emph{Exemplary denoising results for real-world noised images (four averaged images, top rows: \cellar{}, bottom rows: \parkingLot{}).} 
    Brightness and contrast are adapted for better visualization.
    FullMeta+BM3D best removes the noise while preserving image details.
    In contrast, FBI-Denoiser and Blind2Unblind remove noise best visually, but smooth the entire image (both) and introduce square artifacts (Blind2Unblind).
    NLM tends to retain noise at edges.%
    \label{fig:realWorldDenoisingResults}%
    } 
\end{figure*}

%% file: tabMetadataSensitivityAnalysis.tex
\begin{table}[t!]
\caption[Input-output sensitivity analysis of Full-Meta compared to the noise model.]{\emph{Input-output sensitivity analysis of Full-Meta (bottom) compared to the noise model (top).} 
    Input: One parameter is sampled at a time while the rest are fixed to respective maximum values (to generate high noise levels) and the (uncorrupted) mean image intensity to \SI{128}{\DN} (to avoid saturation). 
    Parameter value ranges are provided in \cref{tab:cameraMetadata} and concrete sampled values in the appendix.
    Output: Estimated total noise level (table cells, in DN) per input parameter configuration.
    $\star \colon$ The influence of the pixel clock rate highly depends on metadata that we fixed during the experiments, such as the correlated double sampling dominant time constant.
    $\star\star  \colon$ Simulated CCD sensor (CMOS sensor otherwise).
    \label{tab:cameraMetadataSensitivityAnalysisEstimator}
    }
    \centering
        \setlength{\tabcolsep}{4pt}
        \begin{adjustbox}{max width=1.00\linewidth}
        \begin{tabular}{c|lcccccccccc}
            \toprule
            \multicolumn{2}{c}{}&\multicolumn{10}{c}{Uniform samples of parameter value ranges}\\
            \cmidrule(r){3-12}
            \multicolumn{2}{c}{}& Min & 1 & 2 & 3 & 4 & 5 & 6 & 7 & 8 & Max \\
            \midrule
            \multirow{14}{*}{\rotatebox[origin=c]{90}{Noise Model}}
            &Mean Img Intensity & 5.54 &9.87 &9.94 &10.0& 10.1 &10.1 &10.2 &10.2 &10.3 &6.1\\
            [0.75ex]
            &\textbf{Minimal Metadata} & & & & & & & & & \\
            &Camera Gain & 0.82 & 1.86 & 2.88 & 3.92 & 4.94 & 5.96 & 6.99 & 8.02 & 9.02 & 10.1\\
            &Exposure Time & 4.02 & 5.06 & 5.93 & 6.67 & 7.35 & 7.96 & 8.53 & 9.08 & 9.60 & 10.1\\
            &Sensor Temperature & 3.69 & 3.76 & 3.86 & 4.03 & 4.31 & 4.77 & 5.49 & 6.56 & 8.05 & 10.1\\
            [0.75ex]
            &\textbf{Full Metadata} & & & & & & & & & &\\
            &Dark Signal FoM & 3.96 & 5.02 & 5.89 & 6.65 & 7.34 & 7.95 & 8.53 & 9.06 & 9.57 & 10.1\\
            &Full Well Capacity & 118.4 & 69.6 & 41.4 & 28.6 & 21.8 & 17.6 & 14.8 & 12.8 & 11.3 & 10.1\\
            &$\text{Pixel Clock Rate}^{\star}$ & 3.33 & 3.33 & 3.33 & 3.33 & 3.33 & 3.33 & 3.33 & 3.33 & 3.33 & 3.33\\
            &Sense Node (SN) Gain & 12.9 & 11.7 & 11.1 & 10.8& 10.6 & 10.4 & 10.3 & 10.2 & 10.1 & 10.1\\
            &SN Reset Factor & 9.56 & 9.56 & 9.58 & 9.59 & 9.67 & 9.71 & 9.76 & 9.85 & 9.95 & 10.1\\
            &Sensor Pixel Size & 4.05 & 4.34 & 4.80 & 5.38 & 6.05 & 6.79 & 7.57 & 8.39 & 9.21 & 10.1\\
            &$\text{Thermal Wh. Noise}^{\star\star}$ & 10.0 & 10.0 & 10.0 & 10.0 & 10.0 & 10.0 & 10.0 & 10.0 & 10.1 & 10.1\\
            \midrule
            \multirow{13}{*}{\rotatebox[origin=c]{90}{Noise Source Estimator (Full-Meta)}}
            &Mean Img Intensity & \textcolor{red}{7.05} & 9.57 &9.87 &9.98& 10.1 &10.1 &10.1 &9.95 &9.56 &\textcolor{red}{8.52}\\
            [0.75ex]
            &\textbf{Minimal Metadata} & & & & & & & & & \\
            &Camera Gain & 0.90 & 1.70 & 2.56 & 3.53 & 4.15 & 5.11 & 6.61 & 7.84 & 8.86 & 10.1\\
            &Exposure Time & \textcolor{orange}{5.31} & \textcolor{orange}{6.07} & 6.70 & 7.24 & 7.77 & 8.26 & 8.67 & 9.03 & 9.39 & 10.1\\
            &Sensor Temperature & 4.34 & 4.49 & 4.71 & 4.98 & \textcolor{orange}{5.34} & \textcolor{orange}{5.85} & 6.36 & 7.23 & 8.48 & 10.1\\
            [0.75ex]
            &\textbf{Full Metadata} & & & & & & & & & &\\
            &Dark Signal FoM & 4.76 & \textcolor{orange}{6.17} & 6.47 & 7.20 & 7.71 & 8.19 & 8.68 & 9.19 & 9.78 & 10.1\\
            &Full Well Capacity & \textcolor{red}{167.2} & \textcolor{red}{62.7} & \textcolor{red}{34.4} & 27.7 & 21.4 & 16.3 & 14.7 & 13.03 & 11.5 & 10.1\\
            &$\text{Pixel Clock Rate}^{\star}$ & 4.22 & 4.22 & 4.22 & 4.22 & 4.22 & 4.22 & 4.22 & 4.22 & 4.22 & 4.22\\
            &Sense Node (SN) Gain & 13.5 & \textcolor{orange}{12.8} & 12.1 & 11.4 & 10.8 & 10.6 & 10.5 & 10.4& 10.2 & 10.1\\
            &SN Reset Factor & \textcolor{orange}{8.23} & \textcolor{orange}{8.44} & 8.64 & 8.77 & 8.85 & 8.94 & 9.07 & 9.38 &9.72 & 10.1\\
            &Sensor Pixel Size & 4.88 & 5.22 & 5.56 & 6.01 & 6.51 & 7.05 & 7.72& 8.39 & 9.08 & 10.1\\
            &$\text{Thermal Wh. Noise}^{\star\star}$ & 9.46 & 9.54 & 9.71 & 9.87 & 10.0 & 10.1 & 10.1 &10.1 & 10.1 & 10.1\\
            \bottomrule%
        \end{tabular}%
        \end{adjustbox}
\end{table} 

%% file: conclusion.tex
\section{Conclusion}
We have proposed a noise source estimator that quantifies contributions of individual camera noise sources using an image with metadata to tackle noise at its root causes as opposed to tackling its symptoms.
It is memory-efficient, runs in real-time and its broad range of learned camera systems makes it directly applicable to many mobile agents.
Comparing the three versions of the estimator, we validated the natural hypothesis that the more camera metadata is available and relevant, the better the noise source identification (\fullMetadata{}).
Moreover, the developed estimator (\fullMetadata{}) promotes a reliable application by its ability to detect unexpected influences in image noise and the metadata.
We have evaluated its functionality in extensive experiments including real-world noise from two camera systems, a self-simulated and three standard datasets, the application in two field campaigns with unexpected image noise and metadata changes, and in a sensitivity analysis of the input parameters.
Lastly, the improved total noise estimation has been demonstrated in the context of the downstream vision application of image denoising.
In the future our method could be integrated into a machine's feedback-loop to perform automatic countermeasures \cite{wischow2021camera}.
As our estimators already show notable experimental performance, we follow the baseline \cite{tan2019pixelwise} and leave ablations for the future.

%% file: appendix.tex
\section*{Supplementary Material}
The following supplementary material starts with \gblue{details about the architecture of the proposed noise estimation DNN (\cref{sec:detailsArchitecture}), about} our assumed noise types (\cref{sec:noiseModel}) and follows with the real-world noise acquired by two cameras (\cref{sec:detailsRealWorldNoise}), including details about the used camera metadata (\cref{sec:cameraMetadataDetails}).
Subsequently, we list extended results for the quantitative and the qualitative experiments (Sections~\ref{sec:extendedQuantitativeExperimentResults} and \ref{sec:extendedQualitativeExperimentResults}).
For a better context, we reference corresponding sections, figures and tables from the main paper.

\section{Noise Source Estimation Model Setup}
\label{sec:detailsArchitecture}
\input{tabNNSetup}
\gblue{Table \ref{tab:nnsetup} depicts the layerwise setup of the proposed \fullMetadata{} noise source estimation model and corresponding layer resolutions (cf.~Fig.~\ref{fig:NNModels}, right).
The first part with two residual blocks (each one with five convolutional layers) comprises the processing of the input image patch and is retained from the original DRNE network.
At its end, the network splits into four branches, each for one noise source (PN, DRNE, RN, and $\xi$).
The first three branches have the same structure: %
First, the most important learnt image features for the respective noise type are extracted using 2D global pooling.
Second, the unravelled features are concatenated with additionally passed camera metadata parameters.
Concatenation enables the neural network to learn the relationship between the heterogeneous data independently, without relying on assumptions inherent in a customized data fusion.
Moreover, each branch gets only camera metadata relevant to the noise type according to the theoretical noise model \cite{konnik2014high}.
Subsequently, multiple dense layers allow the network to learn a suitable data fusion and the relationship between the data and the respective noise levels.
Finally, the last branch receives its own 2D-global-pooled features and the results of the previous three branches. 
Analogously, a concatenation of the heterogeneous data and three dense layers follow in order to determine the difference between the image noise and the noise indicated by the metadata.
}

\input{noiseModel}

\section{Real-World Noise Acquisition: Details} 
\label{sec:detailsRealWorldNoise}
The following descriptions of the real-world noise generation and processing refer to \cref{subsec:datasetsWithGT}.

\textbf{Real-World Noise Generation. }%
The noise generation took place in a darkroom with closed camera apertures to prevent light signal.
\gblue{We captured bias frames as RN and dark frames as DCSN images \cite[pp.~163--164]{woodhouse2015astrophotography}.}
Moreover, we disabled all image post-processing and used the highest possible camera bit depths (\num{12} and \num{10} bit) to minimize quantization errors.
To generate RN, we set the exposure time to the minimum of $\SI{0.001}{\second}$ to counteract dark current integration, applied a random camera gain from $[0, 24] \, \si{\deci\bel}$, and generated multiple image sequences.
In order to generate corresponding DCSN images right after an RN image sequence, we sampled another exposure time from $[0.001, 0.2] \, \si{\second}$ but kept the same gain (note that RN is still included in the DCSN images at this point).

\textbf{Real-World Noise Processing. }%
We identify three issues with the raw noise images that needed to be addressed in a post-processing: %
($i$) The DCSN and RN intensity distributions may be truncated with all negative intensity values set to zero, which supports our zero-camera-offset assumption, but also affects the ground truth noise level determination. 
We tackle this issue in four steps: %
($a$) In each noise image histogram we determine the bin $x_{\max}$ that corresponds to the distribution maximum, 
($b$) mirror histogram bins $x \geq 2x_{\max}$ along the vertical axis at $x_{\max}$ to reconstruct bins $x \leq 0$, 
($c$) fit a Gaussian distribution $\mathcal{N}(\mu, \sigma^2)$ into the fixed histogram, 
and ($d$) sample a new noise image from $\mathcal{N}(\mu, \sigma^2)$ (details in supplementary material). 
($ii$) Secondly, the images $I_{\text{DCSN}}$ still contain RN.
We approach this issue during the step ($i.c$) by calculating a rectified Gaussian distribution $\mathcal{N}(\mu_\text{DCSN*}, \sigma_\text{DCSN*}^2)$ with
\begin{equation}
\begin{split}
    \mu_\text{DCSN*} &\doteq \mu_\text{DCSN} - \mu_\text{RN}, \\
    \sigma_\text{DCSN*}^2 &\doteq \sigma^{2}_\text{DCSN} - \sigma^{2}_\text{RN},
\end{split}
\end{equation}
following the central limit theorem for the addition of two statistically independent Gaussian distributed random variables \cite[pp.~230--232]{devore2011Probability}.
Corrected DCSN images in ($i.d$) are then sampled from this rectified distribution.
($iii$) Lastly, we observe residual fixed-pattern noise in multiple noise images (e.g., $I_\text{DCSN}$ may still contain dark current).
To counteract this, for each image set $I_{i \in \{\text{DCSN, RN}\}}$ we calculate pixel-wise means from the respective first \num{20} images and removed this mean image from the remaining images in the set.
The whole real-world noise processing is summarized in \cref{alg:dataPreprocessing}.
\input{algDataPreprocessing}

\section{Camera Metadata: Details} 
\label{sec:cameraMetadataDetails}
\Cref{tab:fixedCameraMetadata,tab:cameraMetadataDefinitions,tab:cameraMetadataSensitivityAnalysisParameterValues} complement \cref{tab:cameraMetadata}.
\Cref{tab:fixedCameraMetadata} provides the fixed camera metadata used in the adopted noise model \cite{konnik2014high},
\cref{tab:cameraMetadataDefinitions} defines all camera parameters used in our study,
\gblue{and \cref{tab:cameraMetadataSensitivityAnalysisParameterValues} lists the used parameter values for the metadata sensitivity analysis in \cref{tab:cameraMetadataSensitivityAnalysisEstimator}.}

\section{Quantitative Experiments (Extended Results)} 
\label{sec:extendedQuantitativeExperimentResults}
\Cref{tab:KITTIResults,tab:SN06SimulatedRealNoiseEstimation} extend the results from \cref{subsec:quantitativeExperiments}.
\Cref{tab:KITTIResults} provides noise estimation results on the KITTI dataset and \cref{tab:SN06SimulatedRealNoiseEstimation} reports results estimating real-world noise from the camera camera \EV76C661{}.

\section{Experiments on Real-world Platforms (Extended Results)} 
\label{sec:extendedQualitativeExperimentResults}
Figures \ref{fig:ICX285ParkingLot} to \ref{fig:ICX285CellarTempCorrupted} extend the results from \cref{subsec:qualitativeExperiments}.
\Cref{subsec:parkingLot,fig:ICX285ParkingLot} first presents noise source estimation from the \parkingLot{} dataset recorded with the camera \ICX285{} (cf.~\cref{subsec:soundQualitativeExperiments}).
Subsequently, \cref{fig:EV76C661CellarParkingLotNoiseSources,fig:EV76C661CellarParkingLotTotalNoise} illustrate the noise source estimation from the \cellar{} and \parkingLot{} datasets recorded with the camera \EV76C661{} (cf.~\cref{subsec:soundQualitativeExperiments}).
\Cref{fig:underexposureExamplesParkingLot,fig:overexposureExamplesCellar} depict corresponding exemplary images of under- and over-exposure effects that we observed in the \cellar{} and \parkingLot{} datasets.
Lastly, \cref{fig:ICX285CellarTempCorrupted} shows noise source estimation of the \cellar{} dataset with corrupted \emph{sensor temperature} metadata (cf.~\cref{subsec:corruptedQualitativeExperiments}).

\subsection{Parking Lot}
\label{subsec:parkingLot}
The \parkingLot{} scenario leads to similar estimations as \cellar{} (compare \cref{fig:ICX285ParkingLot} to \cref{fig:ICX285CellarParkingLotNoiseSources}).
The smaller PN predictions of the noise model result from the lower scene illumination.
In particular, we noticed large under-exposed areas at timestamps where $\hat\sigma_{\text{Noise Model}} < \SI{2}{\DN}$ (\cref{fig:underexposureExamplesParkingLot}).
Although this case is covered in the noise model and considered by our training data augmentation, it still impacts \fullMetadata{} in that the method detects this model/image mismatch in the residual noise plot for the respective time stamps $\{t| [20, 22] \cup [46, 62] \cup [72, 79]\} \, \si{\second}$.
We also observe a similar over-exposure behavior in the \cellar{} dataset (\cref{fig:overexposureExamplesCellar}).
We think that these use-cases are still not sufficiently represented in the training data.
Note that this neither affects \withoutMetadata{} (as it did not learn any residual noise) nor \minimalMetadata{} (which does not correctly estimate residual noise, cf.~\cref{subsec:corruptedQualitativeExperiments}).
\FloatBarrier
\input{tabFixedCameraparameterRanges}
\input{tabCameraParameterDefinitions}
\input{tabCameraParameterSensitivityAnalysis}
\input{tabKITTIResults}
\input{tabSN06SimulatedRealNoiseResults}
\input{figParkingLotUnderexposure.tex}

\input{figCellarOverexposure.tex}
\input{figICX285ParkingLot.tex}
\input{figEV76C661NoiseSources.tex}
\input{figEV76C661CellarParkingLotTotalNoise.tex}
\input{figICX285CellarTempCorrupted.tex}

%% file: tabNNSetup.tex
\begin{table}[h]
    \centering
    \caption{\emph{\gblue{Details of the proposed noise source estimation network architecture (\fullMetadataTitle)\label{tab:nnsetup}}}}
    \begin{adjustbox}{max width=0.99\linewidth}
        \begin{tabular}{ccc}
            \toprule
            & Output size & Layers\\
            \midrule
            & 128 $\times$ 128 $\times$ 1 & Input image patch (grayscale)\\
            & 128 $\times$ 128 $\times$ 64 & Residual block (5 conv layers)\\
            & 128 $\times$ 128 $\times$ 64 & Residual block (5 conv layers)\\
            \midrule
            \parbox[t]{3.0mm}{\multirow{6}{*}{\rotatebox[origin=c]{90}{PN Branch}}}%
            & 64 & Global max pooling\\
            & 66 & Concatenate (add. input: 2 metadata)\\
            & 32 & Dense\\
            & 16 & Dense\\
            & 8 & Dense\\
            & 1 & Output $\sigma_\text{PN}$\\
            \midrule
            \parbox[t]{3.0mm}{\multirow{6}{*}{\rotatebox[origin=c]{90}{DCSN Branch}}}%
            & 64 & Global max pooling\\
            & 69 & Concatenate (add. input: 5 metadata)\\
            & 32 & Dense\\
            & 16 & Dense\\
            & 8 & Dense\\
            & 1 & Output $\sigma_\text{DCSN}$\\
            \midrule
            \parbox[t]{3.0mm}{\multirow{6}{*}{\rotatebox[origin=c]{90}{RN Branch}}}%
            & 64 & Global max pooling\\
            & 69 & Concatenate (add. input: 5 metadata)\\
            & 32 & Dense\\
            & 16 & Dense\\
            & 8 & Dense\\
            & 1 & Output $\sigma_\text{RN}$\\
            \midrule
            \parbox[t]{3.0mm}{\multirow{6}{*}{\rotatebox[origin=c]{90}{$\xi$ Branch}}}%
            & 64 & Global max pooling\\
            & 67 & Concatenate (incl. $\sigma_\text{PN}$, $\sigma_\text{DCSN}$ and $\sigma_\text{RN}$)\\
            & 32 & Dense\\
            & 16 & Dense\\
            & 8 & Dense\\
            & 1 & Output $\sigma_{\xi}$\\
            \bottomrule
        \end{tabular}%
    \end{adjustbox}
\end{table} 

%% file: noiseModel.tex
\section{Noise Formation Process}
\label{sec:noiseModel}

Image noise is ``any undesired information that contaminates an image'' \cite[p.~348]{bookJayaraman2009}, 
and can be modelled as
\begin{equation} \label{eq:generalNoise}
\tilde{I}(x,y) = I(x,y) + I(x,y)^{\gamma} \, u(x,y)\text{,}
\end{equation}
where $I(x,y)$ is the clean signal intensity, $u(x,y)$ is a random, stationary and uncorrelated noise process, and $\tilde{I}(x,y)$ is the corrupted intensity. 
A parameter $\gamma$ controls different noise types.
The amount of noise (or noise level) may be quantified using the standard deviation $\sigma$ of the underlying statistical distribution of $u(x,y)$.

We only consider time-varying noise sources because time-invariant sources (fixed-pattern noise) are addressed during radiometric calibration (before acquisition) and their residuals are assumed to have a minor influence on image quality. 
The most prominent time-varying noise sources are photon shot noise (PN), dark current shot noise (DCSN), and readout noise (RN).
They are modeled as statistically independent from each other and thus (following error propagation) contribute to the total amount of noise $\totalNoise$ as
\begin{equation}
\totalNoise = \sqrt{\photonNoise^2 + \dcNoise^2 + \readoutNoise^2} + \restNoise,
\label{eq:totalNoiseSupp}%
\end{equation}
with an unexpected noise term $\restNoise$ that we introduce in \eqref{eq:restNoise}.
Next, we review each noise source and also refer to \cite{konnik2014high}.

\textbf{Photon Shot Noise. }%
As photons arrive at the sensor, the counting process within the exposure interval undergoes random fluctuations.
This is known as shot noise and follows a Poisson distribution $P_{\lambda}(k)$, with $\lambda = \sigma^2$ as the expected value and variance of the number of arriving photons $k$.
If $k$ is large enough (i.e., in non-low illumination conditions), the Poisson distribution may be approximated by a Gaussian distribution $\mathcal{N(\lambda, \lambda)}$ using the Central Limit Theorem \cite[p.~225]{devore2011Probability}.
The higher the number of arriving photons, the higher the number of random fluctuations; 
hence PN behaves signal-dependent and can be described by \eqref{eq:generalNoise} when setting $\gamma = 1$ and $u(x,y) \sim P_{\lambda}(k)$.

\textbf{Dark Current Shot Noise. }%
Similar to PN, DCSN originates from the random arrival of dark current electrons and follows the same distribution. 
Dark current emerges from thermally generated electrons at different sensor material regions.
The amount of generated electrons depends, among others, mainly on the pixel area, temperature and exposure time \cite[ch.~7.1.1]{janesick2001Scientific}.
DCSN is signal-independent, hence $\gamma = 0$ in \eqref{eq:generalNoise}.

\textbf{Readout Noise. }%
RN refers to the imperfections due to the sensor's electronic circuitry converting charge into digital values and it is attributed to the on-chip amplification and conversion processing units \cite[p.~197]{Dussault2004Noise}. 
Although readout noise can be reduced to a negligible level in scientific cameras, its impact is still significant for industry-grade sensors that lack of noise reduction \cite[ch.~7.2.9]{janesick2001Scientific}. 
We incorporate sense node reset noise (alias kTC noise) and source-follower noise as the main time-varying components.

Both noise sources can be modeled as zero-mean Gaussian processes, where $\sigma$ depends on the temperature. 
The overall readout noise contribution is a signal-independent addition of both noise processes, hence $\gamma = 0$ in \eqref{eq:generalNoise}.

%% file: algDataPreprocessing.tex
{\small
\begin{algorithm}[t]
    \DontPrintSemicolon
    \caption{Real-World Noise Processing}\label{alg:dataPreprocessing}
    \KwData{Raw RN/DCSN image sessions.}
    \KwResult{Offset/FPN corrected image sessions.}
    \BlankLine
    \BlankLine
    $s_\text{fpn}$ = 20 \Comment*[r]{\#imgs for FPN calculation}
     \ForEach {sess  $\in$  sessions}{
          \If{\#(sess.imgs) $>$ $s_\text{fpn}$}
          {
               $\text{imgs}_{\text{RN,fixed}} = \emptyset$\;
               $\text{imgs}_{\text{DCSN,fixed}} = \emptyset$\;
               \ForEach {$(\text{img}_{\text{RN}}, \text{img}_{\text{DCSN}}$)  $\in$  sess.imgs}{ 
                    
                    $\mu_\text{RN}, \sigma_\text{RN}$ $\gets$ fixNoiseDistr($\text{img}_{\text{RN}}$)\;
                    $\text{img}_{\text{RN, fixed}}$ $\gets$ sampleNormal($\mu_\text{RN}, \sigma_\text{RN}$)\;
                    $\text{imgs}_{\text{RN, fixed}}.\text{insert}$($\text{img}_{\text{RN, fixed}}$)\;
                    \BlankLine
                    $\mu_\text{DCSN}, \sigma_\text{DCSN}$ $\gets$ fixNoiseDistr($\text{img}_{\text{DCSN}}$)\;
                    $\mu_\text{DCSN*} = \mu_\text{DCSN} - \mu_\text{RN}$\;
                    $\sigma_\text{DCSN*} = \sqrt{\sigma^{2}_\text{DCSN} - \sigma^{2}_\text{RN}}$\;
                    $\text{img}_{\text{DCSN, fixed}}$ $\gets$ sampleNormal($\mu_\text{DCSN*}, \sigma_\text{DCSN*}$)\;
                    $\text{imgs}_{\text{DCSN, fixed}}.\text{insert}$($\text{img}_{\text{DCSN, fixed}}$)\;
               }
               correctFPN($\text{imgs}_{\text{RN,fixed}}$, $s_\text{fpn}$) \;
               correctFPN($\text{imgs}_{\text{DCSN*,fixed}}$, $s_\text{fpn}$) \;
          }
     }
    \BlankLine
    \BlankLine
    \SetKwFunction{FMain}{fixNoiseDistr}
    \SetKwProg{Fn}{Function}{:}{}
     \Fn{\FMain{img}}{
        hist $\gets$ histogram(img)\;
        \vspace{-5pt}
        $x_{\text{max}}$ $\gets$ argmax(hist) \KwSty{where} $x_{\text{max}} \stackrel{!}{>} 0$ \;
        $\text{hist}_{\text{fixed}}$ $\gets$ fixHistogram(hist, $x_{\text{max}}$) \;
        $\mu, \sigma$ $\gets$ fitNormal($\text{hist}_{\text{fixed}}$)\;
        \KwRet $\mu, \sigma$\;
    } 
    \BlankLine
    \BlankLine
    \SetKwFunction{FMain}{fixHistogram}
    \SetKwProg{Fn}{Function}{:}{}
     \Fn{\FMain{hist, $x_\text{max}$}}{
        $\text{hist}_{\text{fixed}} = \emptyset$ \;
        \ForEach {(bin, val)  $\in$  hist}{ 
            \If{$bin \geq 2x_\text{max}$}
            {
                $\text{hist}_{\text{fixed}}$.insert($(2x_\text{max} - \text{bin}, \text{val})$)\;
            }
            $\text{hist}_{\text{fixed}}$.insert($(\text{bin}, \text{val})$)\;

        }
        \KwRet $\text{hist}_{\text{fixed}}$\;
    }
    \BlankLine
    \BlankLine
    \SetKwFunction{FMain}{correctFPN}
    \SetKwProg{Fn}{Function}{:}{}
     \Fn{\FMain{$imgs_\text{fixed}$, $s_\text{fpn}$}}{
        $\text{img}_{\text{fpn}}$ $\gets$ meanImg($\text{imgs}_{\text{fixed}}$[0:$s_\text{fpn}$])\;
               \ForEach {$\text{img}_{\text{fixed}}$  $\in$  $\text{imgs}_{\text{fixed}}$[$s_\text{fpn}$:]}{ 
                    $\text{img}_{\text{fixed/noFpn}}$ $\gets$ $\text{img}_{\text{fixed}} - \text{img}_{\text{fpn}}$\;
                    saveImg($\text{img}_{\text{fixed/noFpn}}$, img.filePath)\;
               }
    }
\end{algorithm}
}

%% file: tabFixedCameraparameterRanges.tex
\begin{table}
\ifarxiv \vspace{-30ex} \fi 
    \caption{\emph{Fixed Camera metadata} in the employed noise model.\label{tab:fixedCameraMetadata}}
    \centering
    \begin{tabular}{ll}
        \toprule
        Fixed Parameter & Value \\
        \midrule
        Camera Offset & 0 \si{\DN}\\
        \makecell[l]{CDS Gain} & 1\\
        \makecell[l]{CDS s2s Time} & $10^{-6}$ \si{\second}\\
        \makecell[l]{CDS Time Factor} & 0.5\\
        \makecell[l]{Flicker Noise Corner Freq.} & $10^{-6}$ \si{\hertz}\\
        \makecell[l]{Source Foll. Current Mod.} & $10^{-8}$ \si{\hertz}\\
        \makecell[l]{Source Foll. Gain} & 1\\
        \bottomrule
    \end{tabular}
    \ifarxiv \vspace{6ex} \fi 
\end{table} 

%% file: tabCameraParameterDefinitions.tex
\begin{table}
    \caption{\emph{Camera metadata definitions.}\label{tab:cameraMetadataDefinitions}}    
    \centering
    \begin{adjustbox}{max width=\linewidth}
        \begin{tabular}{ll}
            \toprule
            Camera Parameter & Definition \\
            \midrule
            Camera Gain & Amplification factor applied to the digital camera signal.\\
            Camera Offset & Offset value applied to the digital camera signal.\\
            \makecell[l]{Correlated Double Sampling\\(CDS) Gain} & Left over from CDS that is applied as gain to the signal.\\
            \makecell[l]{CDS Sample-to-Sample\\Time} & \makecell[l]{Time period after which a video is sampled and held\\within the sample-and-hold CDS circuit.} \\
            CDS Time Factor & \makecell[l]{Factor to calculate the dominant time constant from\\the sample-to-saple time.}\\
            Exposure Time & \makecell[l]{Time period in which the sensor is illuminated for\\a single image.}\\
            \makecell[l]{Flicker Noise\\Corner Frequency} & \makecell[l]{Frequency at which the magnitude of a device's white noise\\and flicker noise are equal.}\\
            Sensor Temperature & Temperature of the camera sensor at image acquisition.\\
            Dark Signal Figure of Merit & Quantity to characterize dark signal generation performance.\\
            Full Well Capacity &  Number of charge carriers a camera sensor pixel can hold.\\
            Pixel Clock Rate & \makecell[l]{Rate at which pixels are transferred to fit an entire\\frame of pixels into a single refresh cycle.}\\
            Sense Node Gain & Gain applied for charge to voltage conversion.\\
            Sense Node Reset Factor & Compensation factor of the sense node reset noise from CDS.\\
            Sensor Pixel Size & Height/Width of a single camera sensor pixel.\\
            Sensor Type & Construction type of the camera sensor.\\
            \makecell[l]{Source Follower\\Current Modulation} & Current modulation induced by burst noise.\\
            Source Follower Gain & Voltage amplification applied by the source follower.\\
            Thermal White Noise & White noise component within the source follower.\\
            \bottomrule%
        \end{tabular}%
    \end{adjustbox}
\end{table} 

%% file: tabCameraParameterSensitivityAnalysis.tex
\begin{table}
\caption[Sampled parameter values of camera metadata sensitivity analysis.]{\emph{Camera metadata sensitivity analysis: sampled parameter values.}
    Parameter units are provided in \cref{tab:cameraMetadata}.
    $\star\colon$ Sampled from the non-dB range and converted into \si{\deci\bel} afterwards.
    $\star\star\colon$ Simulated CCD sensor (CMOS sensor otherwise).
    \label{tab:cameraMetadataSensitivityAnalysisParameterValues}
    }
    \centering
        \begin{adjustbox}{max width=\linewidth}
        \setlength{\tabcolsep}{4pt}
        \begin{tabular}{lllllllllll}
            \toprule
            & \multicolumn{10}{c}{Uniform samples of parameter value ranges}\\
            \cmidrule(r){2-11}
            & Min & 1 & 2 & 3 & 4 & 5 & 6 & 7 & 8 & Max \\
            \midrule
            Mean Img Intensity [DN] & 0 & 28 & 56 & 85 &113& 141&170 &198 & 226 & 255\\
            [2ex]
            \textbf{Minimal Metadata}  & & & & & & & & & \\
            $\text{Camera Gain}^{\star}$ & 0 & 8.52 & 12.74 & 15.56 & 17.69 & 19.40 & 20.83 & 22.05 & 23.12 & 24.08\\
            Exposure Time &  0.001 & 0.02 & 0.05 & 0.07 & 0.09 & 0.11 & 0.13 & 0.16 & 0.18 & 0.2\\
            Sensor Temperature &  0.00 & 8.89 & 17.78 & 26.67 & 35.56 & 44.44 & 53.33 & 62.22 & 71.11 & 80.00\\
            [2ex]
            \textbf{Full Metadata}  & & & & & & & & & &\\
            Dark Signal FoM & 0.00 & 0.11 & 0.22 & 0.33 & 0.44 & 0.55 & 0.66 & 0.77 & 0.88 & 1.00\\
            Full Well Capacity [in k]& 2.00 & 12.89 & 23.78 & 34.67 & 45.56 & 56.44 & 67.33 & 78.22 & 89.11 & 100.00\\
            Pixel Clock Rate & 8.00 & 23.78 & 39.56 & 55.33 & 71.11 & 86.89 & 102.67 & 118.44 & 134.22 & 150.00\\
            Sense Node (SN) Gain & 1.00& 1.44 & 1.89 & 2.33 & 2.78 & 3.22 & 3.67 & 4.11 & 4.56 & 5.00\\
            SN Reset Factor & 0.00 & 0.11 & 0.22 & 0.33 & 0.44 & 0.55 & 0.66 & 0.77 & 0.88 & 1.00\\
            Sensor Pixel Size & 0.01 & 0.02 & 0.03 & 0.04 & 0.05 & 0.06 & 0.07 & 0.08 & 0.09 & 1.00\\
            $\text{Thermal Wh. Noise}^{\star\star}$ & 0.10 & 0.76 & 1.41 & 2.07& 2.72 & 3.38 & 4.03 & 4.69 & 5.34 & 6.00\\
            \bottomrule%
        \end{tabular}%
        \end{adjustbox}
\end{table} 

%% file: tabKITTIResults.tex
\begin{table}
\caption{\emph{Noise source estimation on synthetically corrupted (\emph{Random}) and real-world noised (\ICX285{} and \EV76C661{}) KITTI dataset.} The best results per camera and method are highlighted in bold. Note that the uncorrupted KITTI dataset may already contain considerable image noise (cf. \cref{fig:datasetsGTNoise}) and that the results should therefore be interpreted with caution.\label{tab:KITTIResults}}    
    \centering
    \begin{adjustbox}{max width=\linewidth}
        \setlength{\tabcolsep}{4pt}
        \begin{tabular}{llcccccccccccc}
            \toprule
            &&\multicolumn{3}{c}{Photon Shot Noise}&\multicolumn{3}{c}{DCSN}&\multicolumn{3}{c}{Readout Noise} &\multicolumn{3}{c}{Total Noise}%
            \\
            \cmidrule(r){3-5}\cmidrule(r){6-8}\cmidrule(r){9-11}\cmidrule(r){12-14}
            & & Bias & Std & RMS & Bias & Std & RMS & Bias & Std & RMS & Bias & Std & RMS \\
            \midrule
            \parbox[t]{3.0mm}{\multirow{4}{*}{\rotatebox[origin=c]{90}{Random}}}
                &\baselineTitle%
                        & - & - & - %
                        & - & - & - %
                        & - & - & - %
                        & 0.18 & \textbf{0.22} & \textbf{0.28}\\
                &\PGETitle \cite{byun2021fbi}%
                        & 2.03 & 1.16 & 2.33 %
                        & - & - & - %
                        & - & - & - %
                        & 4.00 & 4.80 & 6.24\\
                &\withoutMetadataTitle%
                        & 0.16 & 0.67 & 0.69 %
                        & 0.18 & 2.36 & 2.37 %
                        & 0.61 & 2.33 & 2.41 %
                        & 0.32 & 0.98 & 1.04\\
                &\minimalMetadataTitle%
                        & \textbf{0.04} & 0.66 & 0.66 %
                        & 0.14 & 1.50 & 1.51 %
                        & \textbf{0.04} & 1.92 & 1.92 %
                        & \textbf{0.01} & 0.78 & 0.78\\
                &\fullMetadataTitle%
                        & 0.11 & \textbf{0.14} & \textbf{0.18} %
                        & \textbf{0.05} & \textbf{0.31} & \textbf{0.32} %
                        & 0.10 & \textbf{0.38} & \textbf{0.40} %
                        & 0.03 & 0.34 & 0.35\\
            [0.9ex]%
            \parbox[t]{3.0mm}{\multirow{4}{*}{\rotatebox[origin=c]{90}{ICX285}}}
                &\baselineTitle%
                        & - & - & - %
                        & - & - & - %
                        & - & - & - %
                        & \textbf{0.09} & 0.32 & 0.33\\
                &\PGETitle \cite{byun2021fbi}%
                        & 2.51 & 1.02 & 2.71 %
                        & - & - & - %
                        & - & - & - %
                        & 3.03 & 1.50 & 3.41\\
                &\withoutMetadataTitle%
                        & 0.66 & 0.57 & 0.88 %
                        & 0.53 & 0.60 & 0.80 %
                        & 0.03 & \textbf{0.64} & \textbf{0.64} %
                        & 0.10 & \textbf{0.03} & \textbf{0.11}\\
                &\minimalMetadataTitle%
                        & 0.83 & 0.19 & 0.85 %
                        & 0.69 & 0.78 & 1.05 %
                        & \textbf{0.02} & 1.27 & 1.27 %
                        & 0.12 & 0.31 & 0.34\\
                &\fullMetadataTitle%
                        & \textbf{0.10} & \textbf{0.12} & \textbf{0.16} %
                        & \textbf{0.16} & \textbf{0.49} & \textbf{0.51} %
                        & 0.90 & 1.10 & 1.43 %
                        & 0.45 & 0.76 & 0.89\\
            [0.9ex]%
            \parbox[t]{3.0mm}{\multirow{4}{*}{\rotatebox[origin=c]{90}{EV76C661}}} 
             &\baselineTitle%
                        & - & - & - %
                        & - & - & - %
                        & - & - & - %
                        & 0.05 & 0.23 & 0.23\\
            &\PGETitle \cite{byun2021fbi}%
                        & 2.61 & 0.82 & 2.74 %
                        & - & - & - %
                        & - & - & - %
                        & 3.70 & 1.25 & 3.90\\
                &\withoutMetadataTitle%
                        & 0.60 & 0.52 & 0.79 %
                        & \textbf{0.00} & 1.36 & 1.36 %
                        & \textbf{0.09} & \textbf{1.06} & \textbf{1.07} %
                        & 0.12 & \textbf{0.03} & \textbf{0.12}\\
                &\minimalMetadataTitle%
                        & 1.11 & 0.24 & 1.14 %
                        & 0.43 & 1.18 & 1.26 %
                        & 0.76 & 1.49 & 1.67 %
                        & \textbf{0.02} & 0.18 & 0.18\\
                &\fullMetadataTitle%
                        & \textbf{0.25} & \textbf{0.13} & \textbf{0.28} %
                        & 0.78 & \textbf{0.89} & \textbf{1.19} %
                        & 0.43 & 1.73 & 1.78 %
                        & 0.21 & 1.03 & 1.05\\
            \bottomrule
        \end{tabular}%
    \end{adjustbox}
\end{table} 

%% file: tabSN06SimulatedRealNoiseResults.tex
\begin{table}
\caption{\emph{Noise source estimation on real-world noise extracted from the e2V EV76C661 camera sensor.} DCSN and RN with corresponding metadata were recorded from the camera. PN was generated synthetically using the real metadata. The best results per method and dataset are highlighted in bold. \label{tab:SN06SimulatedRealNoiseEstimation}}
    \centering
    \begin{adjustbox}{max width=\linewidth}
        \setlength{\tabcolsep}{4pt}
        \begin{tabular}{llcccccccccccc}
            \toprule
            &&\multicolumn{3}{c}{Photon Shot Noise}&\multicolumn{3}{c}{DCSN}&\multicolumn{3}{c}{Readout Noise} &\multicolumn{3}{c}{Total Noise}%
            \\
            \cmidrule(r){3-5}\cmidrule(r){6-8}\cmidrule(r){9-11}\cmidrule(r){12-14}
            & & Bias & Std & RMS & Bias & Std & RMS & Bias & Std & RMS & Bias & Std & RMS \\
            \midrule
            \parbox[t]{3.0mm}{\multirow{4}{*}{\rotatebox[origin=c]{90}{Sim}}}
                &\baselineTitle%
                        & - & - & - %
                        & - & - & - %
                        & - & - & - %
                        & 0.14 & \textbf{0.22} & \textbf{0.26}\\
                &\PGETitle \cite{byun2021fbi}%
                        & 3.02 & 0.95 & 3.17 %
                        & - & - & - %
                        & - & - & - %
                        & 3.65 & 1.13 & 3.82\\
                &\withoutMetadataTitle%
                        & 0.47 & 0.55 & 0.73 %
                        & \textbf{0.08} & 1.29 & 1.29 %
                        & \textbf{0.32} & \textbf{1.15} & \textbf{1.19} %
                        & 0.28 & \textbf{0.22} & \textbf{0.26}\\
                &\minimalMetadataTitle%
                        & 1.39 & 0.26 & 1.41 %
                        & 0.41 & 1.14 & \textbf{1.21} %
                        & 0.80 & 1.59 & 1.78 %
                        & 1.88 & 0.46 & 1.94\\
                &\fullMetadataTitle%
                        & \textbf{0.28} & \textbf{0.08} & \textbf{0.30} %
                        & 0.79 & \textbf{0.92} & \textbf{1.21} %
                        & 0.51 & 1.74 & 1.81 %
                        & \textbf{0.02} & 0.26 & \textbf{0.26}\\
            [0.9ex]%
            \parbox[t]{3.0mm}{\multirow{4}{*}{\rotatebox[origin=c]{90}{Tamp.17}}}
                &\baselineTitle%
                        & - & - & - %
                        & - & - & - %
                        & - & - & - %
                        & 0.31 & \textbf{0.37} & 0.48\\
                &\PGETitle \cite{byun2021fbi}%
                        & 3.17 & 1.17 & 3.37 %
                        & - & - & - %
                        & - & - & - %
                        & 3.39 & 1.64 & 3.77\\
                &\withoutMetadataTitle%
                        & 0.38 & 0.62 & 0.73 %
                        & \textbf{0.09} & 1.37 & 1.37 %
                        & 0.69 & \textbf{1.30} & \textbf{1.47} %
                        & \textbf{0.02} & 0.52 & 0.52\\
                &\minimalMetadataTitle%
                        & 1.38 & 0.33 & 1.41 %
                        & 0.45 & 1.17 & 1.25 %
                        & 0.59 & 1.76 & 1.85 %
                        & 1.69 & 0.64 & 1.81\\
                &\fullMetadataTitle%
                        & \textbf{0.34} & \textbf{0.15} & \textbf{0.37} %
                        & 0.79 & \textbf{0.93} & \textbf{1.22} %
                        & \textbf{0.44} & 1.77 & 1.82 %
                        & 0.13 & 0.39 & \textbf{0.41}\\
            [0.9ex]%
            \parbox[t]{3.0mm}{\multirow{4}{*}{\rotatebox[origin=c]{90}{Udacity}}} 
            &\baselineTitle%
                        & - & - & - %
                        & - & - & - %
                        & - & - & - %
                        & 0.10 & 0.42 & 0.43\\
            &\PGETitle \cite{byun2021fbi}%
                        & 2.52 & 0.81 & 2.65 %
                        & - & - & - %
                        & - & - & - %
                        & 3.41 & 0.98 & 3.55\\
            &\withoutMetadataTitle%
                        & 0.32 & 0.46 & 0.56 %
                        & \textbf{0.07} & 1.32 & 1.32 %
                        & 0.48 & \textbf{1.10} &\textbf{ 1.21} %
                        & \textbf{0.06} & 0.42 & 0.43\\
            &\minimalMetadataTitle%
                        & 0.94 & 0.16 & 0.96 %
                        & 0.44 & 1.12 & 1.20 %
                        & 0.64 & 1.58 & 1.70 %
                        & 1.39 & 0.47 & 1.47\\
            &\fullMetadataTitle%
                        & \textbf{0.13} & \textbf{0.10} & \textbf{0.17} %
                        & 0.72 & \textbf{0.93} & \textbf{1.18} %
                        & \textbf{0.37} & 1.71 & 1.75 %
                        & 0.28 & \textbf{0.29} & \textbf{0.40}\\
            \bottomrule
        \end{tabular}%
    \end{adjustbox}
\end{table} 

%% file: figParkingLotUnderexposure.tex
\begin{figure}
    \noindent
    \includegraphics[width=0.48\linewidth, trim=16 16 16 16, clip]{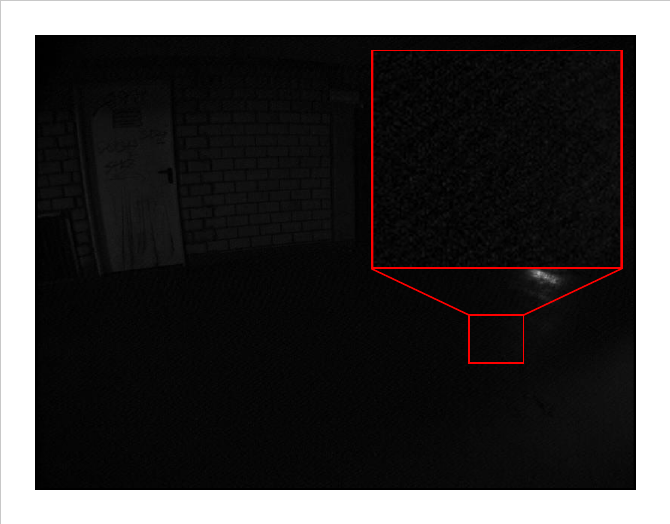}
    \hfill
    \includegraphics[width=0.48\linewidth, trim=16 16 16 16, clip]{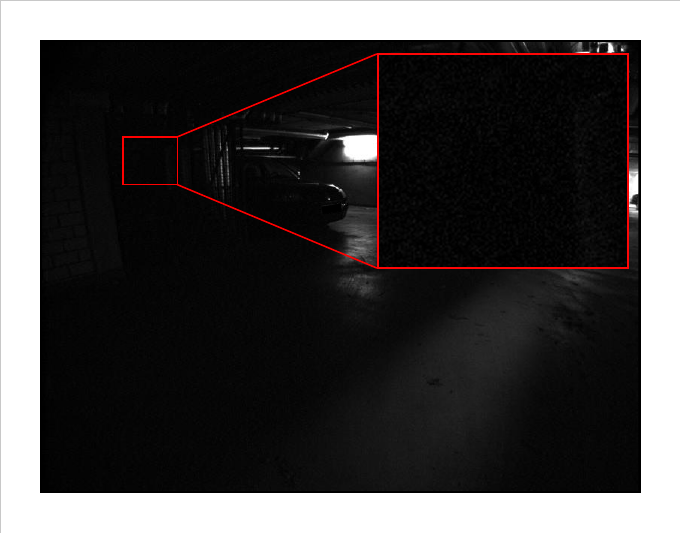}
    \caption{\emph{Exemplary \parkingLot{} images with under-exposed areas. } 
    Timestamps: $t = \SI{50}{\second}$ (Left) and $t = \SI{60}{\second}$ (Right). 
    Details are in \cref{subsec:soundQualitativeExperiments}.
    }
    \label{fig:underexposureExamplesParkingLot}
\end{figure}

%% file: figCellarOverexposure.tex
\begin{figure}
    \noindent
    \includegraphics[width=0.48\linewidth, trim=16 16 16 16, clip]{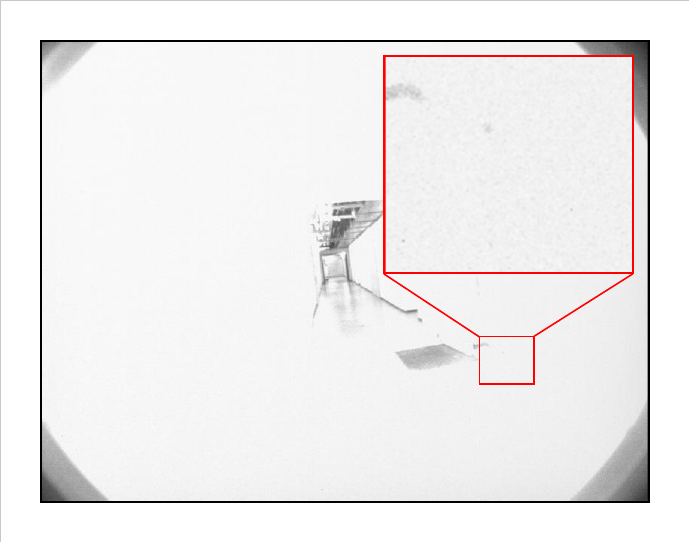}
    \hfill
    \includegraphics[width=0.48\linewidth, trim=16 16 16 16, clip]{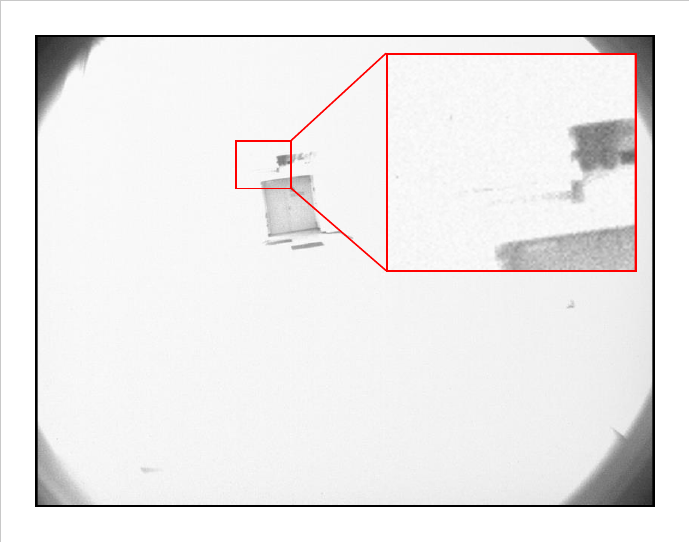}
    \caption{\emph{Exemplary \cellar{} images with over-exposed areas. } Timestamps: $t = \SI{13.3}{\second}$ (Left) and $t = \SI{102.8}{\second}$ (Right). 
    Compare to \cref{fig:EV76C661CellarParkingLotNoiseSources}. 
    Details in \cref{subsec:soundQualitativeExperiments}.
    }
    \label{fig:overexposureExamplesCellar}
\end{figure}

%% file: figICX285ParkingLot.tex
\begin{figure*}
    \tikzsetnextfilename{FigICX285ParkingLot}
    \begingroup
        \pgfplotsset{every axis/.style={scale=0.52}}
        \pgfplotsset{cycle list/Set1, cycle multiindex* list={
                mark list*\nextlist
                Set1\nextlist
            }
        }
        \input{plotICX285ParkingLotNoiseSources.tex}
    \endgroup
    \caption{\emph{Noise source estimation (dataset: \parkingLot{}, camera: \ICX285{})}.
    The model \fullMetadata{} produces anomalies in its estimates roughly at timestamps $\{t| [20, 22] \cup [46, 62] \cup [72, 79]\} \, \si{\second}$, where we noticed large under-exposed areas in the dataset (cf. \cref{fig:underexposureExamplesParkingLot}).
    }
    \label{fig:ICX285ParkingLot}
\end{figure*}
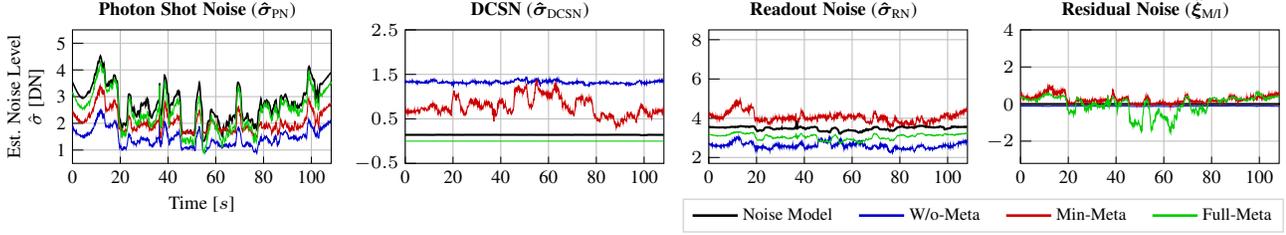

%% file: plotICX285ParkingLotNoiseSources.tex
\pgfplotstableread[col sep = comma]{dataICX285parkingLot1Branch.csv}\csvParkingLotA
\pgfplotstableread[col sep = comma]{dataICX285parkingLot4Branch.csv}\csvParkingLotB
\pgfplotstableread[col sep = comma]{dataICX285parkingLot4BranchConf.csv}\csvParkingLotC
\pgfplotstableread[col sep = comma]{dataICX285parkingLot4BranchAll.csv}\csvParkingLotD
\pgfplotstableread[col sep = comma]{dataICX285parkingLotnoiseModel.csv}\csvParkingLotE
\begin{tikzpicture}
{\scriptsize

\begin{axis}[%
    name=axis1,
    clip mode=individual,
    tick pos=left,
    xmajorgrids,
    xmin=0, xmax=10.83,
    xtick style={color=black},
    xtick={0, 2, 4, 6, 8, 10, 12},
    xticklabels={0,20, 40, 60, 80, 100, 120},
    ymajorgrids,
    ymin=0.5, ymax=5.50,
    ytick style={color=black},
    ytick={1, 2, 3, 4, 5},
    yticklabels={1, 2, 3, 4, 5},
    xlabel near ticks,
    ylabel near ticks,
    xlabel={Time [$s$]},
    xlabel style={yshift=-4pt},
    ylabel style={align=center}, 
    ylabel={Est. Noise Level \\ $\hat{\sigma}$ [DN]},
    title={\textbf{Photon Shot Noise} ($\bm{\photonNoiseEstimate}$)},
    title style={yshift=16pt},
    yscale=0.6,
    xscale=0.965
]
    
\addplot[color=black, semithick] table [x=0, y=4] {\csvParkingLotE};
\addplot[color=blue!80!black] table [x=0, y=4] {\csvParkingLotB};
\addplot[color=red!80!black] table [x=0, y=4] {\csvParkingLotC};
\addplot[color=green!80!black] table [x=0, y=4] {\csvParkingLotD};
\end{axis}

\begin{axis}[%
    name=axis2,
    at=(axis1.right of south east), 
    anchor=left of south west,
    xshift=2ex,
    clip mode=individual,
    tick pos=left,
    xmajorgrids,
    xmin=0, xmax=10.83,
    xtick style={color=black},
    xtick={0, 2, 4, 6, 8, 10, 12},
    xticklabels={0, 20, 40, 60, 80, 100, 120},
    ymajorgrids,
    ymin=-0.5, ymax=2.5,
    ytick style={color=black},
    ytick={-0.5, 0.5, 1.5, 2.5},
    xlabel near ticks,
    ylabel near ticks,
    title={\textbf{DCSN} ($\bm{\dcNoiseEstimate}$)},
    title style={yshift=16pt},
    yscale=0.6,
    xscale=0.965
]

\addplot[color=blue!80!black] table [x=0, y=5] {\csvParkingLotB};
\addplot[color=red!80!black] table [x=0, y=5] {\csvParkingLotC};
\addplot[color=green!80!black] table [x=0, y=5] {\csvParkingLotD};
\addplot[color=black, thick] table [x=0, y=5] {\csvParkingLotE};
\end{axis}
    
\begin{axis}[%
        name=axis3,
        at=(axis2.right of south east), 
        anchor=left of south west,
        xshift=2.5ex,
        clip mode=individual,
        tick pos=left,
        xmajorgrids,
        xmin=0, xmax=10.83,
        xtick style={color=black},
        xtick={0, 2, 4, 6, 8, 10, 12},
        xticklabels={0,20, 40, 60, 80, 100, 120},
        ymajorgrids,
        ymin=1.7, ymax=8.5,
        ytick style={color=black},
        ytick={2, 4, 6, 8},
        xlabel near ticks,
        ylabel near ticks,
        title={\textbf{Readout Noise} ($\bm{\readoutNoiseEstimate}$)},
        title style={yshift=16pt},
        yscale=0.6,
        xscale=0.965
    ]

\addplot[color=blue!80!black] table [x=0, y=6] {\csvParkingLotB};
\addplot[color=red!80!black] table [x=0, y=6] {\csvParkingLotC};
\addplot[color=green!80!black] table [x=0, y=6] {\csvParkingLotD};
\addplot[color=black, thick] table [x=0, y=6] {\csvParkingLotE};
\end{axis}
    
\begin{axis}[%
        name=axis4,
        at=(axis3.right of south east), 
        anchor=left of south west,
        xshift=1.5ex,
        clip mode=individual,
        tick pos=left,
        xmajorgrids,
        xmin=0, xmax=10.83,
        xtick style={color=black},
        xtick={0, 2, 4, 6, 8, 10, 12},
        xticklabels={0,20, 40, 60, 80, 100, 120},
        ymajorgrids,
        ymin=-3.2, ymax=4.0,
        ytick style={color=black},
        ytick={-2, 0, 2, 4},
        xlabel near ticks,
        ylabel near ticks,
        title={\textbf{Residual Noise} ($\bm{\restNoiseEstimation}$)},
        title style={yshift=16pt},
        yscale=0.6,
        xscale=0.965,
        legend cell align={left},
        legend image post style={scale=1.0},
        legend columns=4,
        legend style={
          nodes={scale=0.72, transform shape},
          text opacity=1,
          at={(0, 0)},
          anchor=north east,
          /tikz/every even column/.append style={column sep=8pt},
          font={\small\color{black}},
          line width=0.6pt,
          draw=lightgray,
          draw opacity=0.6,
          fill=none
        },
        legend entries={Noise Model,
                \withoutMetadataTitle,
                \minimalMetadataTitle,
                \fullMetadataTitle},
        legend to name=icx285ParkingLotLegend
    ]

\addlegendimage{color=black, no markers, line width=1.0pt, draw opacity=1.0}
\addlegendimage{color=blue!80!black, line width=1.0pt, draw opacity=1.0}
\addlegendimage{color=red!80!black, line width=1.0pt, draw opacity=1.0}
\addlegendimage{color=green!80!black, line width=1.0pt, draw opacity=1.0}

\addplot[color=black, thick] coordinates {(0,0) (12, 0)};
\addplot[color=blue!80!black] table [x=3, y=7] {\csvParkingLotB};
\addplot[color=red!80!black] table [x=3, y=7] {\csvParkingLotC};
\addplot[color=green!80!black] table [x=3, y=7] {\csvParkingLotD};
\end{axis}
}
\end{tikzpicture}
\tikzexternaldisable
    \hspace*{255pt}
    \raisebox{6pt}[0cm][0cm]{\ref*{icx285ParkingLotLegend}}
\vspace{-5pt}
\tikzexternalenable

%% file: figEV76C661NoiseSources.tex
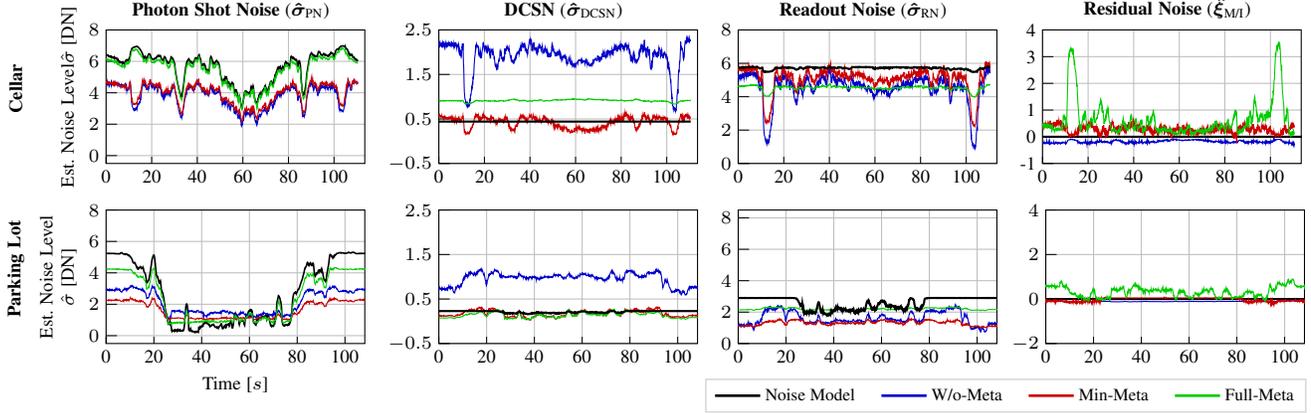
\begin{figure*}
    \tikzsetnextfilename{FigEV76C661CellarParkingLotNoiseSources}
    \begingroup
        \pgfplotsset{every axis/.style={scale=0.52}}
        \input{plotEV76C661CellarParkingLotNoiseSources.tex}
    \endgroup
    \caption{\emph{Noise source estimation (dataset: \cellar{} and \parkingLot{}, camera: \EV76C661{})}. Cellar: All models produce an anomaly in their estimates roughly at timestamps $\{t| [11, 17] \cup [100, 107] \} \, \si{\second}$ that correspond to large over-exposed areas in the images (cf.~\cref{fig:overexposureExamplesCellar}).}
    \label{fig:EV76C661CellarParkingLotNoiseSources}
\end{figure*}

%% file: plotEV76C661CellarParkingLotNoiseSources.tex
\pgfplotstableread[col sep = comma]{dataEV76C661cellar4Branch.csv}\csvCellarB
\pgfplotstableread[col sep = comma]{dataEV76C661cellar4BranchConf.csv}\csvCellarC
\pgfplotstableread[col sep = comma]{dataEV76C661cellar4BranchAll.csv}\csvCellarD
\pgfplotstableread[col sep = comma]{dataEV76C661cellarnoiseModel.csv}\csvCellarE
\pgfplotstableread[col sep = comma]{dataEV76C661parkingLot4Branch.csv}\csvParkingLotB
\pgfplotstableread[col sep = comma]{dataEV76C661parkingLot4BranchConf.csv}\csvParkingLotC
\pgfplotstableread[col sep = comma]{dataEV76C661parkingLot4BranchAll.csv}\csvParkingLotD
\pgfplotstableread[col sep = comma]{dataEV76C661parkingLotnoiseModel.csv}\csvParkingLotE
\begin{tikzpicture}
{\scriptsize
  
\begin{axis}[%
    name=axis1,
    clip mode=individual,
    tick pos=left,
    xmajorgrids,
    xmin=0, xmax=11.35,
    xtick style={color=black},
    xtick={0, 2, 4, 6, 8, 10, 12},
    xticklabels={0,20, 40, 60, 80, 100, 120},
    ymajorgrids,
    ymin=-0.5, ymax=8,
    ytick style={color=black},
    ytick={0, 2, 4, 6, 8},
    xlabel near ticks,
    ylabel={Est. Noise Level \\ $\hat{\sigma}$ [DN]},
    ylabel style={align=center}, 
    ylabel near ticks,
    title={\textbf{Photon Shot Noise} ($\bm{\photonNoiseEstimate}$)},
    title style={yshift=16pt},
    yscale=0.6,
    xscale=0.965
]
    
\addplot[color=black, semithick] table [x=0, y=4] {\csvCellarE};
\addplot[color=blue!80!black] table [x=0, y=4] {\csvCellarB};
\addplot[color=red!80!black] table [x=0, y=4] {\csvCellarC};
\addplot[color=green!80!black] table [x=0, y=4] {\csvCellarD};
\end{axis}

\begin{axis}[%
    name=axis2,
    at=(axis1.right of south east), 
    anchor=left of south west,
    xshift=2ex,
    clip mode=individual,
    tick pos=left,
    xmajorgrids,
    xmin=0, xmax=11.35,
    xtick style={color=black},
    xtick={0, 2, 4, 6, 8, 10, 12},
    xticklabels={0,20, 40, 60, 80, 100, 120},
    ymajorgrids,
    ymin=-0.5, ymax=2.5,
    ytick style={color=black},
    ytick={-0.5, 0.5, 1.5, 2.5},
    xlabel near ticks,
    title={\textbf{DCSN} ($\bm{\dcNoiseEstimate}$)},
    title style={yshift=16pt},
    yscale=0.6,
    xscale=0.965
]

\addplot[color=blue!80!black] table [x=0, y=5] {\csvCellarB};
\addplot[color=red!80!black] table [x=0, y=5] {\csvCellarC};
\addplot[color=green!80!black] table [x=0, y=5] {\csvCellarD};
\addplot[color=black, thick] table [x=0, y=5] {\csvCellarE};
    \end{axis}
    
\begin{axis}[%
        name=axis3,
        at=(axis2.right of south east), 
        anchor=left of south west,
        xshift=2ex,
        clip mode=individual,
        tick pos=left,
        xmajorgrids,
        xmin=0, xmax=11.35,
        xtick style={color=black},
        xtick={0, 2, 4, 6, 8, 10, 12},
        xticklabels={0,20, 40, 60, 80, 100, 120},
        ymajorgrids,
        ymin=0, ymax=8,
        ytick style={color=black},
        ytick={0, 2, 4, 6, 8},
        xlabel near ticks,
        title={\textbf{Readout Noise} ($\bm{\readoutNoiseEstimate}$)},
        title style={yshift=16pt},
        yscale=0.6,
        xscale=0.965
    ]

\addplot[color=blue!80!black] table [x=0, y=6] {\csvCellarB};
\addplot[color=red!80!black] table [x=0, y=6] {\csvCellarC};
\addplot[color=green!80!black] table [x=0, y=6] {\csvCellarD};
\addplot[color=black, thick] table [x=0, y=6] {\csvCellarE};
\end{axis}
    
\begin{axis}[%
        name=axis4,
        at=(axis3.right of south east), 
        anchor=left of south west,
        xshift=2ex,
        clip mode=individual,
        tick pos=left,
        xmajorgrids,
        xmin=0, xmax=11.35,
        xtick style={color=black},
        xtick={0, 2, 4, 6, 8, 10, 12},
        xticklabels={0, 20, 40, 60, 80, 100, 120},
        ymajorgrids,
        ymin=-1, ymax=4.0,
        ytick style={color=black},
        ytick={-1, 0, 1, 2, 3, 4},
        yticklabels={-1, 0, 1, 2, 3, 4},
        xlabel near ticks,
        title={\textbf{Residual Noise} ($\bm{\restNoiseEstimation}$)},
        title style={yshift=16pt},
        yscale=0.6,
        xscale=0.965
    ]

\addplot[color=black, thick] coordinates {(0,0) (12, 0)};
\addplot[color=blue!80!black] table [x=3, y=7] {\csvCellarB};
\addplot[color=red!80!black] table [x=3, y=7] {\csvCellarC};
\addplot[color=green!80!black] table [x=3, y=7] {\csvCellarD};
\end{axis}

\begin{axis}[%
    name=axis5,
    at={(axis1.below south west)},
    anchor=north west,
    yshift=-2ex, 
    clip mode=individual,
    tick pos=left,
    xmajorgrids,
    xmin=0, xmax=10.83,
    xtick style={color=black},
    xtick={0, 2, 4, 6, 8, 10, 12},
    xticklabels={0,20, 40, 60, 80, 100, 120},
    ymajorgrids,
    ymin=-0.5, ymax=8,
    ytick style={color=black},
    ytick={0, 2, 4, 6, 8},
    xlabel near ticks,
    xlabel={Time [$s$]},
    xlabel style={yshift=-4pt},
    ylabel={Est. Noise Level \\ $\hat{\sigma}$ [DN]},
    ylabel near ticks,
    ylabel style={align=center},
    yscale=0.6,
    xscale=0.965
]
    
\addplot[color=black, semithick] table [x=0, y=4] {\csvParkingLotE};
\addplot[color=blue!80!black] table [x=0, y=4] {\csvParkingLotB};
\addplot[color=red!80!black] table [x=0, y=4] {\csvParkingLotC};
\addplot[color=green!80!black] table [x=0, y=4] {\csvParkingLotD};
\end{axis}

\begin{axis}[%
    name=axis6,
    at=(axis5.right of south east), 
    anchor=left of south west,
    xshift=2ex,
    clip mode=individual,
    tick pos=left,
    xmajorgrids,
    xmin=0, xmax=10.83,
    xtick style={color=black},
    xtick={0, 2, 4, 6, 8, 10, 12},
    xticklabels={0, 20, 40, 60, 80, 100, 120},
    ymajorgrids,
    ymin=-0.5, ymax=2.5,
    ytick style={color=black},
    ytick={-0.5, 0.5, 1.5, 2.5},
    xlabel near ticks,
    yscale=0.6,
    xscale=0.965
]

\addplot[color=blue!80!black] table [x=0, y=5] {\csvParkingLotB};
\addplot[color=red!80!black] table [x=0, y=5] {\csvParkingLotC};
\addplot[color=green!80!black] table [x=0, y=5] {\csvParkingLotD};
\addplot[color=black, thick] table [x=0, y=5] {\csvParkingLotE};
\end{axis}
    
\begin{axis}[%
        name=axis7,
        at=(axis6.right of south east), 
        anchor=left of south west,
        xshift=2ex,
        clip mode=individual,
        tick pos=left,
        xmajorgrids,
        xmin=0, xmax=10.83,
        xtick style={color=black},
        xtick={0, 2, 4, 6, 8, 10, 12},
        xticklabels={0,20, 40, 60, 80, 100, 120},
        ymajorgrids,
        ymin=0, ymax=8.5,
        ytick style={color=black},
        ytick={0, 2, 4, 6, 8},
        xlabel near ticks,
        yscale=0.6,
        xscale=0.965
    ]

\addplot[color=blue!80!black] table [x=0, y=6] {\csvParkingLotB};
\addplot[color=red!80!black] table [x=0, y=6] {\csvParkingLotC};
\addplot[color=green!80!black] table [x=0, y=6] {\csvParkingLotD};
\addplot[color=black, thick] table [x=0, y=6] {\csvParkingLotE};
\end{axis}
    
\begin{axis}[%
        name=axis8,
        at=(axis7.right of south east), 
        anchor=left of south west,
        xshift=1ex,
        clip mode=individual,
        tick pos=left,
        xmajorgrids,
        xmin=0, xmax=10.83,
        xtick style={color=black},
        xtick={0, 2, 4, 6, 8, 10, 12},
        xticklabels={0,20, 40, 60, 80, 100, 120},
        ymajorgrids,
        ymin=-2, ymax=4.0,
        ytick style={color=black},
        ytick={-2.0, 0, 2, 4},
        xlabel near ticks,
        yscale=0.6,
        xscale=0.965,
        legend cell align={left},
        legend image post style={scale=1.0},
        legend columns=4,
        legend style={
          nodes={scale=0.72, transform shape},
          text opacity=1,
          at={(0, 0)},
          anchor=north east,
          /tikz/every even column/.append style={column sep=8pt},
          font={\small\color{black}},
          line width=0.6pt,
          draw=lightgray,
          draw opacity=0.6,
          fill=none
        },
        legend entries={Noise Model,
                \withoutMetadataTitle,
                \minimalMetadataTitle,
                \fullMetadataTitle},
        legend to name=EV76C661CellarParkingLotLegend
    ]

\addlegendimage{color=black, no markers, line width=1.0pt, draw opacity=1.0}
\addlegendimage{color=blue!80!black, line width=1.0pt, draw opacity=1.0}
\addlegendimage{color=red!80!black, line width=1.0pt, draw opacity=1.0}
\addlegendimage{color=green!80!black, line width=1.0pt, draw opacity=1.0}

\addplot[color=black, thick] coordinates {(0,0) (12, 0)};
\addplot[color=blue!80!black] table [x=3, y=7] {\csvParkingLotB};
\addplot[color=red!80!black] table [x=3, y=7] {\csvParkingLotC};
\addplot[color=green!80!black] table [x=3, y=7] {\csvParkingLotD};
\end{axis}

\node [rotate=90] at (-1.2, 1.0) {\textbf{Cellar}}; 
\node [rotate=90] at (-1.2, -1.4) {\textbf{Parking Lot}}; 
}
\end{tikzpicture}

\tikzexternaldisable
    \hspace*{263.5pt}
    \raisebox{6pt}[0cm][0cm]{\ref*{EV76C661CellarParkingLotLegend}}
\vspace{-5pt}
\tikzexternalenable

%% file: figEV76C661CellarParkingLotTotalNoise.tex
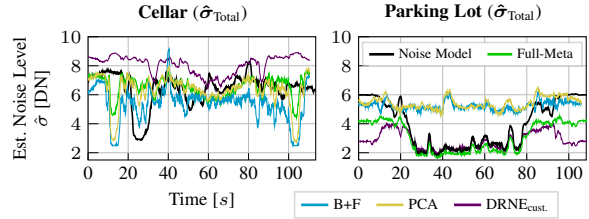
\begin{figure*}
    \tikzsetnextfilename{FigEV76C661CellarParkingLotTotalNoise}
    \begingroup
        \pgfplotsset{every axis/.style={scale=0.48}}
        \centering
        \input{plotEV76C661CellarParkingLotTotalNoise.tex}
    \endgroup
    \caption{\emph{Total noise estimation (dataset: \cellar{} and \parkingLot{}, camera: \EV76C661{}).} Compare to \cref{fig:EV76C661CellarParkingLotNoiseSources}. 
    Details in \cref{subsec:soundQualitativeExperiments}.}
    \label{fig:EV76C661CellarParkingLotTotalNoise}
\end{figure*}

%% file: plotEV76C661CellarParkingLotTotalNoise.tex
\pgfplotstableread[col sep = comma]{dataEV76C661cellar1Branch.csv}\csvCellarA
\pgfplotstableread[col sep = comma]{dataEV76C661cellar4Branch.csv}\csvCellarB
\pgfplotstableread[col sep = comma]{dataEV76C661cellar4BranchConf.csv}\csvCellarC
\pgfplotstableread[col sep = comma]{dataEV76C661cellar4BranchAll.csv}\csvCellarD
\pgfplotstableread[col sep = comma]{dataEV76C661cellarnoiseModel.csv}\csvCellarE
\pgfplotstableread[col sep = comma]{dataEV76C661cellarB+F.csv}\csvCellarF
\pgfplotstableread[col sep = comma]{dataEV76C661cellarPCA.csv}\csvCellarG

\pgfplotstableread[col sep = comma]{dataEV76C661parkingLot1Branch.csv}\csvParkingLotA
\pgfplotstableread[col sep = comma]{dataEV76C661parkingLot4Branch.csv}\csvParkingLotB
\pgfplotstableread[col sep = comma]{dataEV76C661parkingLot4BranchConf.csv}\csvParkingLotC
\pgfplotstableread[col sep = comma]{dataEV76C661parkingLot4BranchAll.csv}\csvParkingLotD
\pgfplotstableread[col sep = comma]{dataEV76C661parkingLotnoiseModel.csv}\csvParkingLotE
\pgfplotstableread[col sep = comma]{dataEV76C661parkingLotB+F.csv}\csvParkingLotF
\pgfplotstableread[col sep = comma]{dataEV76C661parkingLotPCA.csv}\csvParkingLotG
\begin{tikzpicture}
{\scriptsize

\begin{axis}[%
        name=axis1,
        clip mode=individual,
        tick pos=left,
        xmajorgrids,
        xmin=0, xmax=11.35,
        xtick style={color=black},
        xtick={0, 2, 4, 6, 8, 10, 12},
        xticklabels={0,20, 40, 60, 80, 100, 120},
        ymajorgrids,
        ymin=1.5, ymax=10,
        ytick style={color=black},
        xlabel near ticks,
        ylabel near ticks,
        xlabel={Time [$s$]},
        xlabel style={yshift=-4pt},
        ylabel style={align=center}, 
        ylabel={Est. Noise Level \\ $\hat{\sigma}$ [DN]},
        title={\textbf{Cellar} ($\bm{\hat\sigma_\text{Total}}$)},
        title style={yshift=16pt},
         yscale=0.6,
        xscale=0.92,
            legend cell align={left},
            legend image post style={scale=0.6},
            legend columns=3,
            legend style={
              nodes={scale=0.72, transform shape},
              inner xsep=2pt, 
              inner ysep=1pt,
              text opacity=1,
              at={(1.05,1.6)},
              anchor=north east,
              /tikz/every even column/.append style={column sep=4pt},
              font={\footnotesize\color{black}},
              line width=0.6pt,
              draw=lightgray,
              draw opacity=0.6
            },
            legend entries={B+F,
                PCA,
                \baselineTitle
            },
            legend to name=EV76C661TotalNoiseComparisonMethodsLegend
    ]

    \addlegendimage{color=cyan!80!black, line width=1.0pt, draw opacity=1.0}
    \addlegendimage{color=yellow!80!black, line width=1.0pt, draw opacity=1.0}
    \addlegendimage{color=violet!80!black, line width=1.0pt, draw opacity=1.0}
    
    \addplot[color=black, semithick] table [x=3, y=7, col sep=comma] {\csvCellarA};
    \addplot[color=violet!80!black] table [x=3, y expr={sqrt(\thisrowno{4}^2 + \thisrowno{5}^2 + \thisrowno{6}^2)}] {\csvCellarE};
    \addplot[color=green!80!black] table [x=3, y expr={sqrt(\thisrowno{4}^2 + \thisrowno{5}^2 + \thisrowno{6}^2) - \thisrowno{7}}] {\csvCellarD};
    \addplot[color=cyan!80!black] table [x=0, y=7] {\csvCellarF};
    \addplot[color=yellow!80!black] table [x=0, y=7] {\csvCellarG};

\end{axis}

\begin{axis}[%
        name=axis2,
        at=(axis1.right of south east), 
        anchor=left of south west,
        xshift=1ex, 
        clip mode=individual,
        tick pos=left,
        xmajorgrids,
        xmin=0, xmax=10.83,
        xtick style={color=black},
        xtick={0, 2, 4, 6, 8, 10, 12},
        xticklabels={0,20, 40, 60, 80, 100, 120},
        ymajorgrids,
        ymin=1.5, ymax=10,
        ytick style={color=black},
        yscale=0.6,
        xscale=0.92,
        xlabel near ticks,
        ylabel near ticks,
        title={\textbf{Parking Lot} ($\bm{\hat\sigma_\text{Total}}$)},
        title style={yshift=16pt},
        legend cell align={left},
            legend image post style={scale=0.55},
            legend columns=3,
            legend style={
              nodes={scale=0.66, transform shape},
              inner xsep=2pt, 
              inner ysep=1pt,
              text opacity=1,
              at={(1.07,1.6)},
              anchor=north east,
              /tikz/every even column/.append style={column sep=4pt},
              font={\footnotesize\color{black}},
              line width=0.6pt,
              draw=lightgray,
              draw opacity=0.6
            },
            legend entries={Noise Model,
            \fullMetadataTitle
            }
    ]

    \addlegendimage{color=black, no markers, line width=1.0pt, draw opacity=1.0}
    \addlegendimage{color=green!80!black, line width=1.0pt, draw opacity=1.0}
    
    \addplot[color=violet!80!black] table [x=3, y=7] {\csvParkingLotA};
    \addplot[color=black, semithick] table [x=3,y expr={sqrt(\thisrowno{4}^2 + \thisrowno{5}^2 + \thisrowno{6}^2)}] {\csvParkingLotE};
    \addplot[color=green!80!black] table [x=3, y expr={sqrt(\thisrowno{4}^2 + \thisrowno{5}^2 + \thisrowno{6}^2) - \thisrowno{7}}] {\csvParkingLotD};
    \addplot[color=cyan!80!black] table [x=0, y=7] {\csvParkingLotF};
    \addplot[color=yellow!80!black] table [x=0, y=7] {\csvParkingLotG};
\end{axis}
}
\end{tikzpicture}

\tikzexternaldisable
    \hspace*{254.7pt}
    \raisebox{10pt}[0cm][0cm]{\ref*{EV76C661TotalNoiseComparisonMethodsLegend}}
    \vspace{6pt}
\tikzexternalenable

%% file: figICX285CellarTempCorrupted.tex
\begin{figure*}
    \tikzsetnextfilename{FigICX285CellarTempCorrupted}
    \begingroup
        \pgfplotsset{every axis/.style={scale=0.52}}
        \input{plotICX285CellarTempCorrupted.tex}
    \endgroup
     \caption{\emph{Noise source estimation on synthetically doubled \emph{sensor temperature} metadata (dataset: \cellar{}, camera: \ICX285{}).} 
     Compare to \cref{fig:ICX285CellarParkingLotNoiseSources}. 
     Details in \cref{subsec:corruptedQualitativeExperiments}.}
    \label{fig:ICX285CellarTempCorrupted}
\end{figure*}
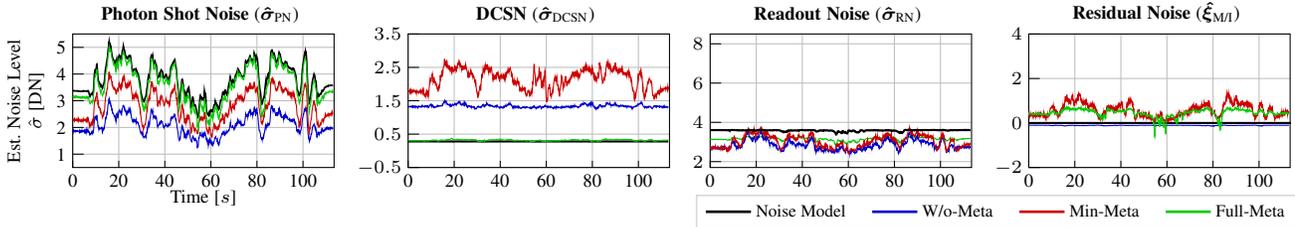

%% file: plotICX285CellarTempCorrupted.tex
\pgfplotstableread[col sep = comma]{dataICX285cellar+doubleTemp4Branch.csv}\csvCellarB
\pgfplotstableread[col sep = comma]{dataICX285cellar+doubleTemp4BranchConf.csv}\csvCellarC
\pgfplotstableread[col sep = comma]{dataICX285cellar+doubleTemp4BranchAll.csv}\csvCellarD
\pgfplotstableread[col sep = comma]{dataICX285cellar+doubleTempnoiseModel.csv}\csvCellarE
\begin{tikzpicture}
{\scriptsize
  
\begin{axis}[%
    name=axis1,
    clip mode=individual,
    tick pos=left,
    xmajorgrids,
    xmin=0, xmax=11.35,
    xtick style={color=black},
    xtick={0, 2, 4, 6, 8, 10, 12},
    xticklabels={0, 20, 40, 60, 80, 100, 120},
    ymajorgrids,
    ymin=0.5, ymax=5.50,
    ytick style={color=black},
    ytick={1, 2, 3, 4, 5},
    yticklabels={1, 2, 3, 4, 5},
    xlabel={Time [$s$]},
    xlabel style={yshift=-4pt},
    xlabel near ticks,
    ylabel near ticks,
    ylabel style={align=center}, 
    ylabel={Est. Noise Level \\ $\hat{\sigma}$ [DN]},
    title={\textbf{Photon Shot Noise} ($\bm{\photonNoiseEstimate}$)},
    title style={yshift=16pt},
    yscale=0.6,
    xscale=0.975,
    legend cell align={left},
        legend image post style={scale=1.0},
        legend columns=4,
        legend style={
          nodes={scale=0.72, transform shape},
          text opacity=1,
          at={(0, 0)},
          anchor=north east,
          /tikz/every even column/.append style={column sep=8pt},
          font={\small\color{black}},
          line width=0.6pt,
          draw=lightgray,
          draw opacity=0.6,
          fill=none
        },
        legend entries={Noise Model,
                \withoutMetadataTitle,
                \minimalMetadataTitle,
                \fullMetadataTitle},
        legend to name=cellarDoubleTempLegend
    ]

\addlegendimage{color=black, no markers, line width=1.0pt, draw opacity=1.0}
\addlegendimage{color=blue!80!black, line width=1.0pt, draw opacity=1.0}
\addlegendimage{color=red!80!black, line width=1.0pt, draw opacity=1.0}
\addlegendimage{color=green!80!black, line width=1.0pt, draw opacity=1.0}
]
    
\addplot[color=black, semithick] table [x=0, y=4] {\csvCellarE};
\addplot[color=blue!80!black] table [x=0, y=4] {\csvCellarB};
\addplot[color=red!80!black] table [x=0, y=4] {\csvCellarC};
\addplot[color=green!80!black] table [x=0, y=4] {\csvCellarD};
\end{axis}

\begin{axis}[%
    name=axis2,
    at=(axis1.right of south east), 
    anchor=left of south west,
    xshift=2ex,
    clip mode=individual,
    tick pos=left,
    xmajorgrids,
    xmin=0, xmax=11.35,
    xtick style={color=black},
    xtick={0, 2, 4, 6, 8, 10, 12},
    xticklabels={0,20, 40, 60, 80, 100, 120},
    ymajorgrids,
    ymin=-0.5, ymax=3.5,
    ytick style={color=black},
    ytick={-0.5, 0.5, 1.5, 2.5, 3.5},
    xlabel near ticks,
    ylabel near ticks,
    title={\textbf{DCSN} ($\bm{\dcNoiseEstimate}$)},
    title style={yshift=16pt},
    yscale=0.6,
    xscale=0.975
]

\addplot[color=blue!80!black] table [x=0, y=5] {\csvCellarB};
\addplot[color=red!80!black] table [x=0, y=5] {\csvCellarC};
\addplot[color=black, thick] table [x=0, y=5] {\csvCellarE};
\addplot[color=green!80!black] table [x=0, y=5] {\csvCellarD};
    \end{axis}
    
\begin{axis}[%
        name=axis3,
        at=(axis2.right of south east), 
        anchor=left of south west,
        xshift=2ex,
        clip mode=individual,
        tick pos=left,
        xmajorgrids,
        xmin=0, xmax=11.35,
        xtick style={color=black},
        xtick={0, 2, 4, 6, 8, 10, 12},
        xticklabels={0,20, 40, 60, 80, 100, 120},
        ymajorgrids,
        ymin=1.7, ymax=8.5,
        ytick style={color=black},
        ytick={2, 4, 6, 8},
        xlabel near ticks,
        ylabel near ticks,
        title={\textbf{Readout Noise} ($\bm{\readoutNoiseEstimate}$)},
        title style={yshift=16pt},
        yscale=0.6,
        xscale=0.975
    ]

\addplot[color=green!80!black] table [x=0, y=6] {\csvCellarD};
\addplot[color=blue!80!black] table [x=0, y=6] {\csvCellarB};
\addplot[color=red!80!black] table [x=0, y=6] {\csvCellarC};
\addplot[color=black, thick] table [x=0, y=6] {\csvCellarE};
\end{axis}
    
\begin{axis}[%
        name=axis4,
        at=(axis3.right of south east), 
        anchor=left of south west,
        xshift=2ex,
        clip mode=individual,
        tick pos=left,
        xmajorgrids,
        xmin=0, xmax=11.35,
        xtick style={color=black},
        xtick={0, 2, 4, 6, 8, 10, 12},
        xticklabels={0, 20, 40, 60, 80, 100, 120},
        ymajorgrids,
        ymin=-2.0, ymax=4.0,
        ytick style={color=black},
        ytick={-2, 0, 2, 4},
        xlabel near ticks,
        ylabel near ticks,
        title={\textbf{Residual Noise} ($\bm{\restNoiseEstimation}$)},
        title style={yshift=16pt},
        yscale=0.6,
        xscale=0.975
    ]

\addplot[color=black, thick] coordinates {(0,0) (12, 0)};
\addplot[color=blue!80!black] table [x=3, y=7] {\csvCellarB};
\addplot[color=red!80!black] table [x=3, y=7] {\csvCellarC};
\addplot[color=green!80!black] table [x=3, y=7] {\csvCellarD};
\end{axis}
}
\end{tikzpicture}

\tikzexternaldisable
    \hspace*{260.0pt}
    \raisebox{5pt}[0cm][0cm]{\ref*{cellarDoubleTempLegend}}
\vspace{1pt}
\tikzexternalenable